\documentclass[mnsc,nonblindrev]{informs3} % current default for manuscript submission

\OneAndAHalfSpacedXI
\usepackage{titlesec}
\titleformat{\subsubsection}[block]{\normalfont\normalsize\bfseries}{\thesubsubsection}{1em}{}
\usepackage{bm}
\usepackage{tikz}
\usetikzlibrary{decorations.pathreplacing}
\usepackage{graphicx}
\usepackage{caption}
\usepackage{subcaption}
\usepackage{multirow}
\usepackage{diagbox}
\usepackage{hyperref}
\usepackage{afterpage}
\usepackage{amsmath, amssymb}

\usepackage{amsmath}
\usepackage{booktabs}
\usepackage{algorithmic}
\usepackage{natbib}
 \bibpunct[, ]{(}{)}{,}{a}{}{,}%
\newcommand{\bibfontsize}{\fontsize{6}{7}\selectfont}
\def\bibfont{\bibfontsize}

\usepackage{nicefrac}       % compact symbols for 1/2, etc.
\usepackage{microtype}      % microtypography
\usepackage{xcolor}         % colors
\usepackage{graphicx}
\usepackage{xcolor, colortbl}
\usepackage{bm}
\usepackage{graphicx}
\usepackage{multirow}
\usepackage{diagbox}
\usepackage{afterpage}
\usepackage{amsmath, amssymb}
\usepackage{algorithmic}
\usepackage{csquotes}
 
\ECRepeatTheorems

\EquationsNumberedThrough   
\TheoremsNumberedThrough
\MANUSCRIPTNO{MS-0000-0000.00}

\usepackage{algorithm}
\usepackage[color=yellow]{todonotes}
\usepackage{xcolor}
\hypersetup{colorlinks=true,pdfnewwindow=true,pdfstartview=FitH,%
pdftitle="",pdfauthor="",%
urlcolor=blue!90!red!45!black,citecolor=blue!90!red!45!black,linkcolor=red!90!black}
\usepackage{enumitem}
\usepackage{tikz}
\definecolor{mygray}{RGB}{195,195,195}

\begin{document}
%%%%%%%%%%%%%%%%

%\setcounter{page}{1}
%\DOI{10.1287/xxxx.0000.0000}%

%\RUNAUTHOR{Jones}
\RUNAUTHOR{Radvand et al.}

%\RUNTITLE{A Cryptography-Based Approach to Privacy-Preserving Revenue Management}
\RUNTITLE{A Training-free Method for LLM Text Attribution}
% \TITLE{Bundling Information Goods of Decreasing Value}
\TITLE{A Training-free Method for LLM Text Attribution \\
%\AUTHOR{\bf(Authors' names blinded for peer review)}
   %\small\textcolor{red}{Compiled \today}
}

%\ARTICLEAUTHORS{%
%\AUTHOR{\bf(Authors' names blinded for peer review)}
%} % end of the block

\ARTICLEAUTHORS{%
\AUTHOR{Tara Radvand}
\AFF{Ross School of Business, University of Michigan, United States, \EMAIL{\href{mailto:tararad@umich.edu}{tararad@umich.edu}}} %, \URL{}}
\AUTHOR{Izak Duenyas}
\AFF{Ross School of Business, University of Michigan, United States, \EMAIL{\href{mailto:duenyas@umich.edu}{duenyas@umich.edu}}} %, \URL{}}
\AUTHOR{Ambuj Tewari}
\AFF{Department of Statistics, University of Michigan, United States, \EMAIL{\href{mailto:tewaria@umich.edu }{tewaria@umich.edu}}}
 %Enter all authors
} % end of the block

\ABSTRACT{
Verifying the provenance of text is increasingly important for firms, educational institutions, and online platforms as Large Language Models (LLMs) produce output that is nearly indistinguishable from human-generated content. We study the problem of determining whether a given text was generated by a particular LLM while controlling the false positive rate. We model LLM-generated text as a sequential stochastic process and develop training-free statistical tests to (i) distinguish between text produced by two known sets of LLMs and (ii) determine whether text was generated by a known LLM or by a distinguishable unknown source, such as a human or another model. We prove that both Type I and Type II errors decay exponentially with text length, establish analogous guarantees for black-box access via sampling, and provide an information-theoretic lower bound showing that there exist model pairs for which no statistical test can make both errors decay faster than exponentially with text length. Numerical experiments empirically evaluate the tests in practical settings and demonstrate strong overall performance, including under many adversarial edits. Our framework provides rigorous guarantees for LLM provenance detection, with applications to content verification, institutional compliance, and misinformation mitigation.}

\maketitle
\vspace{-10mm}
\section{Introduction}

Large Language Models (LLMs) are widely used in journalism, academic research, publishing, and daily business tasks. As these tools advance, determining whether a specific LLM produced a text becomes increasingly difficult. Yet, provenance verification is essential for integrity, accountability, and valuation in digital communication: it helps assign responsibility for content, assess how customers value content from different sources, prevent model quality from deteriorating when AI providers train models on too much machine-generated data, enforce the use of models that satisfy organizational compliance requirements, and help users assess what information to trust.
%Large Language Models (LLMs) are widely used in journalism, academic research, and daily business tasks. As these tools advance, determining whether a specific LLM produced a text becomes increasingly difficult. Yet, provenance verification is essential for integrity and accountability in digital communication: it helps assign responsibility for content, enforce the use of models that satisfy organizational compliance requirements, and prevent model quality from deteriorating when AI providers train models on too much machine-generated data.

Provenance verification helps assign responsibility, as unauthorized LLM use allows individuals to present content they did not author. The need to identify AI-generated content spans examinations, peer review, and publishing, including in academic conferences \citep{neurips2026aipapers} and, following controversies and costly publication withdrawals, in book publishing and literary prizes \citep{newyorktimes2026aifiction,guardian2026publishers,commonwealth2026update,csmonitor2026publishing}.\footnote{Among the cited sources, \citet{csmonitor2026publishing} interviewed one of the authors of this paper. The citation and this footnote are omitted from the anonymous-review version to preserve author anonymity.} Yet, AI use is often undisclosed: \citet{ScienceAIUse2025} reports that 36\% of abstracts among 7{,}177 submitted manuscripts contained at least some AI-generated text, but only 9\% of papers disclosed AI use, making reliable provenance verification necessary. Provenance verification also enables managers to assess whether content source affects customers' valuations and willingness to pay \citep{raj2026artificial}, e.g., by comparing the customers' willingness to pay for human- versus AI-generated content or for a certain LLM's outputs versus those of competing LLMs. Additionally, AI providers use provenance information to protect model development: they must prevent excess synthetic data from entering training corpora and inducing model collapse \citep{shumailov2023curse}. Provenance verification also helps organizations enforce appropriate use of in-house LLMs developed or approved to satisfy security, compliance, or academic-integrity requirements. Universities increasingly use proprietary systems such as Purdue's GenAI Studio and the University of Michigan's UM GPT, and corporations are increasingly using internal platforms such as Goldman Sachs' GS AI Assistant, JPMorgan's LLM Suite, and Samsung's Gauss, among others, to comply with organizational policies. Accurate attribution to sanctioned or non-sanctioned models is therefore essential for accountability and policy enforcement in these settings. Finally, at a societal scale, LLMs enable large-scale, targeted content generation that can influence audiences, as illustrated by attempts to manipulate financial markets \citep{nbc2024fakebiden}, political discourse \citep{nyt2023hairpin}, and consumer sentiment \citep{jakesch2023co}. Provenance verification therefore helps assign responsibility for misleading or harmful LLM-generated content and allows users to assess what information to trust.

These practical needs motivate our research question. LLMs generate text sequentially and probabilistically: at each step, the next token is drawn from a conditional probability distribution that depends on the prompt and all preceding tokens. A manager may have full knowledge of these conditional next-token probability distributions (the white-box setting) or only sampling access to them (the black-box setting).\footnote{Open-weight model families such as Llama, Mistral, Qwen, and Gemma release their model weights and therefore permit white-box access. Several closed-weight APIs also provide token-level log-probabilities, including OpenAI’s Chat Completions API, Cohere’s Chat API, and the Gemini generateContent API when supported by the selected model. A few closed-weight models, such as Anthropic’s Claude API and Google’s Gemini Live API do not expose token-level log-probabilities. In these cases, the conditional next-token distributions can instead be estimated by repeatedly sampling the model, using the preceding tokens, which corresponds to the black-box setting.} Our research question is whether, given a finite text, the manager can determine with high accuracy whether it was generated by a particular LLM while guaranteeing that the probability of falsely attributing text from another distinguishable source to that model remains low. For managerial use, it is also important to characterize how these guarantees depend on the relevant problem parameters, including the text length and the manager-chosen tolerance that governs the trade-off between the false-positive rate and the true-positive rate. Characterizing these relationships allows the manager to determine, for example, how the required text length must change to maintain the same false-positive and true-positive guarantees when the tolerance is adjusted for practical reasons.

The literature on LLM text detection consists of three main streams: learning-based detectors, watermarking, and zero-shot detectors. In learning-based methods, neural classifiers train models on labeled human and LLM-generated text (e.g., as in \citealt{guo2023close}), but such methods require a separate classifier for each evolving LLM family, rely on extensive collections of human-written passages that raise privacy concerns, and generalize poorly outside their training domains. As an example of poor generalization, \citet{liang2023gpt} shows that neural detectors misclassified TOEFL essays by non-native English speakers as LLM-generated in 48–76\% of cases. Watermarking-based methods (e.g., as in \citealp{li2025statistical}) modify the generation process to embed hidden statistical signals in the model’s token choices that are unnoticeable to humans but detectable by algorithms that know the watermarking rule. Their main limitation is that they require cooperation from AI providers, whose incentives regarding openness, competition, and user flexibility may discourage adoption \citep{nature2024watermarking,brookings2024watermarking}. For example, an internal OpenAI survey found that approximately 30\% of ChatGPT users would decrease their use of the product if it deployed watermarking and a rival did not \citep{businessinsider2024homeworkcheating,verge2024openaiwatermark}.

The limitations of learning-based and watermarking approaches motivate interest in zero-shot detectors. A zero-shot detector requires neither labeled examples for detector training nor provider-embedded signals and instead identifies LLM-generated text using statistical properties of the text under evaluation (e.g., as in \citealp{mitchell2023detectgpt}). This reliance on statistical properties creates an opportunity to design statistical tests with provable guarantees for content provenance. However, most existing zero-shot detection methods remain heuristic and lack theoretical guarantees that ensure high accuracy in detecting text generated by a given LLM while maintaining a low false-positive rate.

%This reliance on statistical properties enables the design of statistical tests with provable guarantees for content provenance. However, to our knowledge, this has not been done prior to our work. Existing zero-shot detection methods remain heuristic and lack theoretical guarantees that ensure high accuracy in detecting text generated by a given LLM, while maintaining a low false-positive rate.

\noindent\textbf{What is the research gap?} Despite progress in LLM attribution, existing methods before our work do not provide user-controllable, theoretically grounded guarantees for determining whether a text was generated by a given LLM while keeping the false-positive probability below a user-specified low level. This gap is especially consequential in high-stakes disciplinary and regulatory settings: even a low empirically observed false-positive rate is insufficient when a detector relies on heuristic-based methods, because consequential decisions require a transparent, interpretable basis for attribution along with explicit error guarantees. Students have lost points or gone through difficult misconduct investigations after being accused of using AI, even when they denied it and evidence suggested otherwise \citep{nyt2025falseai,guardian2024aicheating}\footnote{A New York court recently annulled an academic-integrity finding involving a 100\% Turnitin score and ordered the student's record expunged after the university failed to consider contrary evidence or provide a meaningful opportunity to be heard \citep{newby2026}. These concerns have prompted institutional responses: Vanderbilt University disabled Turnitin's AI detector after noting that it had no insight into how the tool worked; the University of Pittsburgh and Johns Hopkins University likewise disabled the tool because of false-positive concerns \citep{vanderbilt2023turnitin,pitt2026academicintegrity,jhu2025detection}.}. False attribution also has serious consequences in book publishing: it can cost an author a contract and damage their reputation, and publishing and literary-prize organizations have cautioned that evidence produced by nontransparent detection methods should not be treated as conclusive for consequential decisions \citep{authorsguild2026detectors,commonwealth2026update,silman2026publishing}. Recent evaluations also show that detector performance varies across tools, highlighting the importance of detection thresholds according to the decision maker's tolerance for false attribution \citep{jabarian2025artificial}. A method addressing this gap would be most useful if its guarantees remained informative for practical-length texts, which are central in many provenance-sensitive settings, and if its error probabilities decreased rapidly with text length, enabling meaningful guarantees for relatively practical-length texts.

\begin{figure}[H]
    \centering
\includegraphics[width=0.45\linewidth]{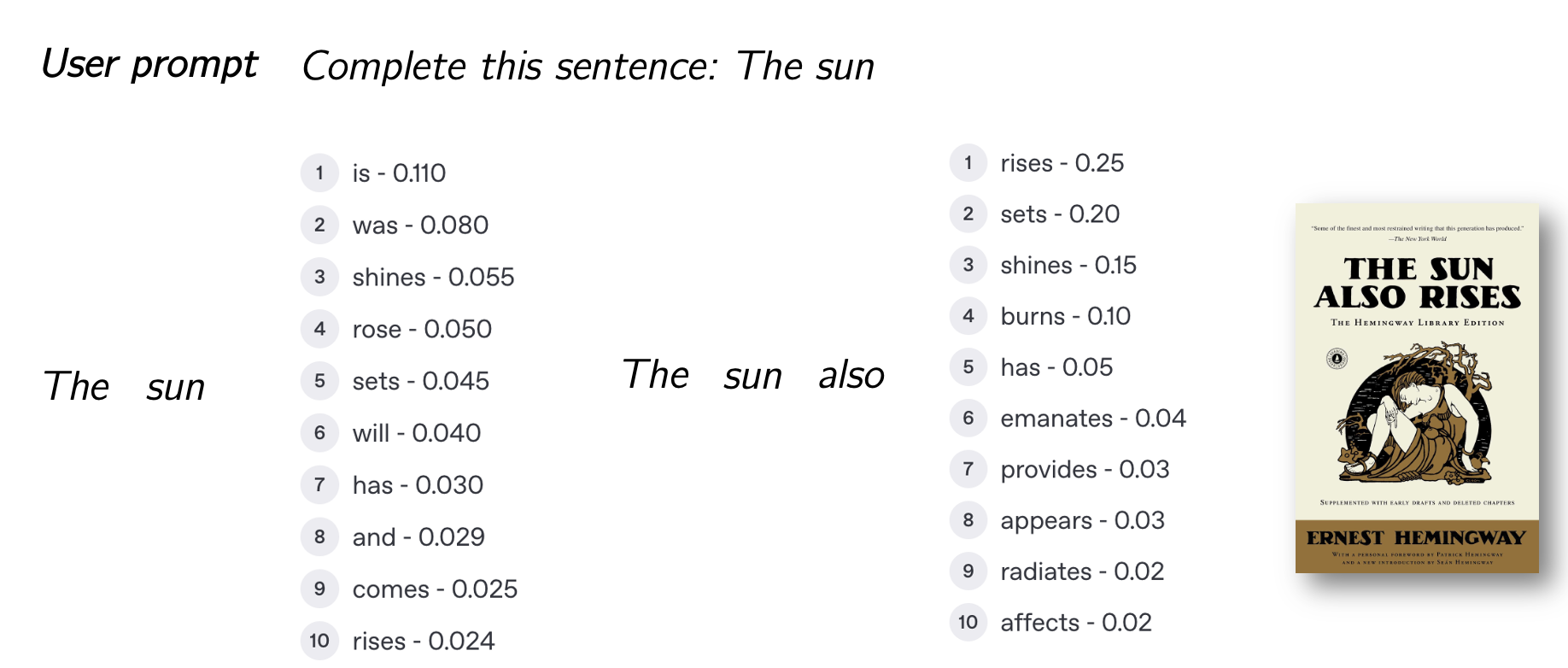}
    \caption{Illustration of token-by-token LLM text generation}
\label{fig:llm_text_generation}
\end{figure}

We provide a theoretical framework for determining whether a given finite-length string was produced by language model $A$ (the evaluator model) or by another source $B$, which may be another language model, a human, or any other text-generation process. Here, the evaluator model $A$ is the model used to assess the text, while the generative source $B$ is the system that actually produced it. We model LLM-generated text as a sequential stochastic process with complete dependence on history to capture the token-by-token text generation in auto-regressive models: at each step, the model uses the context available so far (all previously generated tokens) to assign probabilities to candidate next tokens, and then selects one token according to this context-dependent distribution. Because the context is updated after each generated token, the distribution over the next token can change significantly from one step to the next; therefore, the current token cannot be treated as independent of earlier tokens. Figure~\ref{fig:llm_text_generation} illustrates this dependence: after the prompt \emph{“Complete this sentence: The sun,”} the model assigns a distribution over candidate next tokens; selecting a relatively low-probability continuation such as \emph{“also”} can still occur, and it can substantially reshape the subsequent distribution—here shifting the probability distribution toward tokens like \emph{“rises”} that align with the well-known book title \emph{The Sun Also Rises}.

Building on this model, we develop finite-sample hypothesis tests for LLM provenance that take as input a text string and a given evaluator LLM\footnote{For the evaluator model, we consider three settings. For ease of exposition, we first present theoretical results in a white-box setting. We then extend these results to a black-box setting and characterize the sufficient sample size required to achieve analogous Type I and Type II errors as in the white-box case (Appendix \ref{sec:blackbox}). Finally, we present numerical experiments for a no-access setting, in which we have no access to the evaluator model, not even through sampling (Section \ref{blackboxnumerical}).}, and assess whether that evaluator could plausibly have produced the text. First, we construct composite tests that distinguish whether a text was produced by some model in a set $\mathcal{A}$ or by a model in a disjoint set $\mathcal{B}$, capturing distinctions such as sanctioned in-house models versus non-sanctioned models. Second, we develop a composite test that distinguishes whether the text was generated by $A$ or by a distinguishable unknown alternative “not $A$,” which may include other LLMs, humans, or other language generation processes. Our theoretical results show that the Type I and Type II errors of our tests decay exponentially with string length.

The backbone of our tests is the convergence of two statistical properties: the \textit{log-perplexity} of a string under an evaluator model $A$, and the \textit{average cross-entropy} between a generator $B$ and the evaluator $A$. Log-perplexity is the average negative log-likelihood per token that $A$ assigns to the observed text. It measures how surprising the string is to $A$: low values indicate that $A$ finds the sequence predictable, while high values indicate that $A$ finds the string unexpected. The average cross-entropy between a generator $B$ and the evaluator $A$ describes what log-perplexity is when text is drawn from $B$ and evaluated under $A$. It represents the long-run average log-perplexity that $A$ assigns to text generated by $B$, and when $A$ itself is the generator, this quantity reduces to the entropy rate of $A$, capturing the model’s level of uncertainty about its own next-token predictions. Because we base our tests on the convergence of log-perplexity and average cross-entropy, it helps to see why convergence to average cross-entropy provides the right benchmark. Example \ref{iidtokengenerators} provides intuition.

%This quantity, in isolation, is widely used in attribution and detection tasks, e.g., \citep{mitchell2023detectgpt,solaiman2019release}, motivated by the idea that a model is unlikely to produce text it finds surprising. 
%\begin{example} \label{AversusgreedyA}\textbf{($A$ versus greedy $A$).} Imagine that the unknown generator $B$ is a greedy version of $A$ that always outputs the most likely next token according to $A$. The text generated by $B$ would consist of very high-probability tokens under $A$, leading to its log-perplexity under $A$ being lower than that of typical samples drawn from $A$. As a result, a rule that declares ``$A$ generated the text, since the log-perplexity under $A$ is small'' would then misclassify greedy-$A$'s text as being even more representative of $A$ than the text produced by $A$.\end{example}

\begin{example}\label{iidtokengenerators}
\textbf{(i.i.d.\ token generators).} Consider a simplified setting where both $A$ and $B$ generate tokens independently from fixed distributions: $A$ draws tokens i.i.d.\ from $p$, and $B$ draws tokens i.i.d.\ from $q$. By the weak law of large numbers and the central limit theorem, if $A$ generates the text, then for a sequence of length $N$, the log-perplexity that $A$ assigns to its own output converges to the entropy $H(p)$, with fluctuations shrinking on the order of $1/\sqrt{N}$. If instead $B$ generates the text, the average log-perplexity under $A$ converges to the cross-entropy $H(q,p)$. Thus, the difference between the log-perplexity and $H(p)$ serves as a detection signal: it concentrates near zero when $A$ generates the text and converges to $H(q,p)-H(p)$ when $B$ generates it.
\end{example}

Importantly, in our setting, aligned with practice, LLMs produce non-i.i.d.\ tokens. Hence, the limit theorems and concentration inequalities for i.i.d.\ random variables are not applicable in this setting. By representing the deviation between log-perplexity and average cross-entropy as an average of martingale differences and deriving the required conditional sub-exponential bounds, we show that the log-perplexity concentrates around the average cross-entropy, with the deviation probability decreasing exponentially in text length, making attribution feasible for finite, practical-length texts. This convergence property distinguishes (with high probability) text generated by $A$ from text generated by distinguishable alternative sources, and this is the foundation for our tests.

%Thus, i.i.d.\ limit theorems do not provide the finite-sample guarantee required for attribution. We exploit the sequential generation structure by centering each token’s surprisal under evaluator model A by its conditional expectation under generator B; the resulting deviation between log-perplexity and average cross-entropy is an average of martingale differences with respect to the growing token history. Deriving the corresponding conditional sub-exponential scales yields an exponential concentration bound, making attribution feasible for finite, possibly short texts. This convergence property distinguishes, with high probability, text generated by A from text generated by other sources and forms the foundation of our tests.

\noindent\textbf{Main contributions.} We study LLM provenance under an auto-regressive text-generation model, where each token depends on all prior tokens, making finite-sample analysis difficult because i.i.d.\ tools do not apply. Our key technical idea is to define per-token random variables so that the deviation between the log-perplexity and its associated entropy/cross-entropy target can be written as an average of martingale differences. This martingale representation allows us to invoke concentration inequalities for martingales and thereby obtain explicit, non-asymptotic error guarantees for hypothesis tests on finite, practical-length texts. Below, we summarize our key contributions.
%Our key idea rewrites the gap between log-perplexity and its entropy/cross-entropy target as an average of martingale differences, enabling sharp martingale concentration bounds and explicit, non-asymptotic error guarantees for hypothesis tests on finite (even short) texts.
%We study LLM provenance under an auto-regressive model of text generation. Dependence on the history is a defining feature of how LLMs produce text. That is, conditioned on the prompt and the evolving prefix, each next token is drawn from a context-dependent distribution. The full dependence of history makes rigorous finite-sample analysis complex, since standard i.i.d.\\ tools, e.g., laws of large numbers or classical concentration for independent random variables, do not apply. Our key technical idea is to define per-token random variables so that the deviation between the log-perplexity and its associated entropy/cross-entropy target can be written as an average of martingale differences. This martingale representation makes the problem tractable: it allows us to invoke sharp concentration inequalities for martingales and thereby obtain explicit, non-asymptotic error guarantees for hypothesis tests on finite, possibly short texts. Below, we summarize our key contributions.

\textit{Statistical tests.} We develop two zero-shot statistical tests with theoretical guarantees for finite, practical-length text: an \emph{attribution} test and a \emph{detection} test. In Section \ref{sec:multi}, the attribution test identifies whether the text was generated by an LLM from a known in-house set or a known non-sanctioned set; it applies when the candidate models can be specified in advance, such as when an institution distinguishes approved models from a specified set of prohibited models. In Section \ref{sec:human}, the detection test identifies whether the text was generated by one known LLM or by an unknown distinguishable alternative source that can be an unknown language model, a human, or any other text generation process; it applies, for example, when a university checks whether a particular banned LLM was used for an assignment or when an organization prohibits a specified LLM. The tests address different practical settings and also require different information. For both tests, we prove that Type I and Type II errors decay exponentially with text length, so additional tokens rapidly reduce errors and improve reliability. Numerical experiments in Sections \ref{exp:attribution} and \ref{whiteboxnumerical} empirically evaluate the tests in practical settings and demonstrate strong overall performance. The text-length experiments in Section~\ref{app:length experiments} further show strong performance on practical-length texts.\smallskip

%Numerical experiments in Sections \ref{exp:attribution} and \ref{whiteboxnumerical} validate these results, and the text-length experiments in Section~\ref{app:length experiments} demonstrate strong performance on practical-length texts. 
 
%\todo[inline]{\color{red} Section 5.2 is not a more specific version of 5.1, but it shows that I need to provide discussion and clarification on the differences between the two. Let us discuss in our meeting and I will revise accordingly.}

%We also provide supporting experiments showing that log-perplexity converges to the average string entropy as text length increases, and that it works for practical-length texts of approximately 100 tokens in Appendix \ref{app:additional experiments}.

\textit{Concentration inequalities for convergence of dependent random variables.} To establish our theoretical results, we consider a sequence of discrete random variables generated under probability laws $p_n^B(\cdot)$. In Theorem \ref{thm:maindiff}, we propose and prove an exponential decay concentration inequality that bounds the tail probability of the difference between the log-perplexity of this sequence evaluated under another sequence of probability laws $p_n^A(\cdot)$, and the average cross-entropy of $p_n^B(\cdot)$ and $p_n^A(\cdot)$, over a finite alphabet. The special case in which (i) random variables are independent, categorical variables and (ii) $p_n^{A}(\cdot)=p_n^{B}(\cdot)$, is studied in \cite{zhao2022optimal}, and their paper identifies extending these results to non-independent random variables as an intriguing open question. Since LLMs generate text sequences that depend on previously generated tokens and our setting also allows $p_n^{A}(\cdot)\neq p_n^{B}(\cdot)$, we generalize the results in \cite{zhao2022optimal} by relaxing (i) and (ii). Doing so, we address the open question in \cite{zhao2022optimal}.

\textit{Sufficient sample size for the black-box setting.} We extend our theory to a black-box setting in which we have only sampling access to the evaluator model and cannot query its conditional probability distributions (Appendix \ref{sec:blackbox}). In this setting, we construct empirical log-perplexity and empirical average cross-entropy from a finite number of samples and explicitly characterize the sufficient sample size required to achieve analogous Type~I and Type~II error guarantees as in the white-box setting. Section \ref{BBexperiments} empirically examines how detection performance changes with the number of samples and shows that, by $m=1000$, performance nearly matches the exact-probability benchmark for most settings, while some settings may require a larger sample size.

The rest of this paper is organized as follows: Section \ref{litrev} reviews the literature. Section \ref{sec:modelandbackground} presents our mathematical model. Section \ref{sec:theoreticalresults} provides concentration bounds for the log-perplexity around entropy and cross-entropy targets. Section \ref{sec:stats} presents our statistical tests and the upper bound on their Type I and Type II errors. Appendix \ref{sec:blackbox} extends our tests in Section \ref{sec:stats} to a black-box setting where the evaluator log-probabilities must be approximated via sampling. For clarity, we first present our tests and guarantees in the white-box setting in Section \ref{sec:stats} and then show how they extend to the black-box setting in Appendix \ref{sec:blackbox}, so readers can understand the core statistical ideas while also seeing how our results apply to the black-box with sampling setting. Section \ref{sec:lowerbounds} provides a lower bound by showing that there exist model pairs for which no statistical test can make both errors decay faster than exponentially in text length. Section \ref{sec:experiments} reports our numerical experiments. Section \ref{sec:conclusion} concludes. Appendices \ref{app:concentrationinequalroadmap}-\ref{appendixlowerbound} present the proofs, and Appendix \ref{app:experimentdetails} presents black-box-with-sampling setting experiments and the experiment details.

\section{Literature review} \label{litrev}
We first include the related literature in operations management (OM), and then review the literature on learning-based, watermarking, and zero-shot methods.

\noindent\textbf{Concentration inequalities in OM.} Our paper relates to a broad OM literature using finite-sample concentration inequalities to derive high-probability performance guarantees for sequential, history-dependent stochastic systems in which data are endogenously generated by past decisions and thus are non-i.i.d.\ A major stream studies learning while doing in inventory and revenue management, where firms jointly optimize and learn unknown demand over time, including dynamic inventory control \citep{ChenChao2020} and data-driven posted-pricing mechanisms \citep{jin2024sample}. In the predict-then-optimize framework, other work establishes finite-sample generalization bounds linking predictive accuracy to the quality of downstream optimization decisions \citep{el2019generalization}. Concentration-based confidence bounds are also central to online learning and bandit models, providing finite-time guarantees under adaptive feedback \citep{kalvit2023complexity,JohariKambleKanoria2021Matching,baek2023ts}. Related work examines sequential, history-dependent decision-making \citep[]{Alizamir2022SearchPressure}. Our paper connects to this literature by studying LLM text generation, where each token’s probability distribution depends on all preceding tokens, yielding a sequential, history-dependent stochastic process. Our technical idea is to define per-token random variables so that the deviation between log-perplexity and its entropy/cross-entropy target is written as an average of martingale differences. This representation lets us invoke martingale concentration inequalities and obtain explicit, non-asymptotic error guarantees for our hypothesis tests on finite, practical-length texts.

\noindent\textbf{AI Governance in OM.} When text can influence consumer decisions, firm reputations, or sales, there are incentives to fabricate it, as illustrated by review fraud and strategic manipulation \citep{Dellarocas2006StrategicManipulation,LucaZervas2016FakeIt}. LLMs can accelerate this fabrication by reducing the cost of generating persuasive content at scale. Relevant work emphasizes the need for responsible AI capabilities in organizations and guardrails in education \citep{Tambe2026Reskilling,FugenerWalznerGupta2026Roles,BastaniEtAl2025Guardrails}. Other related work examines how organizations should design human–AI collaboration, including the incentives needed to sustain human oversight \citep{BastaniCachon2025ContractingParadox} and governance frameworks that clarify accountability when humans rely on AI outputs \citep{BastaniCachon2025ContractingParadox,PoulidisEtAl2025ActionAttention}. Here, a reliable LLM attribution and detection method supports governance by enabling provenance checks and informing audit-design decisions, including the minimum text length and black-box sampling effort required, as well as how many sanctioned and non-sanctioned LLMs can be included to maintain the specified error guarantees.

\noindent\textbf{Learning-based methods.} Learning-based detectors train supervised classifiers on labeled human- and LLM-generated text. Existing approaches fine-tune transformer models \citep{hu2023radar}, use embeddings, loss-based measures, or similarity-based training \citep{verma2023ghostbuster,guo2024detective,zhang2026regurgitative}, or rely on lexical and stylistic features such as n-grams \citep{fagni2021tweepfake}. Their main limitation is poor generalization: classifiers may learn generator- or domain-specific patterns, while evaluations restricted to a narrow set of models and domains may overstate their performance on unseen settings \citep{li2024mage}. Our tests are training-free and also cover multi-model attribution and sampling-only access, with both errors decaying exponentially in text length. 

%\citet{wu2026advancing} theoretically justify using longer, easier-to-classify texts to supervise harder detection tasks; our method requires no labeled training data and bounds the Type~I and Type~II errors of the final detection rule for finite-length texts.

\noindent\textbf{Watermarking.} Watermarking changes an LLM's token-selection rule to embed a detectable signal. \citet{kirchenbauer2023watermark} divide the vocabulary into green and red lists and make green tokens slightly more likely. \citet{liu2024adaptive} insert the signal only when several next tokens are plausible and use the preceding text to determine which tokens to favor. \citet{dathathri2024scalable} instead sample several candidate tokens and use a key-based rule to select among them. Because these methods require the provider to modify generation and the verifier to know the key or detection rule, they cannot be applied post hoc to unwatermarked text. This dependence may discourage adoption \citep{nature2024watermarking,brookings2024watermarking}; OpenAI reportedly withheld a ChatGPT watermark partly because it could \enquote{discourage users or drive them to competitors} \citep{verge2024openaiwatermark}. In this provider-dependent setting, \citet{li2025statistical,li2025likelihood} develop statistical tests to determine whether a passage contains a watermark embedded during generation. Their tests can distinguish watermarked LLM output from unwatermarked text, but only through the embedded signal and its associated key; they do not address attributing ordinary, unwatermarked text to a particular model. Other watermarking research examines whether a black-box API deploys a watermark and where watermarked segments occur within mixed-source documents \citep{gloaguen2025blackbox,zhao2025watermarkedsegments}. Our paper studies whether any text, including unwatermarked text, was generated by a specified model or by a model in a specified set. We provide explicit finite-$N$ bounds on both Type~I and Type~II errors and study both attribution between sanctioned and non-sanctioned model sets and (A)-versus-not-(A) detection, where the alternative may include human authors.

\noindent\textbf{Zero-shot methods.}
Zero-shot detectors evaluate a passage using statistical quantities such as likelihood, token rank, entropy, and perplexity \citep{solaiman2019release,gehrmann2019gltr,su2023detectllm,he2024mgtbench,vasilatos2023howkgpt,wang2023m4}. DetectGPT compares the probability assigned to an original passage with those of small, meaning-preserving revisions, whereas Fast-DetectGPT uses conditional next-token probabilities to reduce this computation \citep{mitchell2023detectgpt,bao2024fast}. Binoculars compares perplexity measures from two related language models \citep{hans2024spotting}. These methods are primarily empirical and do not provide finite-length bounds on both Type~I and Type~II errors. \cite{chakraborty2023possibilities} establish detectability from multiple text samples and derive sample-complexity bounds for a detector that assumes knowledge of the human and machine text distributions. \citet{chen2025online} study online detection from a sequence of independent passages generated by a common source, rather than attribution from one passage whose tokens are sequentially dependent. Their approach requires external human-written reference data. Our method addresses single-passage provenance: it provides explicit finite-$N$ bounds on both Type~I and Type~II errors and covers attribution between sanctioned and non-sanctioned model sets and A-versus-not-A detection. Subsequent to our paper, \citet{zhu2025reliably} proposed a multi-scaled conformal-prediction framework that uses human-written calibration texts to control the false-positive rate under an i.i.d.\ calibration-and-test assumption. Their framework does not provide Type~II error guarantees, model-attribution guarantees, or finite-sample error bounds that decrease with the length N of a single audited text. In contrast, our method provides explicit finite-$N$ upper bounds on both Type~I and Type~II errors for model-specific detection and attribution without requiring human calibration data.

\section{Problem Setup and Notation}\label{sec:modelandbackground} 

In this section, we present our model, which formalizes the text-generation process, introduce the evaluator and generator models, define the statistical properties used in our detection method, and outline our hypotheses. Throughout the paper, boldface denotes vectors. The prompt is \(\mathbf{X}\) and the length-\(N\) text is \(\mathbf{Y}_N=[Y_1,\ldots,Y_N]\). We use capital letters such as \(Y_n\) for random tokens and lower-case letters such as \(y_n\) for their realizations. For a positive integer \(N\), \([N]=\{1,2,\ldots,N\}\), and for any set \(S\), \(|S|\) denotes its cardinality. We write \(\mathbb{P}\{\cdot\}\) and \(\mathbb{E}[\cdot]\) for probability and expectation, and \(\mathbf{1}\{\cdot\}\) for the indicator function. For a model \(M\), \(P^{M}\{\mathbf{Y}_N\mid \mathbf{X}\}\) denotes the induced distribution over length-\(N\) strings and \(p_n^{M}\) denotes the conditional next-token pmf, so \(p_n^{M}\{y\}=P^{M}\{Y_n=y\mid \mathbf{Y}_{n-1},\mathbf{X}\}\). In the black-box setting, we denote by \(\hat p_n^{A}\), \(\hat l_A\), and \(\hat h_N(A,A)\) the corresponding empirical analogues of the white-box quantities \(p_n^{A}\), \(l_A\), and \(h_N(A,A)\); \(\log\) denotes the natural logarithm and \(\exp\) denotes its inverse. For asymptotic comparisons as \(N\to\infty\), \(f_N=O\{g_N\}\) means there exist constants \(C>0\) and \(N_0\) such that \(|f_N|\le C|g_N|\) for all \(N\ge N_0\), and \(f_N=o\{g_N\}\) means \(f_N/g_N\to 0\).

%\(m\) denotes the sampling budget per token position and \(\delta\in(0,1)\) denotes the sampling confidence parameter. 

\textbf{Text generation.}
We represent any piece of text $\textbf{Y}=[Y_1, Y_2, \dots, Y_N]$ as a sequence of tokens from a finite vocabulary set $Y_n \in \mathcal{X}$ with size $K:=|\mathcal{X}|$\footnote{This representation does not require all models to use the same tokenizer. 
To establish that our theoretical results extend to models with different tokenizers, for each pairwise comparison between models $A$ and $B$, we fix a common deterministic tokenizer $\mathsf T$ (e.g., the evaluator model's tokenizer). We only need to assume that, under $\mathsf{T}$, each raw-text sequence admits a unique tokenization path for every model being compared, so that the corresponding induced conditional next-token probabilities are uniquely defined and computable.
If $P_A$ and $P_B$ denote the distributions of the raw text generated by the two models, then $P_A\circ\mathsf{T}^{-1}$ and $P_B\circ\mathsf{T}^{-1}$ are the corresponding induced distributions on the same token space. We define $p_n^A(\cdot)$ and $p_n^B(\cdot)$ for that comparison as the conditional next-token distributions associated with these induced distributions. Hence, within each pairwise comparison, both models are evaluated on the same token sequence and have the same length $N$.}. Let $M$ be a generative model described by $\textbf{Y}=M(\textbf{X})$, where $\textbf{X}$ denotes the user prompt as a sequence of tokens and the output denoted by $\textbf{Y}$ consists of a sequence of tokens $\textbf{Y}=[Y_1, Y_2, \dots, Y_N, \dots]$. Following practical implementations of large language models (e.g., refer to \citealp{radford2018improving,brown2020gpt3,openai2023gpt4,you2024linear}), we consider an auto-regressive language model that generates tokens sequentially. In an auto-regressive model, each token is drawn conditional on the user prompt $\mathbf{X}$ and all previously generated tokens.
In particular, the model $M$ first draws a random value for the first token, say, $Y_1=y_1$ by sampling from the distribution $p^{M}(Y_1 | \textbf{X})$, and then for each token $n \in [2,N]$, the model sequentially determines a distribution for the token given prompt $\textbf{X}$ and all the randomly chosen values $y_1, y_2, \dots, y_{n-1}$. So, we define a sequence of probability distributions $p^{M}(\textbf{Y}_N | \textbf{X})$ over $\textbf{Y}_N \in \mathcal{X}^N$ where $\textbf{Y}_N= [Y_1, Y_2, \dots, Y_N]$ is a substring of $\textbf{Y}$ consisting of the first $N$ tokens. The sequence of probability distributions is determined as \vspace{-4mm}
\begin{equation}\label{eq: bayes}  
\scalebox{0.9}{$
P^{M}(\textbf{Y}_N | \textbf{X}) = \prod_{n=1}^{N} p^{M}_{n}(\textbf{Y}_n), \quad \text{where} \quad  
p^{M}_n(\textbf{Y}_n) = P^{M}(Y_n | Y_1, Y_2, \dots, Y_{n-1}, \textbf{X})
$}
\end{equation}
%For each attribution problem, we fix a common deterministic tokenizer $\mathsf{T}$ for all models being compared (e.g., the tokenizer of a designated evaluator model).

\begin{remark}\label{bayesruleapplication}
    Note that Equation (\ref{eq: bayes}) is an application of the Bayes' rule and holds for any generative model regardless of whether tokens $Y_n$ are sequentially generated. While Equation (\ref{eq: bayes}) holds for all generative models, because conditional distributions $p^{M}_n(\textbf{Y}_n)$ are in general not easily accessible, we apply the rule for sequential models. 
\end{remark}

We present Example \ref{illustrativeexampleformodel} to illustrate how conditional probability distributions with dependence on the prompt and previously generated tokens determine the next-token probability distribution. Each token's probability is determined not in isolation but jointly with the prompt and the preceding tokens, meaning that even one word choice can reshape the distribution of the next
token. For simplicity in the example's exposition, we assume each word represents a token. \smallskip

%\todo[inline]{\color{red}I chose the following example to be the same as our running example (the one we presented in the introduction). Please let me know if you want me to practical-lengthen it or change it.}

\begin{example} \label{illustrativeexampleformodel}
    Consider the prompt \vspace{-3mm}
    \[
    \mathbf{X} = \texttt{Complete the sentence: The sun \dots}
   \vspace{-3mm} \]
    Model $M$ assigns probabilities to the possible next tokens conditional on the prompt $\mathbf{X}$ that includes the previously generated tokens $y_1=\texttt{The}$ and $y_2=\texttt{sun}$. For example, \vspace{-3mm}
    \begin{eqnarray*} 
    &P_3^{M}\big(\text{is} \mid \text{The Sun}\big)& = 0.40, \quad
    P_3^{M}\big(\text{was} \mid \text{The Sun}\big) = 0.25, \quad
    P_3^{M}\big(\text{also} \mid \text{The Sun}\big) = 0.20, \quad\\
    &P_3^{M}\big(\text{shines} \mid \text{The Sun}\big)& = 0.10, \quad
    P_3^{M}\big(\text{beautiful} \mid \text{The Sun}\big) = 0.03, \quad
    P_3^{M}\big(\text{others} \mid \text{The Sun}\big) = 0.02. \vspace{-4mm}
\end{eqnarray*} 
Consider two scenarios for the next token generation: For the first scenario, suppose the model selects the token $y_3=$ \texttt{also}. The updated prefix becomes \texttt{The Sun also}, and the model recomputes the conditional probability distribution for the next token. Because of the linguistic and cultural association with Hemingway's \texttt{The Sun Also Rises}, the probability mass now concentrates on rises, which yields
    \begin{eqnarray*}
    &P_4^{M}(\text{rises} \mid \text{The Sun also})& = 0.88, \quad
    P_4^{M}(\text{sets} \mid \text{The Sun also}) = 0.04, \\\quad
    &P_4^{M}(\text{appears} \mid \text{The Sun also})& = 0.03, \quad
    P_4^{M}(\text{burns} \mid \text{The Sun also}) = 0.02, \\\quad
    &P_4^{M}(\text{others} \mid \text{The Sun also})& = 0.03.
    \end{eqnarray*} 
   For the second scenario, suppose that the model chooses $y_3=$ \texttt{is} instead. The updated prefix becomes \texttt{The Sun is}, and the resulting conditional distribution changes to \vspace{-3mm}
\begin{eqnarray*}
    &P_4^{M}(\text{shining} \mid \text{The Sun is})& = 0.45, \quad
    P_4^{M}(\text{bright}\mid \text{The Sun is}) = 0.30, \\\quad
    &P_4^{M}(\text{hot}\mid \text{The Sun is})& = 0.15, \quad
    P_4^{M}(\text{rises}\mid \text{The Sun is}) = 0.001, \\\quad
    &P_4^{M}(\text{others}\mid \text{The Sun is})& = 0.099.
\end{eqnarray*}
    The contrast between these two cases illustrates how dependence on the prompt and previously generated tokens determines the probability distributions for the next token. After \texttt{also}, the token \texttt{rises} becomes very likely, whereas after \texttt{is}, it nearly vanishes from the distribution, replaced by contextually appropriate alternatives such as \texttt{shining} or \texttt{bright}. 
\end{example}
\textbf{Generator and evaluator models.} Our detection problem involves two roles: Evaluator model $A$ is the language model used to assess whether a given text is consistent with having been generated by $A$. Generator model $B$ is the model that produces the text being evaluated. Model $B$ can be an external LLM, an in-house LLM, or any other process, e.g., a human, that can be represented as a sequential model\footnote{Recall from Remark \ref{bayesruleapplication} that an application of Bayes’ rule yields that any text generation process corresponds to some sequential model $B$ regardless of whether tokens are sequentially generated.}. Provided that $B\neq A$ has a minimum distance from $A$ (presented in Assumption \ref{assum:mindiff}), our test does not require any information about $B$.

%token probabilities are not arbitrarily close to zero (Assumption \ref{assum:crossmodel}) and 

%Our test does not require any information about $B$.

\noindent\textbf{Statistical properties.} Our method relies on the convergence of two statistical properties for a text: log-perplexity and the average cross-entropy. Consider an evaluator model $A$ and a text generator model $B$.

\textit{Log-perplexity:} For a realized finite-length text string $\textbf{y}_N = [y_1, y_2, \ldots, y_N]$ with respect to an evaluator model $A$, it is defined as the per-token inverse likelihood of the string. Formally, perplexity with respect to model $A$ is $p^{A}\big(\textbf{y}_N\mid \mathbf{X}\big) = \left( \prod_{n=1}^{N} p^{A}_n(y_n) \right)^{-\frac{1}{N}}$, and the log-perplexity is
\begin{equation}\label{eq:deflogperplexity}
l_A(\textbf{y}_N)= -\frac{1}{N} \sum_{n=1}^{N} \log(p^{A}_n(y_n))
\end{equation}
Log-perplexity measures how \emph{surprising} a string generated by model $B$ is to the evaluator model $A$. For example, suppose the evaluator model $A$ observes \text{``The Sun dances quietly,''} which is a text generated by an unknown model $B$. When $A$ processes the same prompt ``The sun", it assigns low conditional probabilities to ``dances" and ``quietly"; for instance, \vspace{-2mm}
\[
P_3^{A}(\text{``dances''} \mid \text{``The Sun''}) = 0.005, \quad
P_4^{A}(\text{``quietly''} \mid \text{``The Sun dances''}) = 0.007.
\]
Because these probabilities are small, their logarithms are large in magnitude, and the resulting log-perplexity of the text under $A$ is high. Intuitively, model $A$ finds the word choices made by $B$ unlikely according to $A$'s own learned distribution.

\textit{Average cross-entropy:} Suppose model $B$ generates a string $\mathbf{y}$. The average cross-entropy of model $B$ and the evaluator model $A$ over the sub-string $\mathbf{y}_N$ is
\begin{equation}\label{def:crossentropy}
h_N(B,A)(\textbf{y}_N) = -\frac{1}{N} \sum_{n=1}^{N} \sum_{y_n \in \mathcal{X}} p^{B}_n(y_n) \log(p^{A}_n(y_n)).
\end{equation}
At each token position, $n$, the inner sum in Equation~\eqref{def:crossentropy}
is the expected surprise under evaluator model $A$ of the next token
generated by model $B$, given the prompt and the text generated so far.
Thus, $h_N(B,A)(\mathbf{y}_N)$ averages this conditional expected surprise
along the realized text. If model $A$ is both the generator and evaluator of $\mathbf{y}_N$, then the average cross-entropy and the average entropy over the text $\mathbf{y}_N$ become equivalent.

%The quantity $h_N(B,A)(\mathbf{Y}_N)$ is equal to the expected log-perplexity $l_A(\mathbf{Y}_N)$ when the expectation is taken over all length-$N$ texts $\mathbf{Y}_N$ drawn from model $B$'s generative distribution. Equivalently, $h_N(B,A)=\mathbb{E}_{\mathbf{Y}_N\sim B}\!\left[l_A(\mathbf{Y}_N)\right],$ where the expectation is with respect to the distribution over all sequences $\mathbf{Y}_N = (Y_1,\ldots,Y_N)$ produced by model $B$. If model $A$ is $\mathbf{Y}_N$'s both generator and evaluator, then the average cross-entropy and the average entropy over the text $\mathbf{Y}_N$ become equivalent. 

\noindent\textbf{White-box vs.\ black-box access.} For our theory, we consider two settings: a white-box setting and a black-box setting.

\textit{White-box.} For simplicity of exposition, we first present the \emph{white-box} setting in which we assume full knowledge of the evaluator model's next-token probability distributions $\{p^A_n(\cdot)\}_{n=1}^N$. This assumption allows us to present the main ideas of our statistical tests without additional notation and technicalities related to estimation error. The theoretical results in Sections~\ref{sec:modelandbackground}--\ref{sec:stats} are developed under white-box access to $p^A_n(\cdot)$. White-box access is standard for open-weight LLMs, and also when we have access to the conditional probability distributions for closed-weight models through API commands, and this is considered in the majority of the computer science literature \citep{mitchell2023detectgpt, gehrmann2019gltr}.

\textit{Black-box.} In this setting, we form empirical estimates of the next-token probability distributions $p^A_n$ by sampling from $A$ given a context, that is, a prompt $\mathbf{X}$ and prefix $(Y_1, Y_2, \dots, Y_{n-1})$. These empirical quantities serve as substitutes for the true probability-based metrics in the white-box case. For each position $n$, we obtain $m$ independent and identically distributed (i.i.d.) samples $\hat{Y}^{(1)}_n,\ldots,\hat{Y}^{(m)}_n \sim p^A_n(\cdot),
$
and define the empirical next-token distribution as
$\hat{p}^A_n(y) := \frac{1}{m}\sum_{j=1}^{m}\mathbf{1}\{\hat{Y}^{(j)}_n = y\}.$ Using these empirical estimates, we construct the empirical log-perplexity
$\hat{l}_A(\mathbf{Y}_N)
= -\frac{1}{N}\sum_{n=1}^{N}\log \hat{p}^A_n(Y_n),$ and the empirical average entropy $\hat{h}_N(A,A)(\mathbf{Y}_N)
= -\frac{1}{N}\sum_{n=1}^{N}\sum_{y_n\in\mathcal{X}}\hat{p}^A_n(y_n)\log\hat{p}^A_n(y_n).$

\noindent\textbf{Hypotheses.} We present two hypothesis testing settings: attribution among multiple models and whether the evaluator model $A$ generated the text. For attribution, we compare the log-perplexities of the candidate models, and in the $A$ vs.\ not-$A$ setting, we use the convergence of log-perplexity to the average cross-entropy. 

%as argued in Example~\ref{AversusgreedyA}.

\textit{Attribution among multiple models.} In this setting, we test whether a text is generated by one of the models in $\mathcal{A}=\{A_1,\ldots,A_p\}$ (e.g., external or non-sanctioned models) or by a model in $\mathcal{B}=\{B_1,\ldots,B_q\}$ (e.g., in-house models). 
The null hypothesis $\mathbf{H}_0$ states that the text originates from some model in $\mathcal{B}$, while the alternative $\mathbf{H}_1$ posits that it comes from a model in $\mathcal{A}$. Type~I error occurs when the test incorrectly attributes a text from $\mathcal{B}$ to a model in $\mathcal{A}$, and Type~II error occurs when the test fails to recognize that a text actually originates from $\mathcal{A}$.

\textit{Model $A$ or not model $A$.}
In this setting, we test whether the evaluator model $A$ generated the text. The null hypothesis $\mathbf{H}_0$ assumes that the text is not produced by model $A$, i.e., it is generated by some unknown model $B$ or another source, and the alternative $\mathbf{H}_1$ states that the text originates from $A$. Type~I error occurs when the test incorrectly concludes that model $A$ generated a text, whereas Type~II error arises when it fails to detect a text actually written by $A$.

\section{Concentration inequalities}\label{sec:theoreticalresults} %\vspace{-2mm}

In this section, we formalize the idea of verifying whether the log-perplexity of the string evaluated by $A$ converges to the average cross-entropy of the string under $B$ and $A$. In particular, we need to define some random variables and formalize convergence in terms of random variables. For exposition, we present the random variables and concentration inequalities under the white-box assumption, and dedicate Appendix \ref{sec:blackbox} to our extended results for the black-box setting. 

\noindent\textbf{Random Variables.} We define the random variable \(Z_n^{A}:=-\log\bigl(p_n^{A}(Y_n)\bigr)\) and the conditionally centered random variable \(X_n^{A}:=Z_n^{A}-\mathbb{E}_{p_n^{B}}\bigl[Z_n^{A}\bigr]\). Because \(p_n^{B}\) is the conditional distribution of \(Y_n\) given the prompt and all preceding tokens, \(X_n^{A}\) is conditionally zero-mean; that is, \(\mathbb{E}_{B}\bigl[X_n^{A}\mid \mathbf{X},\mathbf{Y}_{n-1}\bigr]=0\). Additionally, we define a random variable $S_N^{A} := \sum_{i=1}^{N} X^{A}_i$. Note that we only define the random variable $X_n^{A}$ to obtain our concentration inequalities, and our statistical tests in Section \ref{sec:stats} do not require calculating or knowing $X_n^{A}$. The tests only require access to the evaluator model $A$ to compute $Z_n^{A}$ and $\mathbb{E}_{p^{A}_n}[Z_n^{A}]$. In the white-box setting, we have access to and can compute $Z_n^{A}$ and $\mathbb{E}_{p^{A}_n}[Z_n^{A}]$. Consider a string $\bf{Y}$ generated by model $B$, and we want to evaluate the string using model $A$. So, $\mathbb{E}_{p^{B}_n}[Z_n^{A}]$ is
\begin{eqnarray} \label{expectedB}
\mathbb{E}_{p^{B}_n}[Z_n^{A}] = -\sum_{y_n \in \mathcal{X}} p^{B}_n(y_n) \log(p^{A}_n(y_n)) = H(p^{B}_n, p^{A}_n).
\end{eqnarray}
where $H(p^{B}_n,p^{A}_n)$ is the cross-entropy between distributions $p^{B}_n(.)$ and $p^{A}_n(.)$. Note that $\mathbb{E}_{p_n^B}[Z_n^{A}]$ is not necessarily finite. 
From Equation~\eqref{expectedB}, \( \mathbb{E}_{p_n^B}[Z_n^{A}]\) is infinite if there exists \( y_n \in \mathcal{X} \) such that \( p_n^B(y_n) > 0 \) and \( p_n^A(y_n) = 0 \). If such a token \( y_n \) appears in \( \textbf{Y}_N \), then model \( A \) assigns zero probability to the sequence, and we can conclude with certainty (i.e., probability 1) that model \( A \) did not generate the string. Hence, the detection problem becomes trivial. Otherwise, i.e., $\textbf{Y}_N$ does not include any $y_n$ with $p^{A}_n(y_n)=0$, we calculate $\mathbb{E}_{p_n^B}[Z_n^{A}]$ by applying conditional expectation as
\[
\mathbb{E}_{p_n^B}[Z_n^{A} \mid p^{A}_n(y_n) > 0 , \forall y_n \in \textbf{Y}_N ] 
= \sum_{y_n \in \mathcal{X}} -p^{\tilde{B}}_n(y_n) \log(p^{A}_n(y_n)) 
= H(p^{\tilde{B}}_n, p^{A}_n)
,
\]
where for $p^{A}_n(y_n)=0$, we have $p^{\tilde{B}}_n(y_n)=0$ and for $p^{A}_n(y_n)>0$, we have $p^{\tilde{B}}_n(y_n)=p^{B}_n(y_n)/ \sum_{y_k: p^{A}_n(y_k)>0}p^{B}_n(y_k)$.

This conditioning does not remove any case in which the test could incorrectly conclude that \(A\) generated the text. If a token \(y_n\) with \(p_n^B(y_n)>0\) and \(p_n^A(y_n)=0\) appears, then \(A\) assigns zero probability to the observed string and is ruled out with certainty. Otherwise, conditional on observing only tokens to which \(A\) assigns positive probability, the conditional next-token distribution of \(B\) is exactly \(p_n^{\tilde B}\). Hence, the detection problem for model pairs $A$ and $B$ reduces to that of models $A$ and $\tilde{B}$.
The modified pair of distributions $A, \tilde{B}$
satisfies the condition that if $p^A_n(y_n) = 0$, then $p^{\tilde{B}}_n(y_n) = 0$ for all $y_n$, which yields $\mathbb{E}_{p_n^{\tilde{B}}}[Z_n^{A}]$ is finite. Without loss of generality, we exclude the trivial case and only focus on pairs of models for which $\mathbb{E}_{p_n^B}[Z_n^{A}]$ is finite.

\noindent\textbf{Concentration Inequalities.} We provide concentration inequalities to show that if the string is generated by a model $B$ and evaluated by a model $A$, then $\frac{1}{N}\sum_{n=1}^{N} Z_n^{A}$ concentrates around the average cross-entropy of the string under $B$ and $A$, $h_N(B,A)$, with a high probability. If the text is generated and evaluated by model $A$, then $\frac{1}{N}\sum_{n=1}^{N} Z_n^{A}$ concentrates around the average entropy of the string under $A$ with high probability. 
In Appendix \ref{app:concentrationinequalroadmap}, we provide the proofs for the results in this section. These concentration bounds are the backbone of the statistical tests we design in Section \ref{sec:stats}. To ensure the boundedness of our random variable $\mathbb{E}_{p^{B}_n}\big[Z_n^{A}\big]$, we make a parametric assumption on the probability laws $p^{A}_k(.)$ and $p^{B}_k(.)$ below. As we explain in Remark \ref{assumption1practical}, this assumption is consistent with practice. 
\begin{assumption}\label{assum:crossmodel}
    We assume that there exists $\epsilon>0$ such that $p^{A}_n(y_k), p^{B}_n(y_k) \notin (0,\epsilon)$.
\end{assumption}%\vspace{-1mm}
Assumption \ref{assum:crossmodel} ensures that the zero-mean random variable $X_n^{A}=Z_n^{A}-\mathbb{E}_{p_n^B}[Z_n^{A}]$ is well-defined by guaranteeing that  
\begin{eqnarray} \label{outcomeofassumption1}
\mathbb{E}_{p^{B}_n}\big[|Z_n^{A}|\big] = \mathbb{E}_{p^{B}_n}\big[Z_n^{A}\big] 
= \sum_{y_n \in \mathcal{X}} -p^{B}_n(y_n) \log(p^{A}_n(y_n)) 
\le -\log(\epsilon)
\end{eqnarray}
is finite\footnote{Note that the upper-bound $-\log(\epsilon)$ is attainable, e.g., when the tokens are equally likely, and each has probability $\epsilon$. Hence, to guarantee that $\mathbb{E}_{p^{B}_n}\big[|Z_n^{A}|\big]$ is bounded, we need to ensure that $-\log(\epsilon)$ is bounded.}.

\begin{remark}\label{assumption1practical}
    Assumption \ref{assum:crossmodel} implies that models $A$ and $B$ either do not associate any probability to a token $y \in \mathcal{X}$, or they assign a probability of at least $\epsilon$. Our theoretical results depend only on $\log(\epsilon)$. Thus, our theoretical bounds rely primarily on a constant shift in the logarithmic scale. In practice, computers allow for a limited range of representable numbers due to finite precision in floating-point arithmetic, which is aligned with Assumption \ref{assum:crossmodel}. Very small probabilities are either rounded to zero or set to a minimum threshold to maintain numerical stability in computations (\citealp{luo-etal-2023-came,luoCAMEgithub2023,ieee7542019,goldberg1991every}). Here, $\epsilon$ denotes the minimum non-zero conditional probability assigned to any token on the vocabulary set. Machine precision provides the worst-case lower bound for $\epsilon$. Appendix~\ref{app:clipped_log_scores} shows robustness of the results in Sections~\ref{sec:theoreticalresults}--\ref{sec:stats} to a setting that does not require Assumption~\ref{assum:crossmodel}. Appendix~\ref{app:clipped_log_scores} also provides numerical calculations illustrating the resulting error bounds and the corresponding text lengths required to attain the specified error guarantees.
    \end{remark}
    
%LLM training and fine-tuning code often introduce explicit minimum thresholds. For example, the official CAME optimizer (used in GPT-2/T5/BERT-Large experiments, and Llama fine-tuning) implementation uses a default $\epsilon=10^{-16}$, corresponding to $-\log_{10}(\epsilon)=16$ (\citealp{luo-etal-2023-came,luoCAMEgithub2023}).

%OpenAI notes that its open-weight gpt-oss models are released in MXFP4 and can also be run in bfloat16; in bfloat16 the smallest positive representable number is $2^{-133}\approx 9.2\times 10^{-41}$, so probabilities below this scale are effectively rounded to zero, giving $-\ln(\epsilon)\approx 92.19$ (equivalently $-\log_{10}(\epsilon)\approx 40.04$) (\citealp{openaiIntroducingGptOss2025,openaiRunGptOssTransformers2025,nvidiaCUDAProgrammingGuideFloatTypes2025}). 
Theorem \ref{thm:maindiff} presents our concentration inequality, which we employ to design our statistical tests.  

%\vspace{1mm}

%{\color{red} To be combined as one theorem with two parts.}

%\ambuj{yeah, we need to do a better job here -- both theorems use $X_n$ but it refers to different RVs in the two cases. When you merge the two, make sure there is no confusion}

\begin{theorem}\label{thm:maindiff}\small
(a) If model $A$ is the same as model $B$, the following bound holds with the constant $c_1=\frac{4\sqrt{2}}{\log (2)}$ independent of the evaluator model $A$ such that for any $t>0$ we have 
\[
\mathbb{P}\bigg(\frac{1}{N} \left| \sum_{n=1}^{N} X_n^{A} \right| \geq t \bigg) 
\le 2 \exp \left[ 
  -\frac{N t}{2c_1 \log(K)} 
  \min \left(1, \frac{t}{c_1 \log(K)} \right)  
\right].
\]
(b) If model $A$ is not model $B$ and Assumption \ref{assum:crossmodel} holds,
the following bound holds with the constant $c_2=\frac{4\sqrt{2}}{\log (2)}$ independent of models $A$ and $B$ such that for any $t>0$ we have
\[
\mathbb{P}\left(\frac{1}{N} \left| \sum_{n=1}^{N} X_n^{A} \right| \geq t \right)
\le 2 \exp\left[
  -\frac{N t}{-2c_2 \log(\epsilon)} 
  \min\left(1, \frac{t}{-c_2 \log(\epsilon)} \right)
\right].
\]

\end{theorem}

\proof{Proof.}
    For the proof of part (a), see Section \ref{sec:proofmainsame}; for the proof of part (b), see Section \ref{app:proofthmdiff}.
\endproof

\begin{remark}
Our theorem's Part~(a) generalizes Theorem~1.1 (\emph{Main result}) in
\citet{zhao2022optimal}. They consider independent, not necessarily identically distributed, categorical variables and establish a concentration bound that is uniform over their probability parameters. Our setting allows the conditional distribution at each position to depend on the entire preceding history. Moreover, our theorem's part~(a) preserves \cite{zhao2022optimal}'s high-dimensional dependence on $N$ and $K$: when $t\leq c_1\log K$, our bound becomes $2\exp\left\{-\frac{Nt^2}{c_1^2(\log K)^2}\right\},$ so convergence is governed by the same
$(\log K)^2/N$ scaling as in \citet{zhao2022optimal}. Because the independent setting is a special case of our history-dependent setting, \cite{zhao2022optimal}'s lower bound also implies that this dependence on $N$ and $K$ cannot be uniformly improved over the broader class considered here. This extension is not an immediate application of a standard bounded-increment martingale inequality. The random variable $-\log p_n^A(Y_n)$ is not uniformly bounded when conditional token probabilities can approach zero. An important step is
Lemma~\ref{keylemma}, which shows that, conditional on every realized history, the centered increment $X_n^A$ has sub-exponential norm bounded by a constant multiple of $\log K$, uniformly over the
conditional probability distribution. Iterating this history-uniform conditional MGF bound yields the martingale concentration inequality. When the conditional distributions are history-independent,
\citet{zhao2022optimal}'s setting is recovered. Thus, Part~(a) addresses the martingale extension explicitly identified in \cite{zhao2022optimal}'s conclusion. Our theorem's part~(b) additionally treats the cross-model case $A\neq B$, which is not covered by \cite{zhao2022optimal}'s theorem.
\end{remark}

The different logarithmic factors in parts (a) and (b) arise from different bounds on the sub-exponential norm of the martingale increments \(X_n^{A}\). When \(A=B\), a token assigned a very small probability by \(A\) is correspondingly unlikely to be generated, and Lemma~\ref{keylemma} bounds the sub-exponential norm of \(X_n^{A}\) by a constant multiple of \(\log K\). When \(A\neq B\), model \(B\) may generate a token to which the evaluator model \(A\) assigns a very small positive probability. Under Assumption~\ref{assum:crossmodel}, Lemma~\ref{keylemma2} bounds the sub-exponential norm of \(X_n^{A}\) by a constant multiple of \(-\log\epsilon\).

Note that $\frac{1}{N}\left|\sum_{n=1}^{N} X_n^{A}\right|$ equals the difference between the log-perplexity under $A$ and the corresponding average (cross-)entropy. Theorem \ref{thm:maindiff} - (a) states that if the string is generated by the same model as the evaluator model $A$, then the log-perplexity of the string under $A$ converges to the average entropy of the string under $A$, except with an exponentially small probability in string length $N$. Theorem \ref{thm:maindiff} - (b) states that if model $B$ generates the text, then, except with an exponentially small probability in string length $N$, the log-perplexity of the string under the evaluator model $A$ converges to the average cross-entropy of the string under $B$ and $A$. 

Theorem~\ref{thm:maindiff} provides a non-asymptotic upper bound on the probability that the difference between the log perplexity and the average cross entropy, \(\frac{1}{N}\sum_{n=1}^{N} X_n^{A}\), deviates from $0$ by at least the value of \(t\). This probability depends on four parameters: the text length \(N\), the chosen tolerance level \(t\), the alphabet size \(K\), and \(\epsilon\). Among these, the text length exerts the highest influence. With all other parameters fixed, the deviation probability decays exponentially in \(N\), implying that longer texts substantially reduce the probability that the sum \(\frac{1}{N}\sum_{n=1}^{N} X_n^{A}\) does not fall within a $t$-neighborhood of zero.

The tolerance parameter \(t\) also plays a critical role. When \(t\) is small, the exponent in the bound behaves quadratically in \(t\). That is, reducing \(t\) by half (requiring the two empirical averages to be twice as close) necessitates increasing the text length by a factor of four to preserve the same deviation probability. 

The dependence on $K$ and $\epsilon$ is less significant than the parameters $N$ and $t$. These quantities appear only through logarithmic factors in the exponent. Enlarging the alphabet from \(K\) to \(K^2\) doubles \(\log(K)\), which reduces the exponent by at most a factor of $1/2$. Similarly, increasing $\epsilon$ (for example, doubling the number of digits to represent the probability of the next token for each of the tokens in the alphabet) affects the exponent only through changes in \(-\log(\epsilon)\). Thus, while \(K\) and \(\epsilon\) influence the probability bound, their impact is substantially smaller than that of \(N\) and \(t\). Since \(c_1\) and \(c_2\) are numerical constants, the bounds can be evaluated numerically after substituting \(N\), \(t\), \(K\), and, in part~(b), \(\epsilon\).\smallskip

%\todo[inline]{\color{red}Below, I provide some intuition on how this Theorem is being used for defining our statistical tests. I would appreciate it if you could let me know what additional intuition I can provide here.}

Theorem \ref{thm:maindiff} shows that, when model $B$ generates the text $\mathbf{Y}_N$, its log-perplexity under evaluator model $A$, $\frac{1}{N}\sum_{n=1}^{N} Z_n^{A}$, concentrates around the average cross-entropy $h_N(B,A)(\mathbf{Y}_N)$. When $A=B$, the average cross-entropy becomes $h_N(A,A)(\mathbf{Y}_N)$, the average entropy of evaluator model $A$ over the string. Therefore, if $A$ generated the text, the absolute difference between its log-perplexity under $A$ and $h_N(A,A)(\mathbf{Y}_N)$ concentrates around zero. This result provides the basis for our statistical test: we calculate these two quantities and attribute the text to $A$ when their absolute difference is no greater than a specified threshold $t$. The next section formalizes our tests.

\section{Statistical test}\label{sec:stats} %\vspace{-3mm}
We design our statistical tests using the result in Theorem \ref{thm:maindiff} and then evaluate Type I (false positive) and Type II (false negative) errors. In Section \ref{sec:multi}, we design a composite statistical test that determines whether a text is generated by a model that belongs to a set of models $\mathcal{A}$ or belongs to a disjoint set of models $\mathcal{B}$. In Section \ref{sec:human}, we study the case where we do not have access to the log-probabilities for the generator model (e.g., text is written by an unknown LLM, a human, etc.) and design a composite test to identify whether the model $A$ generated the text or not. 

%In Appendix \ref{app:stattestroadmap}, we provide the proofs for the results in this section. 

%In Section \ref{sec:twomodels}, we design a simple statistical test that determines whether a text is generated by a model $A$ or another model $B$. Then, 

%In particular, we consider a finite length text with length $N$ generated by a model $M$.

%\vspace{-2mm}
\subsection{Attribution among multiple LLMs}\label{sec:multi} %\vspace{-2mm}
Given a string $\mathbf{Y}_N$, we design a statistical test to detect whether one of the models in $\mathcal{A}=\{A_1, \ldots, A_p\}$ (non-sanctioned) or one of the models in $\mathcal{B}=\{B_1, \ldots, B_q\}$ (in-house) generated the text. The null hypothesis $\mathbf{H}_0$ is that the text $\mathbf{Y}_N$ is generated by a model in $\mathcal{B}$. We first calculate the random variables $Z^{M}_n=:-\log\big(p^{M}_n(Y_n)\big)$, and sum over $Z^{M}_n=-\log\big(p^{M}_n(Y_n)\big)$ for all models $M \in \mathcal{A}\cup \mathcal{B}$. Our test rejects the null $\mathbf{H}_0$ if for some $A_i \in \mathcal{A}$, we have
\begin{eqnarray*}
\frac{1}{N} \sum_{n=1}^{N} Z^{A_i}_n < \frac{1}{N} \sum_{n=1}^{N} Z^{B_j}_n \qquad \forall B_j \in \mathcal{B}.
\end{eqnarray*}
Type I error occurs when the test incorrectly concludes that the text is generated by one of the models in $\mathcal{A}$, and Type II error occurs when the test fails to identify that the text is generated by one of the models in $\mathcal{A}$. Our theoretical bounds require that the two models do not impose the same probability distribution on the string. Formally, we introduce Assumption \ref{assum:mindiff2m} as a minimum distance between log-probabilities for the two models.

%Note that the results for Type II errors do not depend on any assumptions.

%and incorrectly concludes that it is generated by one of the models in $\mathcal{B}$
\begin{assumption}\label{assum:mindiff2m} (minimum distance, A-B).
    We assume that if the generative and evaluator models are different, for a small positive $\epsilon_1>0$, we have almost surely under the generating model,
\begin{eqnarray*}
    \frac1N \sum_{n=1}^{N} D_{KL}\big(p_n^{B}||p_n^{A}\big) \ge \epsilon_1, \quad  \frac1N \sum_{n=1}^{N} D_{KL}\big(p_n^{A}||p_n^{B}\big) \ge \epsilon_1,
\end{eqnarray*}
where $D_{KL}\big(p_n^B||p_n^A\big)$ is the token-wise Kullback-Leibler (KL) divergence between $p_n^A$ and $p_n^B$, defined as
\begin{eqnarray*}
    D_{KL}\big(p_n^B||p_n^A\big)=\sum_{y \in \mathcal{X}}p_n^B(y)\log \frac{p_n^B(y)}{p_n^A(y)}.
\end{eqnarray*}
\end{assumption} %\vspace{-2mm}
\begin{remark}\label{assumption2practical}
    Note that if the two distributions are identical, no test can distinguish them, and, therefore, some assumption about distributions is always needed for any test to work. Assumption \ref{assum:mindiff2m} ensures that the two models satisfy a minimum distance in terms of their KL divergence over the generated text.\footnote{Using KL divergence as a measure for comparing the learned distribution from a text generation model to the distribution of another text generation model or the distribution of human-written text is common, e.g., see \citealp{pillutla2021mauve} and the references within.} Note that KL divergence, by definition, is a non-negative value that demonstrates the distance between the two distributions over the next word for the two models. 
\end{remark} \smallskip

\noindent\textbf{Relationship with Theorem~\ref{thm:maindiff}.} To see how Theorem~\ref{thm:maindiff} implies this test, consider the simple case of one model $A\in\mathcal{A}$ and one model $B\in\mathcal{B}$, where the null hypothesis is that $B$ generated the text. In this attribution setting, we have access to both models and can therefore compute $\frac{1}{N}\sum_{n=1}^{N}Z_n^A$ and $\frac{1}{N}\sum_{n=1}^{N}Z_n^B$. If $B$ generates $\mathbf{Y}_N$, Theorem~\ref{thm:maindiff}(a) implies that $\frac{1}{N}\sum_{n=1}^{N}Z_n^B$ converges in probability to $h_N(B,B)(\mathbf{Y}_N)$, whereas Theorem~\ref{thm:maindiff}(b) implies that $\frac{1}{N}\sum_{n=1}^{N}Z_n^A$ converges in probability to $h_N(B,A)(\mathbf{Y}_N)$. Hence, $\frac{1}{N}\sum_{n=1}^{N}Z_n^A-\frac{1}{N}\sum_{n=1}^{N}Z_n^B$ converges in probability to $h_N(B,A)(\mathbf{Y}_N)-h_N(B,B)(\mathbf{Y}_N)$, which equals $\frac{1}{N}\sum_{n=1}^{N}D_{\mathrm{KL}}\!\left(p_n^B\|p_n^A\right)$ and is at least $\epsilon_1$ by Assumption~\ref{assum:mindiff2m}. A positive value of this difference, therefore, supports the null hypothesis that $B$ generated the text, whereas a negative value leads to rejection of the null. Finally, to derive Proposition~\ref{thm:statstestmulti} from Theorem~\ref{thm:maindiff}, we apply the theorem with $t=\epsilon_1/2$.
\smallskip

%Our results show that the Type I and Type II errors for our statistical test are approximately $\exp\big(O(-N\epsilon_1)\big)$, which indicates that even for small values of $\epsilon_1$, which can converge to zero with the text length \big(for example, $\epsilon_1=O(N^{-1/2})$\big), our test provides exponentially small Type I and Type II errors in the text length $N$. 

%\textbf{Relationship with Theorem 1.} Here we provide intuition on how theorem 1 implies this test for the simple case of one model A vurses one model B. note that in attribution applications such as... we have access to both models A and B. THerefore, we can calculate both random variables Za and Zb. Additionally, note under the (generator is a) taht Za by theorem 1 must converge to Hn(a,a) and zb converges to hn(b,a) so their difference should converge to kl(b,a). which is always a positive value. hence the positiveness of zb-za is a sign of .... additionally, note that we can substitute t by exactly a lower bound on the kl divergence as we elaborate above in assumption 2.

\begin{proposition}\label{thm:statstestmulti}
    \small If Assumptions \ref{assum:crossmodel} and \ref{assum:mindiff2m} hold, the Type I error for our test is upper bounded by 
\begin{eqnarray*}
2|\mathcal{A}| \exp \left[ -\frac{N (\epsilon_1/2)}{-c_2\log(\epsilon)} 
\min \left(1, \frac{\epsilon_1/2}{-c_2\log(\epsilon)} \right) \right] 
+ 2 \exp \left[ -\frac{N (\epsilon_1/2)}{c_1\log(K)} 
\min \left(1, \frac{\epsilon_1/2}{c_1\log(K)} \right) \right]
\end{eqnarray*}
and the Type II error for our test is upper bounded by
\begin{eqnarray*}
2|\mathcal{B}| \exp \left[ 
  -\frac{N (\epsilon_1/2)}{-c_2\log(\epsilon)} 
  \min \left(1, \frac{\epsilon_1/2}{-c_2\log(\epsilon)} \right) 
\right]
+ 2 \exp \left[ 
  -\frac{N \epsilon_1/2}{c_1\log(K)} 
  \min \left(1, \frac{\epsilon_1/2}{c_1\log(K)} \right) 
\right]
\end{eqnarray*}
 with constants $c_1$, $c_2$, and $\epsilon$ as introduced in Theorem \ref{thm:maindiff}.
\end{proposition} 
\proof{Proof.}
See Section \ref{app:testmulti}.
\endproof
\hfill
\Halmos

Proposition \ref{thm:statstestmulti} demonstrates that both Type I and Type II errors for our composite test decay exponentially in the text length \(N\). Compared to the single-model case, the bounds now have factors \(|\mathcal{A}|\) and \(|\mathcal{B}|\) because we apply a union bound over the finite candidate classes. Consequently, the error bounds scale linearly with class size, and maintaining a fixed target error requires increasing \(N\) by an additive \(\log|\mathcal{A}|\) or \(\log|\mathcal{B}|\) term in the exponent.

 The proposition also involves the separation parameter \(\epsilon_1\) (from Assumption \ref{assum:mindiff2m}), which quantifies the distance between the null and alternative targets. Larger \(\epsilon_1\) yields a larger exponent and therefore faster decay of both error probabilities with \(N\). When \(\epsilon_1\) is small, the hypotheses are inherently difficult to distinguish, and the required text length must grow to achieve a small error. Additionally, each error bound is a sum of two exponentials, one involving \(\log(K)\) and one involving \(-\log(\epsilon)\). These correspond to the two step-size scales in the underlying concentration argument, and the slower-decaying term dominates at moderate \(N\). The dependence on \(K\) and \(\epsilon\) remains logarithmic, so their effect is lower than those of \(N\) and the separation \(\epsilon_1\).

\noindent\textbf{Managerial implications.} A manager designing a model-attribution audit to satisfy prespecified Type~I and Type~II error requirements has two main design levers: (i) the minimum text length \(N\) required for each audited document and (ii) the numbers of models included in the non-sanctioned set \(\mathcal{A}\) and the sanctioned set \(\mathcal{B}\). The first lever determines how much textual evidence $N$ is needed for attribution. Proposition~\ref{thm:statstestmulti} shows that the upper bounds on both errors decrease exponentially with \(N\). The manager can therefore determine the minimum text length necessary to meet the required guarantees and translate this value into report or assignment requirements. For example, employees may be required to explain the assumptions and reasoning underlying a recommendation, while students may be asked to provide explanations, examples, or comparisons rather than only short answers. The second lever determines which models the audit includes. When defining the sanctioned set $\mathcal{B}$, the manager can balance the benefits of offering additional approved models against the desired Type~II error guarantee. Similarly, when defining the non-sanctioned set $\mathcal{A}$, she may prioritize the models of greatest concern and include additional models whenever the desired Type~I error guarantee remains satisfied. The effects of \(|\mathcal{A}|\) and \(|\mathcal{B}|\) are linear because each set size multiplies one component of the corresponding error bound. Alternatively, the manager can retain a larger model set and compensate by increasing \(N\): multiplying a set size by a given factor requires only an additive increase in text length proportional to the logarithm of that factor. Finally, \(K\) and \(\epsilon\) enter the exponential rates only through \(\log K\) and \(\log(1/\epsilon)\). Consequently, as new models adopt larger vocabularies or finer numerical precision, the text length required for the audit need not increase proportionally.

%If the manager requires a smaller Type~I error bound, when defining the non-sanctioned set $\mathcal{A}$, she may prioritize the models of greatest concern and include additional models whenever the desired Type I error guarantee remains satisfied. If she requires a smaller Type~II error bound, when defining the sanctioned set $\mathcal{B}$, she can balance the benefits of offering additional approved models against the desired Type II error guarantee. 

%holding the other parameters fixed, she can reduce the number of non-sanctioned models in \(\mathcal{A}\). 

%she can similarly reduce the number of sanctioned models in \(\mathcal{B}\). 

\subsection{Model $A$ or not model $A$?}\label{sec:human}
%\vspace{-2mm}

%Otherwise, our test accepts the null hypothesis. 
%(e.g., it is generated by another model $B$), and the alternative hypothesis $\mathbf{H}_1$ is that $\mathbf{Y}_N$ is generated by the evaluator model $A$

%Note that, unlike in \ref{sec:multi}, we do not know the log-probabilities for the alternative $\mathbf{H}_1$.

%\ambuj{I am very confused here. The null hyp here is that test is not generated by model $A$ (to which we have white-box access). The writing suggests that the alternative is that text is generated by $A$. Where is the ``human" in this? The test is detecting LLM A vs not LLM A. Isn't it misleading to call this LLM vs human? But on the other hand, the assumption and the proof make reference to a model $B$. Does $B$ correspond to the human? but as far i can see it could really just be any other data source that satisfies assumption 3 and we don't have white-box access to. Also, it then seems that the test is not testing $H_0$: $A$ did not generate the text vs $H_A$: $A$ did. Rather you know a priori that it was either $A$ or $B$ and so $H_0$ is $B$ generated the text and $H_A$ is $A$ generated the text.}
%\ambuj{we probably need a sentence to justify why we made these null and alternate choices and not the other way around. Usually, the null is the default/conservative choice from which you will shift only if enough evidence arrives. So if we accuse someone of having used LLM $A$, they are innocent until proven guilty. So the null should be they did not use LLM A which matches our choice} 

Given a string $\mathbf{Y}_N$, for arbitrary constants $t$, we design a statistical test to detect whether the evaluator model $A$ generated the text. Consistent with the convention in hypothesis testing, the null hypothesis reflects the default/conservative assumption. So, if we accuse someone of having used (a non-sanctioned) LLM $A$, they are innocent until proven guilty. Accordingly, the null $\mathbf{H}_0$ is that the text $\mathbf{Y}_N$ is not generated by the evaluator model $A$; it is generated by any model $B$ that we do not have white-box access to, which could be a human or some other language generation process\footnote{Recall from section \ref{sec:modelandbackground} that an application of the Bayes’ rule yields that any text generation process corresponds to some sequential model $B$ regardless of whether tokens are sequentially generated. Our test does not require any information about $B$.}. The alternative is that the text $\mathbf{Y}_N$ is generated by $A$. We first calculate the random variable $Z_n^{A}=:-\log\big(p^{A}_n(Y_n)\big)$ to then calculate $\frac{1}{N}\sum_{n=1}^{N} Z_n^{A}$. Our test rejects the null if 
\begin{eqnarray*}
   \left| \frac{1}{N} \sum_{n=1}^{N} Z_n^{A} - h_N(A,A)(\mathbf{Y}_N) \right| \le t.
\end{eqnarray*}

Type I error occurs when the test mistakenly concludes that the text is generated by the evaluator model $A$, and Type II error happens when the test incorrectly concludes that the text is not written by $A$. To derive Type I error control, we impose Assumption \ref{assum:mindiff} as a sufficient condition on the distance between $p_n^A$ and $p_n^B$. The Type~II error bound does not require this condition. 

%Again, note that if the two distributions are identical, no test can distinguish them and, therefore, some assumption about distributions is always needed for any test to work. 

\begin{assumption}\label{assum:mindiff}
(minimum distance, A-NA). Our theoretical results require that $|h_N(B,A)(\mathbf{Y}_N)-h_N(A,A)(\mathbf{Y}_N)|>\epsilon_2,$ almost surely under $B$, for a small positive $\epsilon_2>0$.
\end{assumption}

\begin{remark}\label{assumption3practical}
   Assumption~\ref{assum:mindiff} is a separated-hypothesis (indifference-zone) condition used in similar settings in OM \citep{nelson1995using,frazier2014fully} and statistics \citep{romano2005optimal,balakrishnan2019hypothesis}: the Type~I guarantee in Proposition~\ref{thm:statstest} applies to non-$A$ sources satisfying this assumption, or the sufficient condition in Equation~\eqref{eq:sufficient}. Note that $h_N(A,A)$ is upper-bounded by $\log(K)$, although its value can be substantially smaller depending on the evaluator model and the context \citep{gehrmann2019gltr,holtzman2020curious}. A sufficient condition for obtaining an exponentially high-probability lower bound on this separation is the following \footnote{The detection experiments in Section~\ref{whiteboxnumerical}, where human text represents the null and LLM-generated text represents the alternative, provide empirical evidence consistent with Assumption \ref{assum:mindiff} being satisfied in the evaluated settings.}: If the generative and evaluator models are different, then for some positive $\epsilon_2>0$,
\begin{equation}\label{eq:sufficient}
\frac{1}{N} \sum_{n=3}^{N} \mathbb{E} \left[ D_{\text{KL}}\left(p_n^{B} \,\|\, p_n^{A} \right) \,\middle|\, \textbf{Y}_{n-2} \right] \ge h_N(A,A)(\textbf{Y}_N) + \epsilon_2.\footnote{In \eqref{eq:sufficient}, the conditioning on $\mathbf{Y}_{n-2}$ rather than $\mathbf{Y}_{n-1}$, and the start of the sum at $n=3$, arise from the proof of Lemma~\ref{lemma:bounding}: each term $D_{\text{KL}}\big(p_n^{B}\,\|\,p_n^{A}\big)$ is determined by the history $\mathbf{Y}_{n-1}$, so centering it at its conditional expectation given $\mathbf{Y}_{n-2}$ yields the zero-mean increments used in the martingale concentration argument, and beginning the sum at $n=3$ ensures that this conditioning history is nonempty. Because the omitted terms for $n=1,2$ are nonnegative, dropping them only makes the sufficient condition in \eqref{eq:sufficient} more conservative, and the guarantees in Proposition~\ref{thm:statstest} stand unaffected.}
\end{equation}
\end{remark}

%\ambuj{we should point out that some assumption is needed for any test to work. if two distributions are identical, no test can tell them apart}
\begin{proposition}\label{thm:statstest}
    For any $t \ge 0$, the Type II error for our test is upper bounded by
\begin{eqnarray*}
    \scalebox{0.95}{$
  2 \exp \left[ -\frac{N t}{c_1\log(K)} \min \left(1, \frac{t}{c_1\log(K)} \right) \right]
$}
\end{eqnarray*}
    with $c_1$ as introduced in Theorem \ref{thm:maindiff}. Suppose Assumption \ref{assum:crossmodel} holds and, in addition, Assumption \ref{assum:mindiff} \big(or the sufficient condition in (\ref{eq:sufficient})\big) holds, then for any positive $t < \epsilon_2/2$, and for $c_2$ as introduced in Theorem \ref{thm:maindiff}, the Type  I error of our test is upper bounded by  
\begin{eqnarray*} 2 \exp \left[ -\frac{N \epsilon_2/2}{-c_2\log(\epsilon)} \min \left(1, \frac{\epsilon_2/2}{-c_2\log(\epsilon)} \right) \right]
    + 2 \exp \left[ -\frac{N (\epsilon_2/2 - t)}{-c_2\log(\epsilon)} \min \left(1, \frac{\epsilon_2/2 - t}{-c_2\log(\epsilon)} \right) \right],
\end{eqnarray*}
\end{proposition}
\proof{Proof.}
See Section \ref{app:stattestthm}.
\endproof
\hfill
\Halmos

Proposition \ref{thm:statstest} demonstrates that the Type I and Type II errors of our test decrease exponentially in the text length $N$, and the exponential rate's dependence on $K$ and \(\epsilon\) is logarithmic. The parameter \(\epsilon_2\) in Assumption \ref{assum:mindiff} represents a guaranteed minimum separation between the value around which the test statistic concentrates when the text is generated by \(A\) and the value it concentrates around under any non-$A$ generator satisfying Assumption~\ref{assum:mindiff}
or the sufficient condition in Equation~\eqref{eq:sufficient}. Larger \(\epsilon_2\) therefore corresponds to a more distinguishable testing problem: a Type I error (false positive) would require the statistic, under a non-\(A\) source, to deviate sufficiently to cross this separation, an event whose probability is exponentially small in \(N\). This is reflected in the Type I bound through the margins \(\epsilon_2/2\) and \(\epsilon_2/2-t\); increasing \(\epsilon_2\) enlarges these margins and strengthens the exponential rate. The constraint \(t<\epsilon_2/2\) ensures that a strictly positive residual margin \(\epsilon_2/2-t\) remains; as $t$ approaches $\epsilon_2/2$ from below, this residual margin diminishes. In contrast, the Type II bound is driven by same-model concentration when the text is generated by \(A\), and hence depends on \(t\) and \(N\) but not on \(\epsilon_2\); increasing \(t\) reduces the probability of a Type II error (false negative). Note that our theoretical results also apply to a one-sided test that rejects the null hypothesis when $\frac{1}{N} \sum_{n=1}^{N} Z_n^{A} - h_N(A,A)(\mathbf{Y}_N) \le t$\footnote{If model $B$ generates the text, $\frac{1}{N} \sum_{n=1}^{N} Z_n^{A}$ converges to $h_n(B,A)$. Under Equation~\ref{eq:sufficient}, $h_n(B,A)>h_n(A,A)$, except with a probability that decreases exponentially in $N$, which yields that the theoretical guarantees in Proposition \ref{thm:statstest} can be extended to a one-sided test. For more details, please refer to the proof for Proposition \ref{thm:statstest} in Appendix \ref{app:stattestthm}.}.

\noindent\textbf{Managerial implications.} A manager designing an audit to determine whether model \(A\) generated a text has two main design levers: (i) the attribution tolerance \(t\), and (ii) the minimum text length \(N\) required for each audited document. In applying Proposition~\ref{thm:statstest}, the manager may use a conservative lower bound $\underline{\epsilon}_2$. A smaller value of $\underline{\epsilon}_2$ produces a more conservative guarantee, which the manager can offset by increasing the required text length $N$. The first lever determines how strict the attribution rule is. A smaller \(t\) requires the test statistic to be closer to zero before the text is attributed to \(A\). Decreasing \(t\) lowers the \(t\)-dependent component of the Type~I error bound, while leaving the other component unchanged, but increases the Type~II error bound. A smaller \(t\) may therefore be appropriate when an attribution can lead to disciplinary, regulatory, or other consequential action. A larger \(t\) may be appropriate when the audit is used primarily to identify cases for subsequent human review and avoiding missed use of model \(A\) is more important. The second lever determines how much text is needed for the audit. After selecting \(t\), the manager can invert the bounds to determine the minimum \(N\) needed to satisfy the organization's error requirements. This value can be translated into minimum report or assignment lengths---for example, by requiring employees to explain the assumptions and reasoning behind a recommendation or students to provide explanations and supporting examples. The upper bounds on both errors decrease at exponential rates in \(N\). Text length can also compensate for a stricter tolerance: in the quadratic region of the Type~II error bound, reducing \(t\) by a factor \(q\in(0,1)\) requires approximately \(1/q^2\) times as many tokens to preserve the same Type~II guarantee. Thus, halving \(t\) requires approximately four times as much text. Finally, \(K\) and \(\epsilon\) enter the bounds through \(\log K\) and \(\log(1/\epsilon)\). Consequently, as new models adopt larger vocabularies or finer numerical precision, the required audit text length need not increase proportionally.

In Section~\ref{sec:stats}, we established non-asymptotic upper bounds on the Type~I and Type~II errors of our proposed statistical tests. In Section \ref{sec:lowerbounds}, we show that there exist model pairs for which no statistical test can make both Type~I and Type~II errors decay faster than exponentially in the text length $N$.

\section{Lower Bounds}\label{sec:lowerbounds}
%via Information-Theoretic Inequalities

In Section~\ref{sec:theoreticalresults}, we establish (exponentially decaying in text length $N$) upper bounds on the Type~I and Type~II errors of our proposed statistical tests. We now show that there exist models $A$ and $B$ for which no statistical test can make both errors decay faster than exponentially in $N$. To address this, we consider the problem that, given a string $\mathbf{Y}_N$, we design a statistical test to detect whether model $A$ (non-sanctioned) or model $B$ (in-house) generated the text. The null hypothesis $\mathbf{H}_0$ is that the text $\mathbf{Y}_N$ is generated by model $B$. We consider the special case in which the probability laws $p^{A}_n(\mathbf{Y}_n)$ and $p^{B}_n(\mathbf{Y}_n)$ generate text via i.i.d.\ draws from fixed distributions $P_A(\cdot)$ and $P_B(\cdot)$, respectively. Accordingly, we want to test
$H_0:\, \mathbf{Y}_N \sim P_B^{\otimes N},
\quad
H_1:\, \mathbf{Y}_N \sim P_A^{\otimes N}.$

\begin{corollary}\label{cor:lecam-iid}
For any statistical test $T$ that takes a string $\mathbf{Y}_N$ of length $N$, let $\alpha_N(T)$ and $\beta_N(T)$ denote its Type~I and Type~II error probabilities, we have
\begin{equation}\label{eq:lecam-iid}
 \alpha_N(T) + \beta_N(T) \ge  \frac{1}{2} e^{-N \cdot D_{KL}(P_B || P_A)}.
\end{equation}
\end{corollary}
\proof{Proof.}
    See Appendix \ref{app:proofofcorr}.
\endproof
\hfill
\Halmos

Corollary \ref{cor:lecam-iid} establishes that there exist models $A$ and $B$ for which the sum of the Type~I and Type~II errors of any statistical test is bounded below by a quantity that decays exponentially in $N$. Consequently, no statistical test can make both Type~I and Type~II errors decay faster than exponentially in $N$: if both errors decayed faster than exponentially, then their sum would also decay faster than exponentially, contradicting the lower bound. This complements our upper bounds, which show that our proposed tests achieve exponentially decaying Type~I and Type~II errors. Thus, the upper and lower bounds agree on the dependence on text length: exponential decay is achievable, while faster-than-exponential decay of both errors cannot be guaranteed in general.

%No test can outperform the exponential decay in the sample size, and the achievable decay rate is governed by the KL divergence $D(P\|Q)$. In particular, even the best possible test cannot achieve $\alpha_n+\beta_n$ smaller than $\tfrac{1}{2} e^{-n D(P\|Q)}$ asymptotically. This complements our upper bounds, which showed that our test also achieves exponentially decaying errors. Hence, both upper and lower bounds confirm that exponential decay in text length is near-optimal.

%\proof{Proof.}
%See Section \ref{app:lowerboundproof}.
%\endproof
%\hfill
%\Halmos

%{\color{red} Do not refer to the equation in the proof - revise the discussion.}

\section{Experiments}\label{sec:experiments}

Having established theoretical bounds under worst-case settings, in this section we present three sets of experiments to understand the performance of our method in practical settings and the conditions that make attribution and detection harder. We examine performance across fact-based, news, and creative-writing contexts, with particular attention to short, fact-based, Wikipedia-like texts, where human and LLM writing can be relatively similar. We intentionally include models with different training focuses, including one trained mainly on stories and novels, to examine whether such training makes a model’s creative writing more similar to human writing and therefore harder to detect. We also study how performance changes with text length and identify practical text lengths at which additional text provides little further improvement. Finally, we examine a no-access setting in which we cannot observe the evaluator model’s conditional next-token probabilities or generate samples from it, so another accessible model is used to score the text. In this setting, we study how performance changes when text is generated in different ways and when it is changed after generation. Our experimental designs follow settings widely used in the empirical LLM-detection literature (e.g., \citealt{mitchell2023detectgpt,dugan2024raid}). Subsection \ref{exp:attribution} examines performance in an LLM attribution setting with multiple candidate models. Subsection \ref{whiteboxnumerical} examines performance in an LLM detection setting where human text represents the null hypothesis (i.e., not LLM $A$). Subsection \ref{blackboxnumerical}, using a dataset with approximately 6M texts, examines performance in a no-access setting, where a surrogate model is used for scoring, including when an adversary manipulates the text in ways that affect both Type I and Type II errors. For sample-size experiments in the black-box-with-sampling setting, see Appendix~\ref{BBexperiments}.

%Having established theoretical bounds under worst-case settings, in this section we present three sets of experiments to examine the performance of our tests in practical settings. Among other parameters, we examine the impacts of text contexts, whether the generator and evaluator models belong to the same family, and text length on performance. 

%In this section, we present three sets of experiments: Subsection \ref{exp:attribution} empirically evaluates Proposition \ref{thm:statstestmulti}. Subsection \ref{whiteboxnumerical} empirically evaluates Proposition \ref{thm:statstest} where we use human text to represent the null hypothesis (i.e., not LLM $A$). As part of this subsection, \ref{app:length experiments} examines how performance varies with text length. Subsection \ref{blackboxnumerical}, using a dataset with approximately 6M texts, empirically examines performance when an adversary manipulates the text that affects both Type I and Type II errors. For sample-size experiments in the black-box-with-sampling setting, see Appendix~\ref{BBexperiments}.

\subsection{LLM Attribution Experiments}\label{exp:attribution} %\vspace{-2.5mm}

We examine attribution performance across two text contexts and four language models, including the more challenging setting in which the generator and evaluator models belong to the same family. We use two datasets: to represent fact-based information, we use the SQuAD context dataset \citep{rajpurkar2016squad}, which provides context paragraphs drawn from Wikipedia articles, and, to represent a dataset combining human-written creative writing and fact-based information, we use the news articles in the XSum dataset \citep{narayan2018don}. For details about datasets, refer to Appendix \ref{app:datasetdetails}. To create an LLM version of the articles in the SQuAD and XSum datasets, following the literature (e.g., \citealt{mitchell2023detectgpt}), we first randomly sample 500 texts from each dataset. Then, following \cite{mitchell2023detectgpt}, for each sample, we prompt the first 30 tokens of the selected text to each LLM and ask it to generate a 300-token LLM version. 

The language models we use to generate 300-token texts are Llama2 (7B), Llama3 (8B), Falcon (7B), and Qwen (32B). We select these language models to cover architectural diversity and a range of capacities. Llama2 (7B) and Llama3 (8B) are from the same family, creating a challenging attribution setting in which closely related models must be distinguished. Falcon (7B) provides an architecturally distinct model that has also been used as a scoring model to approximate other LLMs in prior work, e.g., in \cite{hans2024spotting} and \cite{zhu2025reliably}. Qwen (32B) serves as a larger, stronger model, allowing us to test performance with a higher-capacity generator. Because the attribution procedure here requires access to the candidate models' conditional next-token probabilities, we choose models for which we have white-box access. 

Table~\ref{tab:error_stats} demonstrates the results of our attribution experiments. In the experiments for each LLM, the evaluator LLM is the set $\mathcal{A}$, and the other three LLMs represent the set $\mathcal{B}$. In Table \ref{tab:error_stats}, the LLM in the headline of each column stands for the evaluator LLM $\mathcal{A}$. The null hypothesis is that the text is generated by one of the models in $\mathcal{B}$. A Type I error occurs when the test incorrectly concludes that the evaluator model generated the text, and a Type II error occurs when the test fails to detect that the evaluator model generated the text. Accordingly, for a given test, the False Positive Rate (FPR) equals the Type I error rate, and the False Negative Rate (FNR) equals the Type II error rate. The key result is that attribution is strong across distinct model families and for the larger Qwen (32B) model, with errors concentrated in the more challenging same-family setting. Across both datasets, FPR never exceeds 2\%, and FNR is at most 4\% in six of the eight model--context combinations. The only larger FNRs occur when Llama3 is the evaluator—14\% on XSum and 10\% on SQuAD—whereas Llama2 has no false negatives in either context. This asymmetric same-family pattern suggests that the difficulty of distinguishing related models can depend on which model serves as the evaluator. The similar results across SQuAD and XSum, together with Qwen (32B)'s zero FNR in both contexts, suggest that within-family model similarity is a more important source of difficulty in this experiment than text context or model scale.

%The results exhibit a comparable error pattern across SQuAD and XSum datasets, suggesting that dataset-specific artifacts do not primarily drive performance. False positives remain low, consistent with the test's conservative approach to attributing text to the evaluator model. False-negative rates are higher when the evaluator and a competing model belong to the same model family, as in the Llama2--Llama3 comparison. This pattern is consistent with closely related models assigning similar log-perplexities to the same text, thereby reducing the separation used by the attribution rule and making the generating model more difficult to identify. In this condition, the likelihood signal separating the evaluator from the other candidates is weaker, so the test more often fails to confidently reject the null. Our results for Qwen demonstrate that our test remains reliable for larger LLMs. 
\vspace{-1mm}
\small
\begin{table}[H]
\centering
\scriptsize
\renewcommand{\arraystretch}{1.3}
\begin{tabular}{llcccccccc}
\toprule
\multicolumn{2}{c}{} & \multicolumn{4}{c}{\textbf{XSum}} & \multicolumn{4}{c}{\textbf{SQuAD}} \\
\cmidrule(lr){3-6} \cmidrule(lr){7-10}
\textbf{Error Type} & \textbf{} & \textbf{Llama2} & \textbf{Llama3} & \textbf{Falcon} & \textbf{Qwen (32B)} & \textbf{Llama2} & \textbf{Llama3} & \textbf{Falcon} & \textbf{Qwen (32B)} \\
\midrule
False Positive Rate  &       & 0.02 & 0.00 & 0.02 & 0.00 & 0.02 & 0.00 & 0.00 & 0.02 \\
False Negative Rate &       & 0.00 & 0.14 & 0.00 & 0.00 & 0.00 & 0.10 & 0.04 & 0.00 \\
\bottomrule
\end{tabular}%\vspace{2mm}
\caption{\small Type I (false-positive) and Type II (false-negative) error rates for LLM attribution on XSum and SQuAD datasets.}
\label{tab:error_stats}
\end{table}
\normalsize
\vspace{-7mm}
%Consistent with intuition, more distinct models are more separable, leading to fewer false negatives while maintaining low false positives across both datasets.
\subsection{Detection Experiments} \label{whiteboxnumerical} 

We examine detection performance across three text contexts and three language models, including a deliberately challenging setting involving GPT-NeoX Erebus (20B), which was trained on story- and novel-based corpora.

%\ambuj{not enough detail below for me to understand how the experiments were setup. suddenly we are talking about positive and negative examples? what does that mean? which test are you actually running? what's the null? what's the alternative? I know you have limited space but you could give full details in appendix}

%\textbf{Experiment setup.} In this section, we design experiments to numerically validate the performance of our proposed statistical test from Proposition
\noindent\textbf{Setup \& Datasets.} To represent the null hypothesis, we use human text (i.e., text not generated by LLM $A$). For human text, in addition to the XSum and SQuAD datasets, we use prompted stories from the Reddit WritingPrompts \citep{fan2018hierarchical} to represent human creative writing. For details on the datasets, see Appendix \ref{app:datasetdetails}. 

To create LLM versions of the human texts in the datasets, following \cite{mitchell2023detectgpt}, we first randomly sample 500 human-written texts from each dataset and prompt the first 30 tokens of each text to the LLM to create a $300$-token LLM-generated version. To compute the detection score, we require access to the evaluator model's conditional next-token probabilities and therefore choose models for which we have white-box access. Specifically, we examine Llama3 (8B), GPT-NeoX Erebus (20B), and Qwen (32B). We intentionally include GPT-NeoX Erebus (20B), which was trained on story- and novel-based corpora, to evaluate our method in a challenging setting where specialization in narrative writing may reduce the distinction between model- and human-written text. Llama3 (8B) is a relatively compact model, while Qwen (32B) is a considerably larger, higher-capacity model, allowing us to examine performance across different model specializations and capacities\footnote{Unlike the attribution setting, where we deliberately chose two models from the same family (Llama2/Llama3) to create challenging same-family vs.\ cross-family attribution scenarios, here we treat each model separately as the focal “$A$” in an $A$-vs-not-$A$ test and therefore prioritize diversity in architecture and scale rather than fine-grained family comparisons.}.

%Llama3 (8B) is a relatively compact model, GPT-NeoX Erebus (20B) (20B) is a larger model with a different training pipeline and stylistic behavior, and Qwen (32B) is a considerably larger, higher-capacity model, allowing us to probe detection performance as model size and capacity increase 

%{\color{red} A few sentences on why we chose these four and why they represent many other models?+ why these are different from the previous experiment?} 
%We have white-box access to all models, which aligns with the requirements of Proposition \ref{thm:statstest}. 
%that the evaluator model does not generate. Hence, the null hypothesis is that the text is generated by a human (not by the evaluator model), and the alternative is that the evaluator model generated the text. 
%%The false positive rate (Type I error) refers to instances where the test incorrectly identifies human-written text as being generated by the evaluator model. Type II error occurs when the test fails to detect that the evaluator model actually generated a text and mistakenly classifies it as human-written.

For this section, since the first 30 tokens are generated by a human for both the null and the alternative hypotheses, we use a one-sided version of our test that rejects the null hypothesis when $\frac{1}{N} \sum_{n=1}^{N} X_n^{A} =\frac{1}{N} \sum_{n=1}^{N} Z_n^{A} - h_N(A,A)(\mathbf{Y}_N) \le t$ to remove the impact of the differences between cross-entropy and log perplexity caused by the initial 30 tokens\footnote{Since the first 30 tokens in both the human and the LLM samples are the same, these 30 tokens yield the same average entropy and log-perplexity under both null and alternative. Denote the random variable $C = \sum_{n=1}^{30} X_n$. Since the first 30 tokens are human-written, the random variable $C$ is not zero-mean, and for both null and alternative, $\sum_{n=1}^{300} X_n$ is biased by the non-zero mean random variable $C$. In this setting, it is inaccurate to use $|\sum_{n=1}^{300} X_n|$ being small to determine null. One approach is to exclude the first 30 tokens and instead use $\left|\sum_{n=31}^{300} X_n\right|$. Another approach is to use the one-sided version of our test that rejects the null hypothesis when $\frac{1}{N} \sum_{n=1}^{N} X_n^{A} =\frac{1}{N} \sum_{n=1}^{N} Z_n^{A} - h_N(A,A)(\mathbf{Y}_N) \le t.$ Since the $\sum_{n=1}^{300} X_n$ under null and alternative is shifted by a constant $C$, the test's performance is not affected. That is why we can use the one-sided version of our test to remove the impact of the initial 30 tokens.}. 
%\footnote{Since the first 30 tokens in both the human and the LLM samples are the same, these 30 tokens yield the same average entropy and log-perplexity under both null and alternative. Denote the random variable $C = \sum_{i=1}^{30} X_n$. Since the first 30 tokens are human-written, the random variable $C$ is not zero-mean, and for both null and alternative, $\sum_{n=1}^{300} X_n$ is biased by the non-zero mean random variable $C$. In this setting, it is inaccurate to use $|\sum_{n=1}^{300} X_n|$ being small to determine null. We instead use the one-sided version of our test that rejects the null hypothesis when $\frac{1}{N} \sum_{n=1}^{N} X_n^{A} =\frac{1}{N} \sum_{n=1}^{N} Z_n^{A} - h_N(A,A)(\mathbf{Y}_N) \le t.$ Since the $\sum_{i=1}^{300} X_i$ under null and alternative is shifted by a constant $C$, the test's performance is not affected. That is why we use the one-sided version of our test to remove the impact of the initial 30 tokens. 

%use the one-sided version of our test
%{\color{red}For a detailed explanation, please refer to Appendix \ref{app:thirtytokens} - move from appendix to here!}
%\subsection{The initial thirty human tokens' impacts}\label{app:thirtytokens}
%Recall that for our white-box experiments in Section \ref{whiteboxnumerical}, we randomly sample 500 texts from each dataset, and prompt the first 30 tokens of each text to the LLM to create an LLM-generated version.
Here, True Positive Rate (TPR) is the rate of correctly spotting that the evaluator model generated the text, and False Positive Rate (FPR) is the rate of incorrectly flagging human text as generated by the evaluator model. The detection rule uses a fixed threshold $t$, and varying this threshold affects both Type I and Type II errors.

\textit{Setting the threshold $t$.} Since different detectors use distinct methodologies and require their own thresholds, it is essential to account for the impact of these thresholds when evaluating the test performance. To address this, the literature typically adopts two approaches: (1) evaluating the TPR at a specified low FPR, e.g., \citep{hans2024spotting,{dugan2024raid}}, or (2) using the Area Under the Receiver Operating Characteristic (AUROC) curve as a summary measure of performance, e.g., \citep{verma2023ghostbuster,mitchell2023detectgpt}. Accordingly, we use two metrics. Our first metric, following \citealt{hans2024spotting}, is TPR when FPR is held less than or equal to 1$\%$. To characterize how methods behave across the full range of thresholds, we also need to report AUROC, which plots TPR against FPR at all possible thresholds. So, our second metric is AUROC.

We compare the performance of our method with established zero-shot baselines. We adopt the experimental setting from \cite{mitchell2023detectgpt} to ensure our results are directly comparable under a consistent experimental setup. In this setting, the (log-)likelihood-based benchmarks use the source model’s average token-wise log-probabilities; the (log-)rank-based benchmarks use the average observed (log-)rank of the tokens in the passage based on the model’s conditional distributions; and the entropy-based benchmarks hypothesize that the model's text is less surprising to the model, leading to lower-entropy predictive distributions. We additionally compare against Binoculars \citep{hans2024spotting}, which discusses its performance in the low-false-positive regime. \cite{hans2024spotting} provides extensive comparisons to a range of recent zero-shot detectors, including Fast-DetectGPT \citep{bao2024fast} and DNA-GPT \citep{yang2024dna}, and reports stronger low-FPR detection performance than these alternatives. Therefore, we benchmark against \cite{hans2024spotting} to provide a conservative reference point.

\noindent\textbf{TPR at FPR $\leq 1\%$.} Table~\ref{tab:tpr} reports detection performance at FPR $\leq 1\%$. Our method achieves the highest (or is tied-highest) TPR in seven of the nine model--dataset settings. The largest improvements relative to the strongest competing method occur on the fact-heavy SQuAD dataset, particularly for Qwen (32B). This setting is challenging because the constrained factual context can make human and LLM continuations similarly probable under the evaluator, weakening methods that rely on likelihood, rank, or entropy in isolation. By comparing log-perplexity with the evaluator model's average entropy, our method retains greater separation between human- and LLM-generated text in these difficult settings. Binoculars is a strong zero-shot competitor in several XSum and WritingPrompts settings, where both methods often attain high performance. On SQuAD, however, our method matches or exceeds Binoculars considerably for every evaluator model, with the largest difference occurring for Qwen (32B). This pattern is consistent with Binoculars' cross-model normalization being less informative when factual constraints make the models' predictions similar, whereas our method compares the observed log-perplexity with the evaluator's own average entropy. For a more detailed comparison with Binoculars, see Appendix~\ref{app:detectordetails}.

% Relying on the convergence of two statistical properties, our approach is less sensitive to this calibration issue. 

\begin{table}[H]
    \centering
    \resizebox{\textwidth}{!}{ % Resizes the table to fit within the page
    \begin{tabular}{lccc|ccc|ccc}
        \toprule
        & \multicolumn{3}{c|}{XSum} & \multicolumn{3}{c|}{SQuAD} & \multicolumn{3}{c}{WritingPrompts} \\
        \cmidrule(lr){2-4} \cmidrule(lr){5-7} \cmidrule(lr){8-10}
        Method & Llama3 (8B) & GPT-NeoX Erebus (20B) & Qwen (32B) & Llama3 (8B) & GPT-NeoX Erebus (20B) & Qwen (32B) & Llama3 (8B) & GPT-NeoX Erebus (20B) & Qwen (32B) \\
        \midrule
        $\log p(x)$ & 0.89 & 0.02 & 0.99* & 0.01 & 0.05 & 0.01 & \textbf{1.00} & 0.46 & \textbf{1.00} \\
        Rank & 0.22 & 0.06 & 0.20 & 0.07 & 0.18 & 0.01 & 0.49 & 0.39 & 0.43 \\
        LogRank & \textbf{0.99} & 0.05 & 0.99* & 0.02 & 0.06 & 0.01 & \textbf{1.00} & 0.54& \textbf{1.00}\\
        Entropy & 0.02 & 0.09 & 0.01 & 0.01 & 0.04 & 0.00 & 0.00 & 0.00 & 0.00 \\
        Binoculars & \textbf{0.99} & 0.89* & 0.99*& \textbf{0.96} & 0.86* & 0.26* & 0.97* & \textbf{0.99} & 0.99 \\
        \textbf{Ours} & 0.97* & \textbf{1.00} & \textbf{1.00} & \textbf{0.96} & \textbf{0.92} & \textbf{0.72} & 0.90 & \textbf{0.99} & \textbf{1.00} \\
        \midrule
        Diff (zero-shot) & -0.02 & 0.11 & 0.01 & 0.00 & 0.06 & 0.46 & -0.10 & 0.00 & 0.00 \\
        \bottomrule
    \end{tabular}
    } %\vspace{2mm}
    \caption{\footnotesize TPR at FPR $\leq 1\%$. The best TPR values are in \textbf{bold}, the second-best values are marked with an asterisk (*). The row Diff(zero-shot) shows our TPR improvement over the strongest zero-shot methods.
}
    \label{tab:tpr}
\end{table}
\vspace{-5mm}

We conduct two additional analyses to examine dimensions of performance not captured by TPR at a fixed low FPR: AUROC evaluates performance across the full range of decision thresholds, and the text-length experiments examine how detection performance changes as more textual content becomes available, how much text is needed for reliable detection, and whether some text contexts require longer passages.

\noindent\textbf{AUROC.} For the AUROC experiments, model samples are generated by sampling from the conditional distribution with $t$/temperature $=1$\footnote{Following common practice, we sample with temperature $T{=}1$, i.e., from the model's unmodified conditional distribution (no logit rescaling before the softmax). For more discussion on the effects of temperature on token conditional probability distribution, see, e.g. \cite{hinton2015distilling} and \cite{holtzman2020curious}. Temperature rescales logits as $\mathrm{softmax}(u/T)$: lower temperatures ($T<1$) concentrate probability mass on the most probable tokens, making generations less diverse, while higher temperatures ($T>1$) produce a softer/flatter distribution that increases randomness by giving relatively more weight to lower-probability tokens.}. Similar to the results in Table~\ref{tab:tpr}, the results in Table~\ref{tab:auroc} demonstrate that our method's improvement is most pronounced on fact-based datasets such as SQuAD. Additionally, the advantage of our method becomes more visible as the generator becomes larger, e.g., Qwen (32B), where cues that guide detectors that rely on surface signals tend to diminish. 

%In these settings, the predictability of valid answers varies widely across contexts: some answers are almost forced, while others depend on rare entities or admit multiple plausible phrasings. Detectors that rely mainly on surface signals can flag human text as LLM-generated text more frequently because both types of text appear unlikely to the evaluator. Our method mitigates this by accounting for context-dependent uncertainty, resulting in stronger threshold-free detection. On more creative datasets such as WritingPrompts, human and LLM text are often already well separated, so the additional gains are naturally smaller. Additionally, the advantage of our method becomes more visible as the generator becomes larger, e.g., Qwen (32B), where cues that guide detectors that rely on surface signals tend to diminish. 

\begin{table}[H] 
    \centering
    \resizebox{\textwidth}{!}{ % Resizes the table to fit within the page
    \begin{tabular}{lccc|ccc|ccc}
        \toprule
        & \multicolumn{3}{c|}{XSum} & \multicolumn{3}{c|}{SQuAD} & \multicolumn{3}{c}{WritingPrompts} \\
        \cmidrule(lr){2-4} \cmidrule(lr){5-7} \cmidrule(lr){8-10}
        Method & Llama3 (8B) & GPT-NeoX Erebus (20B) & Qwen (32B) &Llama3 (8B) & GPT-NeoX Erebus (20B) & Qwen (32B) & Llama3 (8B) & GPT-NeoX Erebus (20B) & Qwen
        (32B) \\
        \midrule
        $\log p(x)$ & 0.98 & 0.84 & 0.99* & 0.91 & 0.75 & 0.60 & 0.99* & 0.95 & 1.00* \\
        Rank & 0.75 & 0.69 & 0.69 & 0.71 & 0.71 & 0.55 & 0.82 & 0.80 & 0.81 \\
        LogRank & 0.98 & 0.87 & 0.99* & 0.93 & 0.81 & 0.62 & \textbf{1.00} & 0.97 & 1.00*\\
        Entropy & 0.39 & 0.70 & 0.40 & 0.37 & 0.66 & 0.70 & 0.04 & 0.36 & 0.02 \\
        DetectGPT & 0.78 & 0.95 & 0.99 *& 0.55 & 0.78 & 0.62 & 0.68 & 0.94 & 0.99 \\
        Binoculars & 0.99* & 0.99* & 0.99 *& 0.98* & 0.99* & 0.94* & 0.99 & 0.99* & 0.99 \\
        \textbf{Ours} & \textbf{0.99} & \textbf{0.99} & \textbf{1.00} & \textbf{0.99} & \textbf{0.99} & \textbf{0.97} & 0.98 & \textbf{1.00} & \textbf{1.00} \\
        \midrule
        Diff (zero-shot) & 0.00 & 0.00 & 0.01 & 0.01 & 0.01 & 0.03 & -0.02 & 0.01 & 0.00 \\
        \bottomrule
         RoBERTa (base) & 0.98* & 0.95 & 0.92* & 0.97* & 0.92 & 0.69* & 0.97* & 0.95* & 0.74* \\
         RoBERTa (large) & 0.98* & 0.98* & 0.92* & 0.95 & 0.93* & 0.68 & 0.96 & 0.93 & 0.65 \\
         \textbf{Ours} & \textbf{0.99} & \textbf{0.99} & \textbf{1.00} & \textbf{0.99} & \textbf{0.99} & \textbf{0.97} & \textbf{0.98} & \textbf{1.00} & \textbf{1.00} \\
        \midrule
        Diff (supervised) & 0.01 & 0.01 & 0.08& 0.02 & 0.06 & 0.28 & 0.01 &  0.05& 0.26 \\
         \bottomrule
    \end{tabular}
    } %\vspace{2mm}
    \caption{\footnotesize AUROC for detecting samples from the given models on the datasets. The best AUROC values are in \textbf{bold}, and the second-best values are marked with an asterisk (*). The rows Diff(zero-shot) and Diff(supervised) show our AUROC improvement over the strongest zero-shot and supervised baseline methods.
}
    \label{tab:auroc}
\end{table}

\subsubsection{Text Length Experiments} \label{app:length experiments}

The purpose of this experiment is to understand how much text is needed for reliable detection in different settings. We examine how performance changes as the text becomes longer and how many tokens are sufficient to achieve a given level of performance across different models and across news, fact-based, and creative-writing contexts. 

For the length experiment, we follow the setting for A-vs-not-A detection experiments in this section, where human text represents the null and text generated by the evaluator model represents the alternative\footnote{We use XSum, SQuAD, and Reddit WritingPrompts, taking the first 30 tokens of the document, context, and story fields, respectively, as the prompt; see Appendix~\ref{app:datasetdetails} for dataset details. The models are Llama3 (8B), GPT-NeoX Erebus (20B), and Qwen (32B).}. For each model--dataset pair, we vary the number of post-prompt tokens included in the statistic over $L \in \{10,20,30,40,50,75,100,125,150\},$ and report TPR at FPR $\leq 1\%$ and AUROC. This setup preserves the same white-box access and one-sided test used in this section, while isolating the role of the number of post-prompt tokens.

Taking the 150-token result as the benchmark for each model--dataset pair, TPR at FPR $\leq 1\%$ converges to within $\pm 2$ percentage points of this benchmark using 30--150 tokens in most cases. This pattern suggests that longer texts provide more evidence for distinguishing human- and LLM-generated content, with performance typically reaching a stable level after a moderate number of tokens. XSum and Writing typically converge with fewer (20--30) tokens, while SQuAD converges with more (50--100) tokens because, in factual contexts, human and LLM generations are often very similar under the evaluator. GPT-NeoX Erebus (20B) makes this domain difference particularly informative. Although its narrative specialization was expected to reduce the separation from human text in creative and news-style domains, its outputs on WritingPrompts and XSum become reliably detectable with comparatively short continuations. Interestingly, SQuAD requires longer continuations for the test statistic to achieve comparable separation between human- and Erebus-generated text. This happens because, in fact-based contexts, the distributions of the difference between log-perplexity and the evaluator model's average entropy are closer for human and Erebus-generated text, so more tokens are required. The managerial takeaway is that 30--150 token continuations are sufficient in most settings, with factual contexts requiring more tokens within this range.

\begin{figure}[H]
\centering
\includegraphics[width=0.6\linewidth]{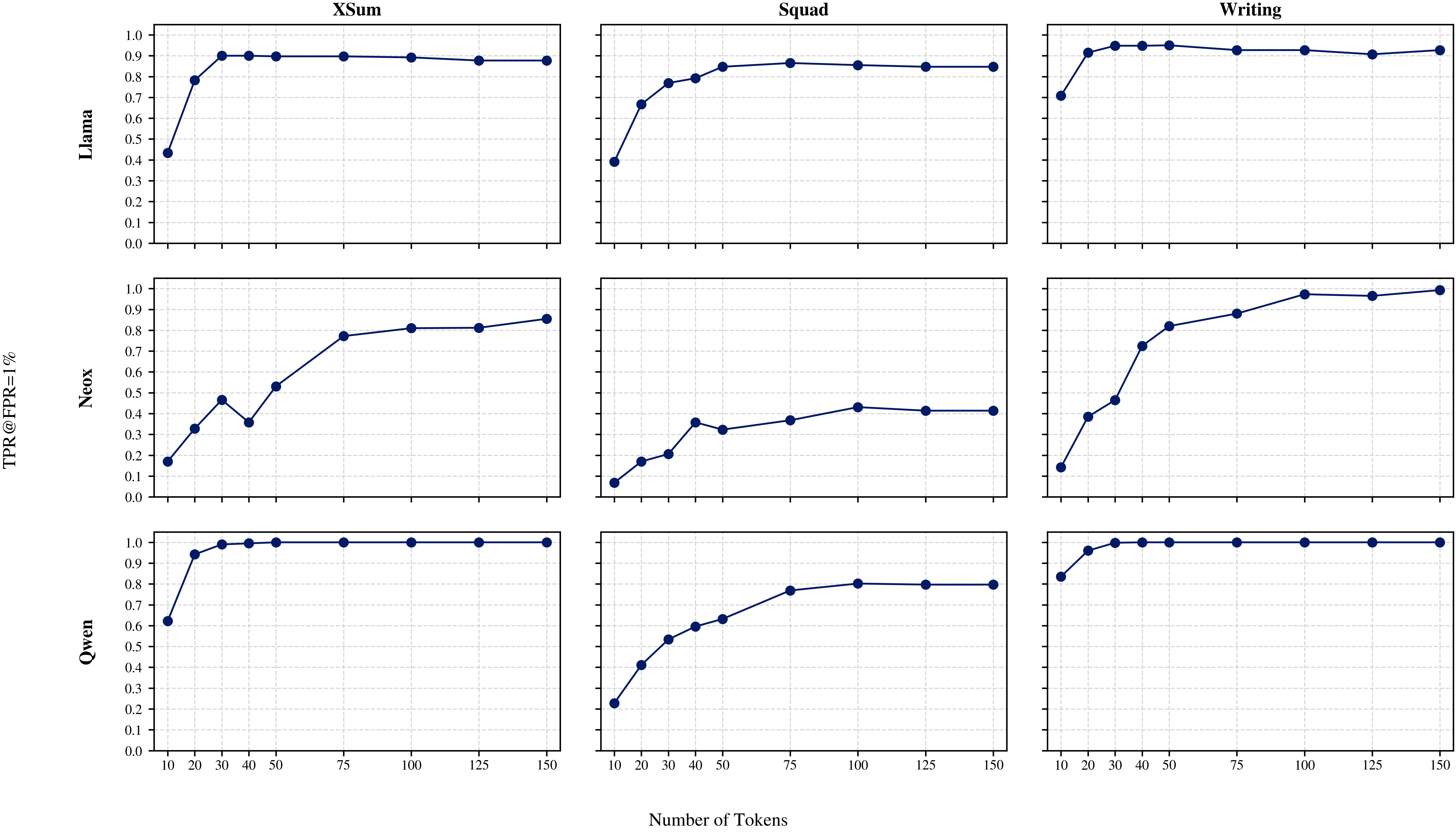}
    \caption{TPR@FPR$\leq1$\% for different text lengths}
    \label{fig:lengthTPRFPR}
\end{figure}

For AUROC, convergence to within $\pm 2$ percentage points of the 150-token benchmark occurs by at most 75 tokens for all model--dataset pairs. XSum and Writing again tend to converge with fewer tokens, while SQuAD generally requires more tokens within this range. This difference (i.e., convergence with fewer tokens) from TPR at FPR $\leq 1\%$ is expected because AUROC presents performance across thresholds, whereas the low-FPR metric evaluates the test under a stringent false-positive constraint.

\begin{figure}[H]
\centering
\includegraphics[width=0.6\linewidth]{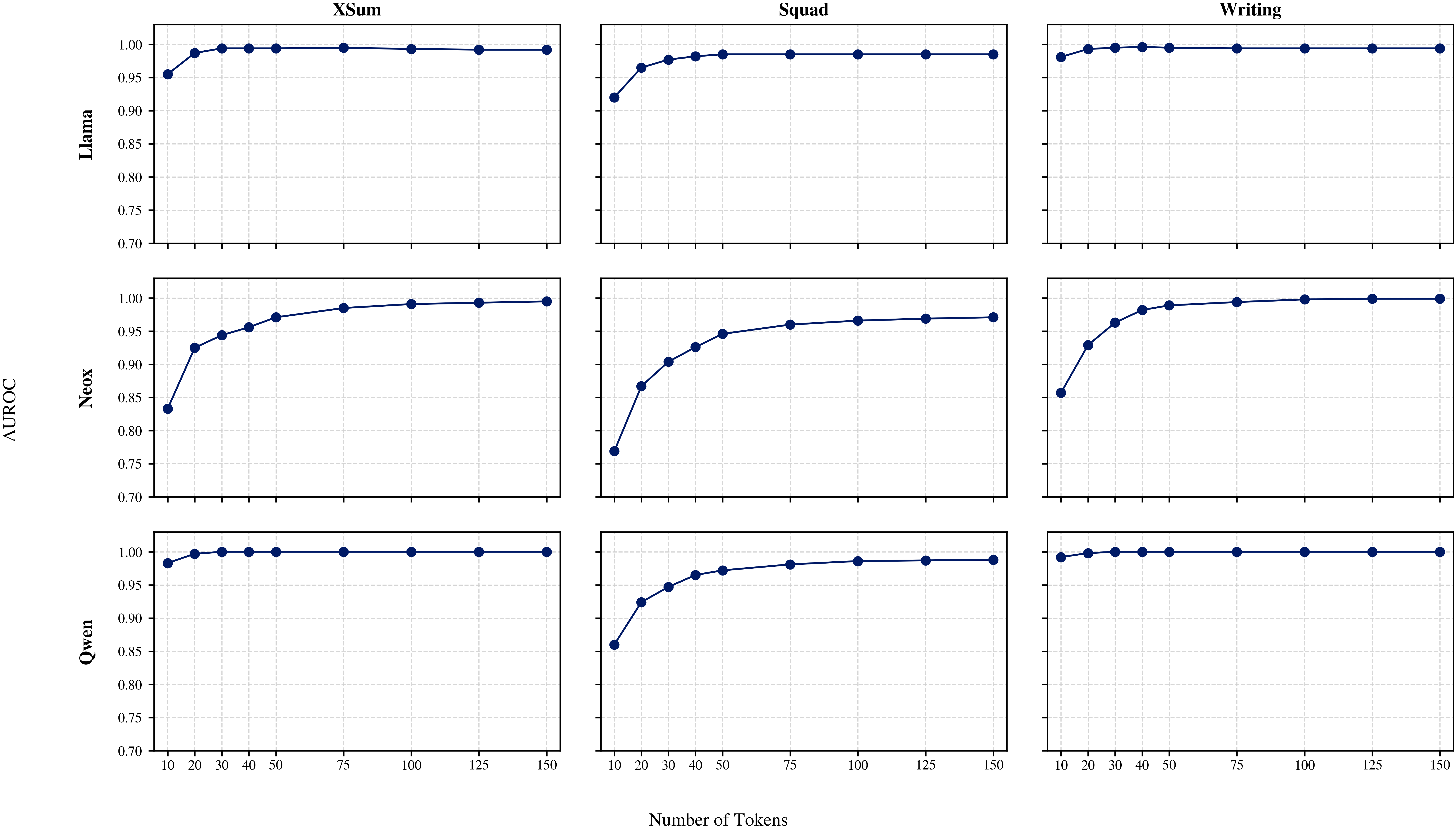}
    \caption{AUROC for different text lengths}
    \label{fig:lengthAUROC}
\end{figure}

\subsection{Robustness Experiments} \label{blackboxnumerical} 

In this section, we empirically examine performance in a setting where an evaluator model is inaccessible. That is, we have no access to an evaluator model—not to its conditional probabilities and not through sampling—and also evaluate performance across decoding strategies and post-generation text manipulations. Human text represents the null, and text generated by an inaccessible language model represents the alternative. Because the target model is inaccessible, we use a surrogate model for scoring, following the widely used working hypothesis that language models are more similar to one another than to human text \citep{pillutla2021mauve}. This working hypothesis motivates using an accessible LLM as a surrogate scoring model for empirical human-versus-LLM detection. We use the scoring model \texttt{tiiuae/falcon-rw-1B} to compute the log-perplexity and average entropy of each text and use the resulting score for classification. This no-access surrogate scoring setup is consistent with prior empirical work on LLM-generated text detection (e.g., \citealp{mitchell2023detectgpt,dugan2024raid})\footnote{This premise is used explicitly by \citep{hans2024spotting,mitchell2023detectgpt} and is consistent with recent comparative studies on matched English news settings reporting that differences among LLM outputs are typically smaller than the differences between LLM and human writing \citep{zamaraeva-etal-2025-comparing,munoz-ortiz-etal-2024-contrasting}. There are few conditions where this observation may not hold; for example, in New York Times--style news, humans and LLMs can look very similar on some basic word-form patterns, which can reduce the gap depending on what is being measured, even though broader grammatical measures still show LLMs clustering more tightly \citep{zamaraeva-etal-2025-comparing}.}. Following \cite{dugan2024raid}, we use TPR at 5$\%$ FPR as our evaluation metric.

We use the RAID dataset \citep{dugan2024raid}, which consists of 14,971 human-written texts from public pre-2022 datasets from 8 domains of abstracts, books, news, poetry, recipes, Reddit, reviews, and Wikipedia. For each human text, \cite{dugan2024raid} create a corresponding generation prompt using the template "Write a recipe for \emph{title}". Then, they generate one output for each of the 11 models, 4 decoding strategies, and 11 adversarial attacks. For details about decoding strategies and adversarial attacks, please refer to Appendix \ref{app:adversarial attacks}. The 11 models are: open-source chat models Llama-c, Mistral-c, and MPT-c; open-source non-chat models Mistral, MPT, and GPT2; closed-source chat models c-GPT, GPT4, and Cohere; and closed-source non-chat models Cohere and GPT3. They consider 4 decoding strategies formed by combining (i) greedy, which picks the highest probability token at each generation, or sampling, which samples according to a generation probability $p_n^{B}(.)$, with (ii) repetition penalty or no repetition penalty. The repetition penalty evades detectors by reducing the likelihood of generating previously used tokens. This is achieved by down-weighting their probabilities by a factor $\theta \in \{1,1.2\}$, resulting in less repetitive output. After generating each text, they apply adversarial attacks that manipulate text on all samples, including human text and the $11\times4$ LLM generations. Because these manipulations are applied to both human and LLM-generated texts, they can affect both Type I and Type II errors. The nine query-free attacks are: Alternative Spelling (AS), Article Deletion (AD), Insert Paragraphs (IP), Number Swap (NS), Misspelling (MS), Synonym Swap (SS), Upper Lower Swap (ULS), Paraphrasing (PP), and Whitespace Addition (WSA). The described process yields a total of 134,739 human texts and 5,928,516 LLM texts\footnote{Given the cost of more than \$10,000 for re-running the experiment for the performance of benchmark detectors on approximately 6M texts, we use the results of \cite{dugan2024raid}'s experiments on the other detectors and run the experiment for our own detector.}. For details about hardware and computation costs, refer to Appendix \ref{app:hardwaredetails}.

We compare our performance with the neural, zero-shot, and commercial detectors in \cite{dugan2024raid}. The neural benchmarks are RoBERTa-B (GPT-2), RoBERTa-L (GPT-2), RoBERTa-B (ChatGPT), and RADAR. The zero-shot benchmarks are GLTR, Fast-DetectGPT, Binoculars, and LLMDet. The commercial benchmarks are GPTZero, Originality, Winston, and ZeroGPT. For details about detectors, please refer to Appendix \ref{app:detectordetails}. 
In presenting our results, we use three coarse tiers ($\leq$ 33\%, 33–67\%, and $\geq$ 67\%) to improve readability of the tables by visually separating detection performances (TPR at 5\% FPR). The colors are intended merely as a visualization aid and should not be interpreted as performance standards or statistical significance thresholds. 

A consistent pattern across methods in Table~\ref{tab:bbperformance} is that greedy generations tend to be easier to detect than sampled generations: greedy decoding concentrates probability mass on high-likelihood tokens, yielding text that is more predictable under a scoring model, whereas sampling introduces lower-probability token choices that often move an LLM text closer to the human distribution. Repetition penalties reshape the generator's token probabilities, and they can either reduce or increase detection probability depending on how they interact with the scorer's tokenization\footnote{A repetition penalty distorts token probabilities. In some settings, it can force the generator to make low-probability lexical substitutions even when the surrounding context remains predictable. That can increase surprisal without a matching increase in contextual uncertainty.}. Our approach remains comparatively steady across the generator/decoding combinations, suggesting that combining log-perplexity with an entropy-based statistic can be less vulnerable to shifts in decoding strategy than methods that rely on a single cue. The table also highlights the fundamental limitation of no-access surrogate scoring: when sampling produces highly diverse outputs that overlap more with the variability of human writing under the scoring model, all detectors, including ours, face a more difficult detection problem at a strict 5\% FPR. Fast-DetectGPT relies on changes in log-likelihood across perturbed versions of the text, and its performance can decline when those perturbations do not produce a consistent change in log-likelihood. Binoculars normalizes perplexity using a second model, which can reduce sensitivity to domain differences but may also reduce separation when the two models assign similar probabilities to the same continuations. In contrast, our method compares the observed log-perplexity with the surrogate model's average entropy and remains comparatively stable across the generator and decoding settings reported in Table~\ref{tab:bbperformance}. For a more detailed comparison with Fast-DetectGPT and Binoculars, see Appendix~\ref{app:detectordetails}. Commercial detectors such as Originality are trained on broad collections of public online and private texts and may combine multiple detection signals. This extensive training exposure can favor performance on benchmark distributions similar to those encountered during development, so the reported results may not fully represent performance on previously unseen domains and writing styles.

\begin{table}[H]
\centering
\scriptsize
\renewcommand{\arraystretch}{0.7}
\begin{tabular}{@{} l *{12}{c} @{}}
\toprule
& \multicolumn{8}{c}{\textbf{Open-Source}} & \multicolumn{4}{c}{\textbf{Closed-Source}}\\
\cmidrule(lr){2-9} \cmidrule(lr){10-13}
& \multicolumn{4}{c}{\textbf{Chat Models}} & \multicolumn{4}{c}{\textbf{Non-Chat Models}} & \multicolumn{2}{c}{\textbf{Chat Models}} & \multicolumn{2}{c}{\textbf{Non-Chat Models}} \\
\cmidrule(lr){2-5} \cmidrule(lr){6-9} \cmidrule(lr){10-11} \cmidrule(lr){12-13}
& \multicolumn{4}{c}{\textit{(Llama-c, mistral-c, mpt-c)}} & \multicolumn{4}{c}{\textit{(mistral, mpt, gpt2)}} & \multicolumn{2}{c}{\textit{(c-gpt, gpt4, cohere)}} & \multicolumn{2}{c}{\textit{(cohere, gpt3)}} \\
\cmidrule(lr){1-1}\cmidrule(lr){2-5} \cmidrule(lr){6-9} \cmidrule(lr){10-11} \cmidrule(lr){12-13}
\textbf{Dec. Strategy}
& \multicolumn{2}{c}{\textbf{greedy}} & \multicolumn{2}{c}{\textbf{sampling}} & \multicolumn{2}{c}{\textbf{greedy}} & \multicolumn{2}{c}{\textbf{sampling}} & \textbf{greedy} & \textbf{sampling} & \textbf{greedy} & \textbf{sampling} \\
\textbf{Rep. Penalty?}
& $\times$ & $\checkmark$ & $\times$ & $\checkmark$ & $\times$ & $\checkmark$ & $\times$ & $\checkmark$ & $\times$ & $\times$ & $\times$ & $\times$ \\
\midrule
R-B GPT2 & \cellcolor{green!30}84.1 & \cellcolor{yellow!30}52.3 & \cellcolor{green!30}77.9 & \cellcolor{red!30}26.2 & \cellcolor{green!30}98.6 & \cellcolor{yellow!30}44.1 & \cellcolor{yellow!30}60.5 & \cellcolor{yellow!30}35.4 & \cellcolor{green!30}70.9 & \cellcolor{yellow!30}41.7 & \cellcolor{yellow!30}65.1 & \cellcolor{yellow!30}52.5 \\
R-L GPT2 & \cellcolor{green!30}79.7 & \cellcolor{yellow!30}41.1 & \cellcolor{green!30}71.4 & \cellcolor{red!30}19.5 & \cellcolor{green!30}98.5 & \cellcolor{yellow!30}43.0 & \cellcolor{green!30}67.2 & \cellcolor{yellow!30}53.4 & \cellcolor{yellow!30}61.4 & \cellcolor{yellow!30}34.7 & \cellcolor{yellow!30}61.1 & \cellcolor{yellow!30}48.6 \\
R-B CGPT & \cellcolor{green!30}80.2 & \cellcolor{yellow!30}63.3 & \cellcolor{green!30}75.0 & \cellcolor{yellow!30}39.3 & \cellcolor{yellow!30}53.3 & \cellcolor{red!30}26.4 & \cellcolor{red!30}14.9 & \cellcolor{red!30}1.7 & \cellcolor{yellow!30}59.1 & \cellcolor{yellow!30}38.1 & \cellcolor{yellow!30}46.5 & \cellcolor{yellow!30}39.0 \\
RADAR & \cellcolor{green!30}88.8 & \cellcolor{green!30}77.4 & \cellcolor{green!30}85.6 & \cellcolor{yellow!30}66.4 & \cellcolor{green!30}91.8 & \cellcolor{yellow!30}63.8 & \cellcolor{yellow!30}48.3 & \cellcolor{red!30}31.8 & \cellcolor{green!30}81.6 & \cellcolor{green!30}75.3 & \cellcolor{green!30}72.2 & \cellcolor{green!30}67.7 \\
\midrule
GPTZero & \cellcolor{green!30}98.8 & \cellcolor{green!30}93.7 & \cellcolor{green!30}98.4 & \cellcolor{green!30}82.5 & \cellcolor{green!30}74.7 & \cellcolor{yellow!30}34.6 & \cellcolor{red!30}9.4 & \cellcolor{red!30}4.8 & \cellcolor{green!30}92.3 & \cellcolor{green!30}88.5 & \cellcolor{yellow!30}60.6 & \cellcolor{yellow!30}53.4 \\
Originality & \cellcolor{green!30}98.6 & \cellcolor{green!30}86.3 & \cellcolor{green!30}97.7 & \cellcolor{green!30}72.5 & \cellcolor{green!30}99.9 & \cellcolor{yellow!30}64.1 & \cellcolor{green!30}89.0 & \cellcolor{yellow!30}51.2 & \cellcolor{green!30}96.8 & \cellcolor{green!30}89.0 & \cellcolor{green!30}91.7 & \cellcolor{green!30}85.4 \\
Winston & \cellcolor{green!30}97.2 & \cellcolor{green!30}90.1 & \cellcolor{green!30}96.6 & \cellcolor{green!30}78.3 & \cellcolor{green!30}68.2 & \cellcolor{yellow!30}49.0 & \cellcolor{red!30}29.5 & \cellcolor{red!30}11.3 & \cellcolor{green!30}96.1 & \cellcolor{green!30}93.7 & \cellcolor{green!30}73.2 & \cellcolor{green!30}68.1 \\
ZeroGPT(*) & \cellcolor{green!30}95.4 & \cellcolor{green!30}80.7 & \cellcolor{green!30}90.5 & \cellcolor{yellow!30}54.9 & \cellcolor{green!30}85.1 & \cellcolor{yellow!30}57.2 & \cellcolor{red!30}16.0 & \cellcolor{red!30}0.3 & \cellcolor{green!30}92.1 & \cellcolor{yellow!30}65.8 & \cellcolor{green!30}83.4 & \cellcolor{green!30}72.7 \\
\midrule
GLTR & \cellcolor{green!30}89.8 & \cellcolor{green!30}67.5 & \cellcolor{green!30}83.9 & \cellcolor{yellow!30}38.3 & \cellcolor{green!30}99.6 & \cellcolor{yellow!30}56.9 & \cellcolor{yellow!30}44.5 & \cellcolor{red!30}0.5 & \cellcolor{green!30}80.7 & \cellcolor{yellow!30}54.3 & \cellcolor{green!30}75.6 & \cellcolor{yellow!30}63.7 \\
Fast-DetectGPT & \cellcolor{green!30}98.6 & \cellcolor{green!30}74.5 & \cellcolor{green!30}96.2 & \cellcolor{yellow!30}40.5 & \cellcolor{green!30}97.8 & \cellcolor{yellow!30}56.1 & \cellcolor{green!30}79.7 & \cellcolor{red!30}0.6 & \cellcolor{green!30}96.0 & \cellcolor{green!30}74.1 & \cellcolor{green!30}93.8 & \cellcolor{green!30}86.3 \\
LLMDet & \cellcolor{yellow!30}55.5 & \cellcolor{red!30}30.2 & \cellcolor{yellow!30}47.5 & \cellcolor{red!30}16.5 & \cellcolor{green!30}74.8 & \cellcolor{red!30}27.0 & \cellcolor{yellow!30}38.4 & \cellcolor{red!30}3.7 & \cellcolor{yellow!30}35.8 & \cellcolor{red!30}18.5 & \cellcolor{yellow!30}40.0 & \cellcolor{red!30}32.9 \\
Binoculars & \cellcolor{green!30}99.9 & \cellcolor{green!30}86.6 & \cellcolor{green!30}99.7 & \cellcolor{yellow!30}60.6 & \cellcolor{green!30}99.9 & \cellcolor{yellow!30}62.3 & \cellcolor{green!30}72.4 & \cellcolor{red!30}0.6 & \cellcolor{green!30}99.2 & \cellcolor{green!30}92.1 & \cellcolor{green!30}99.0 & \cellcolor{green!30}95.0 \\
\midrule
\textbf{Ours} & \cellcolor{green!30}99.4 & \cellcolor{green!30}86.4 & \cellcolor{green!30}97.9 & \cellcolor{yellow!30}62.2 & \cellcolor{green!30}90.8 & \cellcolor{green!30}84.0 & \cellcolor{yellow!30}54.0 & \cellcolor{green!30}73.3 & \cellcolor{green!30}96.4 & \cellcolor{green!30}70.2 & \cellcolor{green!30}94.4 & \cellcolor{green!30}86.3 \\
\bottomrule
\end{tabular}
\caption{\footnotesize TPR at FPR=5$\%$. Cell colors: Red \ensuremath{\leq} 33, Yellow 33--66, Green \ensuremath{\geq} 67.}
\label{tab:bbperformance}
\end{table}
%Results in Table \ref{tab:bbperformance} demonstrate high performance of our detector in black-box detection with and without sampling and repetition penalty. For example, while other detectors have low TPR when detecting open-source non-chat models with sampling and repetition penalty, we achieve a TPR of 73.3$\%$. Furthermore, we observe that, unlike all other zero-shot and learning-based methods, the performance of our framework never falls below the low threshold, with our lowest performance being 54.0$\%$. The average performance of our detector in black-box detection is 82.9$\%$, which (except for Originality, a commercial detector with TPR=85.2$\%$) is higher than the average performance of other detectors, with the second highest performance being Binoculars with TPR=80.6 $\%$.

Table \ref{tab:adversarial} indicates that our detector maintains relatively steady performance under several types of text modification, including alternative spelling, article deletion, paragraph insertion, number swapping, misspelling, case swapping, and paraphrasing. The largest performance reductions occur under synonym swap and whitespace addition. Fortunately, whitespace addition can be removed through a simple pre-processing step that removes added whitespace. Our method does not require training, yet across the ten conditions in Table~\ref{tab:adversarial}, it attains the highest average TPR among the learning-based and zero-shot benchmarks and the second-highest average overall, behind only Originality. This comparison may favor learning-based and commercial detectors (including Originality) because they can benefit when the evaluated texts resemble their training distributions. Commercial detectors may additionally combine multiple detection signals, including learning-based components, further favoring their performance in this benchmark setting. Accordingly, the performance of learning-based and commercial detectors on these evaluated perturbations may not fully represent robustness to previously unseen domains or editing strategies. Among zero-shot baselines, Binoculars is also comparatively robust across several attacks, consistent with the idea that its two-model normalization can mitigate some domain- and formatting-induced shifts that confound single-model likelihood.
\begin{table}[H] 
\renewcommand{\arraystretch}{0.3}
\centering
\begin{tabular}{l|cccccccccc} 
\toprule
 & None & AS & AD & IP & NS & MS & SYN & ULS & WSA & PP \\
\midrule
RoB-B GPT2 & \cellcolor{yellow!30}59.1 & \cellcolor{yellow!30}55.6 & \cellcolor{yellow!30}37.1 & \cellcolor{yellow!30}56.9 & \cellcolor{yellow!30}55.9 & \cellcolor{yellow!30}43.8 & \cellcolor{green!30}71.5 & \cellcolor{red!30}18.8 & \cellcolor{yellow!30}45.2 & \cellcolor{green!30}68.9 \\
RoB-L GPT2 & \cellcolor{yellow!30}56.7 & \cellcolor{yellow!30}52.4 & \cellcolor{yellow!30}33.2 & \cellcolor{yellow!30}55.1 & \cellcolor{yellow!30}51.7 & \cellcolor{yellow!30}39.5 & \cellcolor{green!30}79.4 & \cellcolor{red!30}19.3 & \cellcolor{yellow!30}40.1 & \cellcolor{green!30}72.9 \\
RoB-B CGPT & \cellcolor{yellow!30}44.8 & \cellcolor{yellow!30}43.3 & \cellcolor{yellow!30}38.0 & \cellcolor{red!30}5.2  & \cellcolor{yellow!30}44.3 & \cellcolor{yellow!30}42.1 & \cellcolor{yellow!30}39.6 & \cellcolor{red!30}31.7 & \cellcolor{red!30}0.1  & \cellcolor{yellow!30}49.2 \\
RADAR & \cellcolor{green!30}70.9 & \cellcolor{green!30}70.8 & \cellcolor{green!30}67.9 & \cellcolor{green!30}73.7 & \cellcolor{green!30}71.0 & \cellcolor{green!30}69.5 & \cellcolor{green!30}67.5 & \cellcolor{green!30}70.4 & \cellcolor{yellow!30}66.1 & \cellcolor{green!30}67.3\\
\midrule
GPTZero & \cellcolor{green!30}66.5 & \cellcolor{yellow!30}64.9 & \cellcolor{yellow!30}61.0 & \cellcolor{green!30}66.2 & \cellcolor{yellow!30}65.8 & \cellcolor{yellow!30}65.1 & \cellcolor{yellow!30}61.0 & \cellcolor{yellow!30}56.5 & \cellcolor{green!30}66.2 & \cellcolor{yellow!30}64.0 \\
ZeroGPT & \cellcolor{yellow!30}65.5 & \cellcolor{yellow!30}65.4 & \cellcolor{yellow!30}59.7 & \cellcolor{yellow!30}64.9 & \cellcolor{yellow!30}64.7 & \cellcolor{yellow!30}64.7 & \cellcolor{red!30}18.8 & \cellcolor{yellow!30}54.5 & \cellcolor{yellow!30}64.2 & \cellcolor{yellow!30}46.7 \\
Originality & \cellcolor{green!30}85.0 & \cellcolor{green!30}83.6 & \cellcolor{green!30}71.4 & \cellcolor{green!30}85.1 & \cellcolor{green!30}86.0 & \cellcolor{green!30}78.6 & \cellcolor{green!30}96.5 & \cellcolor{green!30}75.8 & \cellcolor{green!30}84.9 & \cellcolor{green!30}96.7 \\
Winston & \cellcolor{green!30}71.0 & \cellcolor{green!30}68.9 & \cellcolor{green!30}66.9 & \cellcolor{green!30}69.8 & \cellcolor{green!30}69.0 & \cellcolor{green!30}67.5 & \cellcolor{yellow!30}63.6 & \cellcolor{yellow!30}56.8 & \cellcolor{yellow!30}46.8 & \cellcolor{yellow!30}52.6 \\
\midrule
GLTR & \cellcolor{yellow!30}62.6 & \cellcolor{yellow!30}61.2 & \cellcolor{yellow!30}52.1 & \cellcolor{yellow!30}61.4 & \cellcolor{yellow!30}59.9 & \cellcolor{yellow!30}59.8 & \cellcolor{red!30}31.2 & \cellcolor{yellow!30}48.1 & \cellcolor{yellow!30}45.8 & \cellcolor{yellow!30}47.2 \\
Fast-DetectGPT & \cellcolor{green!30}73.6 & \cellcolor{green!30}71.6 & \cellcolor{yellow!30}64.7 & \cellcolor{green!30}72.0 & \cellcolor{green!30}68.2 & \cellcolor{green!30}70.7 & \cellcolor{yellow!30}34.0 & \cellcolor{yellow!30}60.4 & \cellcolor{yellow!30}64.4 & \cellcolor{green!30}71.8 \\
LLMDet & \cellcolor{yellow!30}35.0 & \cellcolor{yellow!30}33.9 & \cellcolor{red!30}27.4 & \cellcolor{red!30}27.2 & \cellcolor{yellow!30}33.8 & \cellcolor{red!30}32.7 & \cellcolor{red!30}27.3 & \cellcolor{red!30}23.4 & \cellcolor{red!30}4.4 & \cellcolor{red!30}28.5 \\
Binoculars & \cellcolor{green!30}79.6 & \cellcolor{green!30}78.2 & \cellcolor{green!30}74.3 & \cellcolor{green!30}71.7 & \cellcolor{green!30}77.1 & \cellcolor{green!30}78.0 & \cellcolor{yellow!30}43.5 & \cellcolor{green!30}73.8 & \cellcolor{green!30}70.1 & \cellcolor{green!30}80.3 \\
\midrule
\textbf{Ours} & \cellcolor{green!30}82.5 & \cellcolor{green!30}80.6 & \cellcolor{green!30}74.4 & \cellcolor{green!30}81.1 & \cellcolor{green!30}74.8 & \cellcolor{green!30}79.8 & \cellcolor{yellow!30}48.6 & \cellcolor{green!30}72.5 & \cellcolor{yellow!30} 60.8 & \cellcolor{green!30}74.2 \\
\bottomrule
\end{tabular}
\caption{\footnotesize TPR @ FPR=5$\%$. Abbreviations are:
AS: Alternative Spelling, AD: Article Deletion, IP: Insert Paragraphs, NS: Number Swap, MS: Misspelling, SYN: Synonym Swap, ULS: Upper Lower Swap, WSA: Whitespace Addition, PP: Paraphrasing. Cell colors: Red $<$ 33, Yellow 33--66, Green \ensuremath{\geq} 67.}
\label{tab:adversarial}
\end{table}

\section{Conclusion} \label{sec:conclusion} %\vspace{-3mm}

We study the problem of LLM provenance verification: given a prompt and a finite-length text, decide whether the text was produced by a particular LLM, while controlling the probability of false positives. This capability is increasingly important for accountability and integrity in digital communication, including assigning responsibility for content, enforcing organizational policies that restrict which models may be used, and supporting AI providers in protecting training pipelines. Motivated in particular by the high cost of accusations (false positives) in regulatory settings, we design zero-shot statistical tests with error guarantees that remain informative at practical text lengths.

We model LLM-generated text as an auto-regressive stochastic process with full dependence on history and build tests based on the convergence of log-perplexity under an evaluator model \(A\) to the average cross-entropy between the evaluator model $A$ and the generative model $B$. Using these concentration results, we develop (i) a composite attribution test that distinguishes whether the text string \(\mathbf{Y}_N\) was generated by a model in a non-sanctioned set \(\mathcal{A}\) or in a disjoint in-house set \(\mathcal{B}\), and (ii) a detection test that distinguishes whether \(\mathbf{Y}_N\) was generated by a known model \(A\) or by an unknown distinguishable alternative ``not \(A\).'' In Propositions \ref{thm:statstestmulti} and \ref{thm:statstest}, we prove finite-\(N\) upper bounds showing that both Type~I and Type~II errors decay exponentially in the text length \(N\). In Appendix \ref{sec:blackbox}, we extend these tests to the black-box setting, where evaluator log-probabilities must be approximated via sampling, and we provide a sample size sufficient to achieve analogous finite-sample errors in Propositions \ref{thm:statstestmultibb} and \ref{thm:statstestsampling}. We also provide a lower bound in Corollary \ref{cor:lecam-iid}, showing that there exist model pairs for which no statistical test can make both Type~I and Type~II errors decay faster than exponentially in the text length \(N\).

%We also provide a lower bound in an i.i.d.\\ benchmark in Corollary \ref{cor:lecam-iid} showing that no test can achieve error rates that decay faster than exponentially in the text length \(N\), matching our upper-bound scaling up to constants.

Our main methodological contribution is Theorem \ref{thm:maindiff}, which gives an exponential-decay concentration inequality for the deviation between the log-perplexity of a dependent token sequence generated under \(B\) and evaluated under \(A\), and the corresponding average cross-entropy between $B$ and $A$, over a finite alphabet. This is a complex problem, and the novelty lies in the reduction required by sequential text generation: expressing the deviation between log-perplexity and average cross-entropy as a martingale with respect to the growing token history and deriving the corresponding increment bounds, after which martingale concentration inequalities yield the non-asymptotic, finite-$N$ guarantees. This result generalizes the i.i.d.\ setting studied in \cite{zhao2022optimal} to the dependent regime relevant for LLMs' text generation. 

We present numerical experiments that empirically examine the performance of our method in practical settings and show that it performs very well across a range of attribution, detection, and no-access settings. In particular, our method attains the highest or tied-highest TPR at FPR $\leq 1\%$ in seven of nine white-box detection settings, achieves AUROC values between $0.97$ and $1.00$ across all nine settings, and remains comparatively stable across many decoding strategies and adversarial edits. The bounds are conservative guarantees; empirically, performance stabilizes at shorter text lengths of 30--150 tokens. The results also identify systematic sources of difficulty: within-family similarity increases attribution errors, and factual contexts often require longer passages than news and creative-writing contexts. In the appendices, we further examine the setting with only sampling access to the evaluator model and show how performance changes with the sample size used to estimate empirical log-perplexity and empirical average cross-entropy. A key managerial takeaway from our experiments is that reliable detection in short, factual, Wikipedia-like tasks requires caution because human and LLM responses can be highly similar in such contexts; managers and educators who design tasks or assignments should therefore require examples, descriptions, or reasoning that provide more textual evidence for detection. A future direction is to develop provable guarantees under adversarial attacks.

\bibliographystyle{informs2014}
% For MS submission
\bibliography{references}

@article{solaiman2019release,
  title={Release strategies and the social impacts of language models},
  author={Solaiman, Irene and Brundage, Miles and Clark, Jack and Askell, Amanda and Herbert-Voss, Ariel and Wu, Jeff and Radford, Alec and Krueger, Gretchen and Kim, Jong Wook and Kreps, Sarah and others},
  journal={arXiv preprint arXiv:1908.09203},
  year={2019}
}

@inproceedings{gehrmann2019gltr,
  title={Gltr: Statistical detection and visualization of generated text},
  author={Gehrmann, Sebastian and Strobelt, Hendrik and Rush, Alexander M},
  booktitle={Proceedings of the 57th annual meeting of the association for computational linguistics: system demonstrations},
  pages={111--116},
  year={2019}
}

@inproceedings{su2023detectllm,
  title={Detectllm: Leveraging log rank information for zero-shot detection of machine-generated text},
  author={Su, Jinyan and Zhuo, Terry and Wang, Di and Nakov, Preslav},
  booktitle={Findings of the Association for Computational Linguistics: EMNLP 2023},
  pages={12395--12412},
  year={2023}
}

@inproceedings{verma2023ghostbuster,
  title={Ghostbuster: Detecting text ghostwritten by large language models},
  author={Verma, Vivek and Fleisig, Eve and Tomlin, Nicholas and Klein, Dan},
  booktitle={Proceedings of the 2024 Conference of the North American Chapter of the Association for Computational Linguistics: Human Language Technologies (Volume 1: Long Papers)},
  pages={1702--1717},
  year={2024}
}

@article{vasilatos2023howkgpt,
  title={Howkgpt: Investigating the detection of chatgpt-generated university student homework through context-aware perplexity analysis},
  author={Vasilatos, Christoforos and Alam, Manaar and Rahwan, Talal and Zaki, Yasir and Maniatakos, Michail},
  journal={arXiv preprint arXiv:2305.18226},
  year={2023}
}

@article{radford2018improving,
  title={Improving language understanding by generative pre-training},
  author={Radford, Alec and Narasimhan, Karthik and Salimans, Tim and Sutskever, Ilya},
  year={2018},
  publisher={San Francisco, CA, USA}
}

@book{vershynin2018high,
  title={High-dimensional probability: An introduction with applications in data science},
  author={Vershynin, Roman},
  volume={47},
  year={2018},
  publisher={Cambridge university press}
}

@article{goldberg1991every,
  title={What every computer scientist should know about floating-point arithmetic},
  author={Goldberg, David},
  journal={ACM computing surveys (CSUR)},
  volume={23},
  number={1},
  pages={5--48},
  year={1991},
  publisher={ACM New York, NY, USA}
}

@article{guo2023close,
  title={How close is chatgpt to human experts? comparison corpus, evaluation, and detection},
  author={Guo, Biyang and Zhang, Xin and Wang, Ziyuan and Jiang, Minqi and Nie, Jinran and Ding, Yuxuan and Yue, Jianwei and Wu, Yupeng},
  journal={arXiv preprint arXiv:2301.07597},
  year={2023}
}

@inproceedings{mitchell2023detectgpt,
  title={Detectgpt: Zero-shot machine-generated text detection using probability curvature},
  author={Mitchell, Eric and Lee, Yoonho and Khazatsky, Alexander and Manning, Christopher D and Finn, Chelsea},
  booktitle={International Conference on Machine Learning},
  pages={24950--24962},
  year={2023},
  organization={PMLR}
}

@inproceedings{kirchenbauer2023watermark,
  title={A watermark for large language models},
  author={Kirchenbauer, John and Geiping, Jonas and Wen, Yuxin and Katz, Jonathan and Miers, Ian and Goldstein, Tom},
  booktitle={International Conference on Machine Learning},
  pages={17061--17084},
  year={2023},
  organization={PMLR}
}

@article{krishna2023paraphrasing,
  title={Paraphrasing evades detectors of ai-generated text, but retrieval is an effective defense},
  author={Krishna, Kalpesh and Song, Yixiao and Karpinska, Marzena and Wieting, John and Iyyer, Mohit},
  journal={Advances in neural information processing systems},
  volume={36},
  pages={27469--27500},
  year={2023}
}

@article{sadasivan2023can,
  title={Can AI-generated text be reliably detected?},
  author={Sadasivan, Vinu Sankar and Kumar, Aounon and Balasubramanian, Sriram and Wang, Wenxiao and Feizi, Soheil},
  journal={arXiv preprint arXiv:2303.11156},
  year={2023}
}

@article{zhao2022optimal,
  title={An optimal uniform concentration inequality for discrete entropies on finite alphabets in the high-dimensional setting},
  author={Zhao, Yunpeng},
  journal={Bernoulli},
  volume={28},
  number={3},
  pages={1892--1911},
  year={2022},
  publisher={Bernoulli Society for Mathematical Statistics and Probability}
}

@article{nature2024watermarking,
  author       = {Nature},
  title        = {Why AI watermarking isn’t working — and what to do instead},
  journal      = {Nature},
  year         = {2024},
  url          = {https://www.nature.com/articles/d41586-024-03418-x},
  note         = {Accessed: 2024-12-12}
}

@article{hans2024spotting,
  title={Spotting llms with binoculars: Zero-shot detection of machine-generated text},
  author={Hans, Abhimanyu and Schwarzschild, Avi and Cherepanova, Valeriia and Kazemi, Hamid and Saha, Aniruddha and Goldblum, Micah and Geiping, Jonas and Goldstein, Tom},
  journal={arXiv preprint arXiv:2401.12070},
  year={2024}
}

@article{narayan2018don,
  title={Don’t give me the details, just the summary},
  author={Narayan, Shashi and Cohen, Shay B and Lapata, Mirella},
  journal={Topic-aware convolutional neural networks for extreme summarization. ArXiv abs/1808.08745},
  year={2018}
}

@inproceedings{rajpurkar2016squad,
  title={Squad: 100,000+ questions for machine comprehension of text},
  author={Rajpurkar, Pranav and Zhang, Jian and Lopyrev, Konstantin and Liang, Percy},
  booktitle={Proceedings of the 2016 conference on empirical methods in natural language processing},
  pages={2383--2392},
  year={2016}
}

@article{fan2018hierarchical,
  title={Hierarchical neural story generation},
  author={Fan, Angela and Lewis, Mike and Dauphin, Yann},
  journal={arXiv preprint arXiv:1805.04833},
  year={2018}
}

@article{ippolito2019automatic,
  title={Automatic detection of generated text is easiest when humans are fooled},
  author={Ippolito, Daphne and Duckworth, Daniel and Callison-Burch, Chris and Eck, Douglas},
  journal={arXiv preprint arXiv:1911.00650},
  year={2019}
}

@article{hu2023radar,
  title={Radar: Robust ai-text detection via adversarial learning},
  author={Hu, Xiaomeng and Chen, Pin-Yu and Ho, Tsung-Yi},
  journal={Advances in Neural Information Processing Systems},
  volume={36},
  pages={15077--15095},
  year={2023}
}

@inproceedings{bao2024fast,
  title={Fast-detectgpt: Efficient zero-shot detection of machine-generated text via conditional probability curvature},
  author={Bao, Guangsheng and Zhao, Yanbin and Teng, Zhiyang and Yang, Linyi and Zhang, Yue},
  booktitle={International Conference on Learning Representations},
  volume={2024},
  pages={24814--24836},
  year={2024}
}

@article{tiangptzero,
  title={Gptzero update v1, January 2023a},
  author={Tian, Edward},
  journal={URL https://gptzero. substack. com/p/gptzero-update-v1},
year={2023}
}

@misc{openai2023gpt4,
  author       = {OpenAI},
  title        = {GPT-4 Research},
  year         = {2023},
  url          = {https://openai.com/index/gpt-4-research},
  note         = {Accessed: February 27, 2025}
}

@misc{nbc2024fakebiden,
  author       = {{NBC News}},
  title        = {Fake Joe Biden robocall tells New Hampshire Democrats not to vote Tuesday},
  year         = {2024},
  url          = {https://www.nbcnews.com/politics/2024-election/fake-joe-biden-robocall-tells-new-hampshire-democrats-not-vote-tuesday-rcna134984},
  note         = {Accessed: February 27, 2025}
}

@misc{nyt2023hairpin,
  author       = {{The New York Times}},
  title        = {The Hairpin blog has become an AI-generated spam content farm},
  year         = {2023},
  url          = {https://www.businessinsider.com/the-hairpin-blog-ai-spam-content-farm-cybersquatting-2024-1},
  note         = {Accessed: February 27, 2025}
}

@inproceedings{jakesch2023co,
  title={Co-writing with opinionated language models affects users’ views},
  author={Jakesch, Maurice and Bhat, Advait and Buschek, Daniel and Zalmanson, Lior and Naaman, Mor},
  booktitle={Proceedings of the 2023 CHI conference on human factors in computing systems},
  pages={1--15},
  year={2023}
}

@article{shumailov2023curse,
  title={The curse of recursion: Training on generated data makes models forget},
  author={Shumailov, Ilia and Shumaylov, Zakhar and Zhao, Yiren and Gal, Yarin and Papernot, Nicolas and Anderson, Ross},
  journal={arXiv preprint arXiv:2305.17493},
  year={2023}
}

@article{fagni2021tweepfake,
  title={TweepFake: About detecting deepfake tweets},
  author={Fagni, Tiziano and Falchi, Fabrizio and Gambini, Margherita and Martella, Antonio and Tesconi, Maurizio},
  journal={Plos one},
  volume={16},
  number={5},
  pages={e0251415},
  year={2021},
  publisher={Public Library of Science San Francisco, CA USA}
}

@article{wang2023m4,
  title={M4: Multi-generator, multi-domain, and multi-lingual black-box machine-generated text detection},
  author={Wang, Yuxia and Mansurov, Jonibek and Ivanov, Petar and Su, Jinyan and Shelmanov, Artem and Tsvigun, Akim and Whitehouse, Chenxi and Afzal, Osama Mohammed and Mahmoud, Tarek and Sasaki, Toru and others},
  journal={arXiv preprint arXiv:2305.14902},
  year={2023}
}

@article{dugan2024raid,
  title={Raid: A shared benchmark for robust evaluation of machine-generated text detectors},
  author={Dugan, Liam and Hwang, Alyssa and Trhlik, Filip and Ludan, Josh Magnus and Zhu, Andrew and Xu, Hainiu and Ippolito, Daphne and Callison-Burch, Chris},
  journal={arXiv preprint arXiv:2405.07940},
  year={2024}
}

@article{keskar2019ctrl,
  title={Ctrl: A conditional transformer language model for controllable generation},
  author={Keskar, Nitish Shirish and McCann, Bryan and Varshney, Lav R and Xiong, Caiming and Socher, Richard},
  journal={arXiv preprint arXiv:1909.05858},
  year={2019}
}

@inproceedings{devlin2019bert,
  title={Bert: Pre-training of deep bidirectional transformers for language understanding},
  author={Devlin, Jacob and Chang, Ming-Wei and Lee, Kenton and Toutanova, Kristina},
  booktitle={Proceedings of the 2019 conference of the North American chapter of the association for computational linguistics: human language technologies, volume 1 (long and short papers)},
  pages={4171--4186},
  year={2019}
}

@article{pillutla2021mauve,
  title={Mauve: Measuring the gap between neural text and human text using divergence frontiers},
  author={Pillutla, Krishna and Swayamdipta, Swabha and Zellers, Rowan and Thickstun, John and Welleck, Sean and Choi, Yejin and Harchaoui, Zaid},
  journal={Advances in Neural Information Processing Systems},
  volume={34},
  pages={4816--4828},
  year={2021}
}

@article{wu2023llmdet,
  title={LLMDet: A third party large language models generated text detection tool},
  author={Wu, Kangxi and Pang, Liang and Shen, Huawei and Cheng, Xueqi and Chua, Tat-Seng},
  journal={arXiv preprint arXiv:2305.15004},
  year={2023}
}

@misc{gptzero_perplexity_burstiness,
  author = {Edward Tian},
  title = {Perplexity, burstiness, and statistical AI detection},
  year = {2023},
  howpublished = {\url{https://gptzero.me/news/perplexity-and-burstiness-what-is-it/}},
  note = {Accessed: 2025-05-10}
}

@inproceedings{bhat2020effectively,
  title={How effectively can machines defend against machine-generated fake news? an empirical study},
  author={Bhat, Meghana Moorthy and Parthasarathy, Srinivasan},
  booktitle={Proceedings of the First Workshop on Insights from Negative Results in NLP},
  pages={48--53},
  year={2020}
}

@article{liang2023mutation,
  title={Mutation-based adversarial attacks on neural text detectors},
  author={Liang, Gongbo and Guerrero, Jesus and Alsmadi, Izzat},
  journal={arXiv preprint arXiv:2302.05794},
  year={2023}
}

@article{guerrero2022mutation,
  title={A Mutation-based Text Generation for Adversarial Machine Learning Applications},
  author={Guerrero, Jesus and Liang, Gongbo and Alsmadi, Izzat},
  journal={arXiv preprint arXiv:2212.11808},
  year={2022}
}

@article{liang2023gpt,
  title={GPT detectors are biased against non-native English writers},
  author={Liang, Weixin and Yuksekgonul, Mert and Mao, Yining and Wu, Eric and Zou, James},
  journal={Patterns},
  volume={4},
  number={7},
  year={2023},
  publisher={Elsevier}
}

@inproceedings{gagiano2021robustness,
  title={Robustness analysis of grover for machine-generated news detection},
  author={Gagiano, Rinaldo and Kim, Maria Myung-Hee and Zhang, Xiuzhen Jenny and Biggs, Jennifer},
  booktitle={Proceedings of the 19th Annual Workshop of the Australasian Language Technology Association},
  pages={119--127},
  year={2021}
}

@inproceedings{pu2023deepfake,
  title={Deepfake text detection: Limitations and opportunities},
  author={Pu, Jiameng and Sarwar, Zain and Abdullah, Sifat Muhammad and Rehman, Abdullah and Kim, Yoonjin and Bhattacharya, Parantapa and Javed, Mobin and Viswanath, Bimal},
  booktitle={2023 IEEE symposium on security and privacy (SP)},
  pages={1613--1630},
  year={2023},
  organization={IEEE}
}

@article{cai2023evade,
  title={Evade chatgpt detectors via a single space},
  author={Cai, Shuyang and Cui, Wanyun},
  journal={arXiv preprint arXiv:2307.02599},
  year={2023}
}

@article{wolff2020attacking,
  title={Attacking neural text detectors},
  author={Wolff, Max and Wolff, Stuart},
  journal={arXiv preprint arXiv:2002.11768},
  year={2020}
}

@article{ScienceAIUse2025,
  author  = {Brainard, Jeffrey},
  title   = {Far More Authors Use {AI} to Write Science Papers Than Admit It, Publisher Reports},
  journal = {Science},
  year    = {2025},
  month   = sep,
  url     = {https://www.science.org/content/article/far-more-authors-use-ai-write-science-papers-admit-it-publisher-reports}
}

@book{lattimore2020bandit,
  title={Bandit algorithms},
  author={Lattimore, Tor and Szepesv{\'a}ri, Csaba},
  year={2020},
  publisher={Cambridge University Press}
}

@misc{nyt2025falseai,
  author       = {{The New York Times}},
  title        = {AI ChatGPT and the Student Cheating Crackdown},
  year         = {2025},
  howpublished = {\url{https://www.nytimes.com/2025/05/17/style/ai-chatgpt-turnitin-students-cheating.html}},
  note         = {Accessed: 2025-10-28}
}

@misc{guardian2024aicheating,
  author       = {The Guardian},
  title        = {‘I received a first but it felt tainted and undeserved’: inside the university AI cheating crisis},
  year         = {2024},
  howpublished = {\url{https://www.theguardian.com/technology/2024/dec/15/i-received-a-first-but-it-felt-tainted-and-undeserved-inside-the-university-ai-cheating-crisis}},
  note         = {Accessed: 2025-10-28}
}

@article{guo2024detective,
  title={Detective: Detecting ai-generated text via multi-level contrastive learning},
  author={Guo, Xun and He, Yongxin and Zhang, Shan and Zhang, Ting and Feng, Wanquan and Huang, Haibin and Ma, Chongyang},
  journal={Advances in Neural Information Processing Systems},
  volume={37},
  pages={88320--88347},
  year={2024}
}

@inproceedings{li2024mage,
  title={MAGE: Machine-generated text detection in the wild},
  author={Li, Yafu and Li, Qintong and Cui, Leyang and Bi, Wei and Wang, Zhilin and Wang, Longyue and Yang, Linyi and Shi, Shuming and Zhang, Yue},
  booktitle={Proceedings of the 62nd Annual Meeting of the Association for Computational Linguistics (Volume 1: Long Papers)},
  pages={36--53},
  year={2024}
}

@article{brookings2024watermarking,
  title        = {Detecting AI fingerprints: A guide to watermarking and beyond},
  author       = {Siddarth Srinivasan},
  journal      = {Brookings Institution},
  year         = {2024},
  month        = {July},
  url          = {https://www.brookings.edu/articles/detecting-ai-fingerprints-a-guide-to-watermarking-and-beyond/},
  note         = {Accessed: 2025-11-08}
}

@article{you2024linear,
  title={When linear attention meets autoregressive decoding: Towards more effective and efficient linearized large language models},
  author={You, Haoran and Fu, Yichao and Wang, Zheng and Yazdanbakhsh, Amir and Lin, Yingyan Celine},
  journal={arXiv preprint arXiv:2406.07368},
  year={2024}
}

@article{brown2020gpt3,
  title     = {Language Models are Few-Shot Learners},
  author    = {Brown, Tom and Mann, Benjamin and Ryder, Nick and Subbiah, Melanie and Kaplan, Jared D. and Dhariwal, Prafulla and Neelakantan, Arvind and Shyam, Pranav and Sastry, Girish and Askell, Amanda and others},
  journal   = {Advances in Neural Information Processing Systems},
  volume    = {33},
  pages     = {1877--1901},
  year      = {2020}
}

@misc{verge2024openaiwatermark,
  author       = {{The Verge}},
  title        = {{OpenAI} won’t watermark {ChatGPT} text because its users could get caught},
  howpublished = {The Verge},
  year         = {2024},
  month        = aug,
  day          = {5},
  url          = {https://www.theverge.com/2024/8/4/24213268/openai-chatgpt-text-watermark-cheat-detection-tool},
  note         = {Accessed: 2025-12-19}
}

@misc{businessinsider2024homeworkcheating,
  author       = {{Business Insider}},
  title        = {Yup, {AI} is basically just a homework-cheating machine},
  howpublished = {Business Insider},
  year         = {2024},
  month        = aug,
  day          = {6},
  url          = {https://www.businessinsider.com/ai-chatgpt-homework-cheating-machine-sam-altman-openai-2024-8},
  note         = {Accessed: 2025-12-19}
}

@article{zhu2025reliably,
  title={Reliably Bounding False Positives: A Zero-Shot Machine-Generated Text Detection Framework via Multiscaled Conformal Prediction},
  author={Zhu, Xiaowei and Ren, Yubing and Cao, Yanan and Lin, Xixun and Fang, Fang and Li, Yangxi},
  journal={arXiv preprint arXiv:2505.05084},
  year={2025}
}

@inproceedings{yang2024dna,
  title={Dna-gpt: Divergent n-gram analysis for training-free detection of gpt-generated text},
  author={Yang, Xianjun and Cheng, Wei and Wu, Yue and Petzold, Linda and Wang, William and Chen, Haifeng},
  booktitle={International Conference on Learning Representations},
  volume={2024},
  pages={48572--48597},
  year={2024}
}

@article{hinton2015distilling,
  title        = {Distilling the Knowledge in a Neural Network},
  author       = {Hinton, Geoffrey and Vinyals, Oriol and Dean, Jeff},
  journal      = {arXiv preprint arXiv:1503.02531},
  year         = {2015},
  url          = {https://arxiv.org/abs/1503.02531}
}

@inproceedings{holtzman2020curious,
  title        = {The Curious Case of Neural Text Degeneration},
  author       = {Holtzman, Ari and Buys, Jan and Du, Li and Forbes, Maxwell and Choi, Yejin},
  booktitle    = {International Conference on Learning Representations (ICLR)},
  year         = {2020},
  url          = {https://openreview.net/forum?id=rygGQyrFvH}
}

@InProceedings{li2025likelihood,
  title     = {A Likelihood Based Approach for Watermark Detection},
  author    = {Li, Xingchi and Li, Guanxun and Zhang, Xianyang},
  booktitle = {Proceedings of The 28th International Conference on Artificial Intelligence and Statistics},
  pages     = {1675--1683},
  year      = {2025},
  editor    = {Li, Yingzhen and Mandt, Stephan and Agrawal, Shipra and Khan, Emtiyaz},
  volume    = {258},
  series    = {Proceedings of Machine Learning Research},
  month     = {03--05 May},
  publisher = {PMLR},
  url       = {https://proceedings.mlr.press/v258/li25d.html}
}

@inproceedings{gloaguen2025blackbox,
  title     = {Black-Box Detection of Language Model Watermarks},
  author    = {Gloaguen, Thibaud and Jovanovi\'{c}, Nikola and Staab, Robin and Vechev, Martin T.},
  booktitle = {International Conference on Learning Representations},
  year      = {2025},
  url       = {https://openreview.net/forum?id=E4LAVLXAHW}
}

@inproceedings{zhao2025watermarkedsegments,
  title     = {Efficiently Identifying Watermarked Segments in Mixed-Source Texts},
  author    = {Zhao, Xuandong and Liao, Chenwen and Wang, Yu-Xiang and Li, Lei},
  booktitle = {Proceedings of the 63rd Annual Meeting of the Association for Computational Linguistics (Volume 1: Long Papers)},
  year      = {2025},
  month     = jul,
  address   = {Vienna, Austria},
  publisher = {Association for Computational Linguistics},
  pages     = {6304--6316},
  doi       = {10.18653/v1/2025.acl-long.316},
  url       = {https://aclanthology.org/2025.acl-long.316/}
}

@InProceedings{chen2025online,
  title     = {Online Detection of {LLM}-Generated Texts via Sequential Hypothesis Testing by Betting},
  author    = {Chen, Can and Wang, Jun-Kun},
  booktitle = {Proceedings of the 42nd International Conference on Machine Learning},
  pages     = {9231--9276},
  year      = {2025},
  editor    = {Singh, Aarti and Fazel, Maryam and Hsu, Daniel and Lacoste-Julien, Simon and Berkenkamp, Felix and Maharaj, Tegan and Wagstaff, Kiri and Zhu, Jerry},
  volume    = {267},
  series    = {Proceedings of Machine Learning Research},
  month     = {13--19 Jul},
  publisher = {PMLR},
  url       = {https://proceedings.mlr.press/v267/chen25bn.html}
}

@inproceedings{luo-etal-2023-came,
    title = "{CAME}: Confidence-guided Adaptive Memory Efficient Optimization",
    author = "Luo, Yang  and
      Ren, Xiaozhe  and
      Zheng, Zangwei  and
      Jiang, Zhuo  and
      Jiang, Xin  and
      You, Yang",
    editor = "Rogers, Anna  and
      Boyd-Graber, Jordan  and
      Okazaki, Naoaki",
    booktitle = "Proceedings of the 61st Annual Meeting of the Association for Computational Linguistics (Volume 1: Long Papers)",
    month = jul,
    year = "2023",
    address = "Toronto, Canada",
    publisher = "Association for Computational Linguistics",
    url = "https://aclanthology.org/2023.acl-long.243/",
    doi = "10.18653/v1/2023.acl-long.243",
    pages = "4442--4453"
}

@misc{luoCAMEgithub2023,
  author       = {Luo, Yang and Ren, Xiaozhe and Zheng, Zangwei and Jiang, Zhuo and Jiang, Xin and You, Yang},
  title        = {CAME Optimizer: Official Implementation},
  year         = {2023},
  howpublished = {\url{https://github.com/yangluo7/CAME}},
  note         = {Accessed: 2026-01-18}
}

@article{LucaZervas2016FakeIt,
  author  = {Luca, Michael and Zervas, Georgios},
  title   = {Fake It Till You Make It: Reputation, Competition, and Yelp Review Fraud},
  journal = {Management Science},
  year    = {2016},
  volume  = {62},
  number  = {12},
  pages   = {3412--3427},
  doi     = {10.1287/mnsc.2015.2304},
  publisher = {INFORMS}
}

@article{Dellarocas2006StrategicManipulation,
  author  = {Dellarocas, Chrysanthos},
  title   = {Strategic Manipulation of Internet Opinion Forums: Implications for Consumers and Firms},
  journal = {Management Science},
  year    = {2006},
  volume  = {52},
  number  = {10},
  pages   = {1577--1593},
  doi     = {10.1287/mnsc.1060.0567},
  publisher = {INFORMS}
}

@article{Tambe2026Reskilling,
  author  = {Tambe, Prasanna B.},
  title   = {Reskilling the Workforce for AI: Domain Expertise and Algorithmic Literacy},
  journal = {Management Science},
  year    = {2026},
  volume  = {72},
  number  = {1},
  pages   = {515--537},
  doi     = {10.1287/mnsc.2022.03968},
  note    = {Published online September 30, 2025},
  publisher = {INFORMS}
}

@article{FugenerWalznerGupta2026Roles,
  author  = {F{\"u}gener, Andreas and Walzner, Dominik D. and Gupta, Alok},
  title   = {Roles of Artificial Intelligence in Collaboration with Humans: Automation, Augmentation, and the Future of Work},
  journal = {Management Science},
  year    = {2026},
  volume  = {72},
  number  = {1},
  pages   = {538--557},
  doi     = {10.1287/mnsc.2024.05684},
  note    = {Published online October 15, 2025},
  publisher = {INFORMS}
}

@article{BastaniEtAl2025Guardrails,
  author  = {Bastani, Hamsa and Bastani, Osbert and Sungu, Alp and Ge, Haosen and Kabak{\c{c}}{\i}, {\"O}zge and Mariman, Rei},
  title   = {Generative AI without guardrails can harm learning: Evidence from high school mathematics},
  journal = {Proceedings of the National Academy of Sciences},
  year    = {2025},
  volume  = {122},
  number  = {26},
  pages   = {e2422633122},
  doi     = {10.1073/pnas.2422633122}
}

@techreport{BastaniCachon2025ContractingParadox,
  author      = {Bastani, Hamsa and Cachon, Gerard P.},
  title       = {The Human-AI Contracting Paradox},
  institution = {SSRN},
  year        = {2025},
  type        = {Working Paper},
  number      = {5962739},
  doi         = {10.2139/ssrn.5962739},
  url         = {https://ssrn.com/abstract=5962739},
  note        = {Posted December 2025}
}

@techreport{PoulidisEtAl2025ActionAttention,
  author      = {Poulidis, Stefanos and Ge, Haosen and Bastani, Hamsa and Bastani, Osbert},
  title       = {Action vs. Attention Signals for Human-AI Collaboration: Evidence from Chess},
  institution = {SSRN},
  year        = {2025},
  type        = {Working Paper},
  number      = {5128584},
  doi         = {10.2139/ssrn.5128584},
  url         = {https://ssrn.com/abstract=5128584},
  note        = {The Wharton School Research Paper}
}

@article{ChenChao2020,
  title={Data-based dynamic pricing and inventory control with censored demand and limited price changes},
  author={Chen, Boxiao and Chao, Xiuli and Wang, Yining},
  journal={Operations Research},
  volume={68},
  number={5},
  pages={1445--1456},
  year={2020},
  publisher={INFORMS}
}

@article{jin2024sample,
  title={Sample complexity of posted pricing for a single item},
  author={Jin, Billy and Kesselheim, Thomas and Ma, Will and Singla, Sahil},
  journal={Advances in Neural Information Processing Systems},
  volume={37},
  pages={82296--82317},
  year={2024}
}

@inproceedings{baek2023ts,
  title={TS-UCB: Improving on Thompson sampling with little to no additional computation},
  author={Baek, Jackie and Farias, Vivek},
  booktitle={International Conference on Artificial Intelligence and Statistics},
  pages={11132--11148},
  year={2023},
  organization={PMLR}
}

@inproceedings{kalvit2023complexity,
  title={Complexity analysis of a countable-armed bandit problem},
  author={Kalvit, Anand and Zeevi, Assaf},
  booktitle={International Conference on Algorithmic Learning Theory},
  pages={850--890},
  year={2023},
  organization={PMLR}
}

@article{Alizamir2022SearchPressure,
  author  = {Alizamir, Saed and de V{\'e}ricourt, Francis and Sun, Peng},
  title   = {Search Under Accumulated Pressure},
  journal = {Operations Research},
  year    = {2022},
  volume  = {70},
  number  = {3},
  pages   = {1393--1409},
  doi     = {10.1287/opre.2019.1880}
}

@article{el2019generalization,
  title={Generalization bounds in the predict-then-optimize framework},
  author={El Balghiti, Othman and Elmachtoub, Adam N and Grigas, Paul and Tewari, Ambuj},
  journal={Advances in neural information processing systems},
  volume={32},
  year={2019}
}

@article{JohariKambleKanoria2021Matching,
  author  = {Johari, Ramesh and Kamble, Vijay and Kanoria, Yash},
  title   = {Matching While Learning},
  journal = {Operations Research},
  year    = {2021},
  volume  = {69},
  number  = {2},
  pages   = {655--681},
  doi     = {10.1287/opre.2020.2013}
}

@article{chakraborty2023possibilities,
  title={On the possibilities of ai-generated text detection},
  author={Chakraborty, Souradip and Bedi, Amrit Singh and Zhu, Sicheng and An, Bang and Manocha, Dinesh and Huang, Furong},
  journal={arXiv preprint arXiv:2304.04736},
  year={2023}
}

@article{munoz-ortiz-etal-2024-contrasting,
  title = {Contrasting Linguistic Patterns in Human and {LLM}-Generated News Text},
  author = {Mu{\~n}oz-Ortiz, Alberto and G{\'o}mez-Rodr{\'i}guez, Carlos and Vilares, David},
  journal = {Artificial Intelligence Review},
  year = {2024},
  volume = {57},
  pages = {265},
  month = aug,
  doi = {10.1007/s10462-024-10903-2},
  url = {https://doi.org/10.1007/s10462-024-10903-2},
  publisher = {Springer}
}

@inproceedings{zamaraeva-etal-2025-comparing,
  title = {Comparing {LLM}-generated and human-authored news text using formal syntactic theory},
  author = {Zamaraeva, Olga  and Flickinger, Dan  and Bond, Francis  and G{\'o}mez-Rodr{\'i}guez, Carlos},
  editor = {Che, Wanxiang  and Nabende, Joyce  and Shutova, Ekaterina  and Pilehvar, Mohammad Taher},
  booktitle = {Proceedings of the 63rd Annual Meeting of the Association for Computational Linguistics (Volume 1: Long Papers)},
  month = jul,
  year = {2025},
  address = {Vienna, Austria},
  publisher = {Association for Computational Linguistics},
  url = {https://aclanthology.org/2025.acl-long.443/},
  doi = {10.18653/v1/2025.acl-long.443},
  pages = {9041--9060},
  ISBN = {979-8-89176-251-0}
}

@techreport{ieee7542019,
  author      = {{IEEE}},
  title       = {{IEEE Standard for Floating-Point Arithmetic}},
  institution = {Institute of Electrical and Electronics Engineers},
  type        = {{IEEE Std}},
  number      = {754-2019},
  year        = {2019},
  pages       = {1--84},
  doi         = {10.1109/IEEESTD.2019.8766229},
  url         = {https://ieeexplore.ieee.org/document/8766229}
}

@article{zhang2026regurgitative,
  title={Regurgitative training: The value of real data in training large language models},
  author={Zhang, Jinghui and Yang, Mochen and Qiao, Dandan and Wei, Qiang},
  journal={Management Science},
  year={2026},
  publisher={INFORMS}
}

@article{liu2024adaptive,
  title={Adaptive text watermark for large language models},
  author={Liu, Yepeng and Bu, Yuheng},
  journal={arXiv preprint arXiv:2401.13927},
  year={2024}
}

@article{dathathri2024scalable,
  title={Scalable watermarking for identifying large language model outputs},
  author={Dathathri, Sumanth and See, Abigail and Ghaisas, Sumedh and Huang, Po-Sen and McAdam, Rob and Welbl, Johannes and Bachani, Vandana and Kaskasoli, Alex and Stanforth, Robert and Matejovicova, Tatiana and others},
  journal={Nature},
  volume={634},
  number={8035},
  pages={818--823},
  year={2024},
  publisher={Nature Publishing Group UK London}
}

@article{li2025statistical,
  title={A statistical framework of watermarks for large language models: Pivot, detection efficiency and optimal rules},
  author={Li, Xiang and Ruan, Feng and Wang, Huiyuan and Long, Qi and Su, Weijie J},
  journal={Annals of statistics},
  volume={53},
  number={1},
  pages={322},
  year={2025}
}

@inproceedings{he2024mgtbench,
  title={Mgtbench: Benchmarking machine-generated text detection},
  author={He, Xinlei and Shen, Xinyue and Chen, Zeyuan and Backes, Michael and Zhang, Yang},
  booktitle={Proceedings of the 2024 on ACM SIGSAC Conference on Computer and Communications Security},
  pages={2251--2265},
  year={2024}
}

@article{raj2026artificial,
  title={The artificial intelligence disclosure penalty: Humans persistently devalue AI-generated creative writing.},
  author={Raj, Manav and Berg, Justin M and Seamans, Rob},
  journal={Journal of Experimental Psychology: General},
  year={2026},
  publisher={American Psychological Association}
}

@article{newyorktimes2026aifiction,
  author  = {{The New York Times}},
  title   = {{A.I.} Is Writing Fiction. Publishers Are Unprepared},
  journal = {The New York Times},
  year    = {2026},
  month   = March,
  day     = {19},
  url     = {https://www.nytimes.com/2026/03/19/books/ai-fiction-shy-girl.html}
}

@article{guardian2026publishers,
  author  = {The {Guardian}},
  title   = {{`Soon Publishers Won't Stand a Chance': Literary World
             in Struggle to Detect AI-Written Books}},
  journal = {The Guardian},
  year    = {2026},
  month   = March,
  day     = {29},
  url     = {https://www.theguardian.com/technology/2026/mar/29/ai-written-books-novel-shy-girl-publishers}
}

@misc{neurips2026aipapers,
  author       = {{NeurIPS}},
  title        = {{AI}-Generated Papers in the {NeurIPS} 2026
                  Position Paper Track},
  howpublished = {NeurIPS Blog},
  year         = {2026},
  month        = June,
  day          = {2},
  url          = {https://blog.neurips.cc/2026/06/02/ai-generated-papers-in-the-neurips-2026-position-paper-track/}
}

@article{csmonitor2026publishing,
  author  = {{The Monitor}},
  title   = {Book Publishers Face Rising Scrutiny over {AI} Usage by Authors},
  journal = {The Christian Science Monitor},
  year    = {2026},
  month   = August,
  day     = {13},
  url     = {https://www.csmonitor.com/Arts-Culture/Books/2026/0812/ai-publishing-commonwealth-prize-shy-girl}
}

@misc{commonwealth2026update,
  author       = {{Commonwealth Foundation}},
  title        = {2026 Commonwealth Short Story Prize Update},
  howpublished = {Commonwealth Foundation},
  year         = {2026},
  month        = June,
  day          = {22},
  note         = {Statement by Razmi Farook, Director-General},
  url          = {https://commonwealthfoundation.com/2026-cw-prize-update/}
}

@misc{authorsguild2026detectors,
  author       = {{Authors Guild}},
  title        = {Can {AI} Detectors Be Trusted? The Authors Guild
                  Put Five of Them to the Test},
  howpublished = {Authors Guild},
  year         = {2026},
  month        = may,
  day          = {26},
  url          = {https://authorsguild.org/news/can-ai-detectors-be-trusted/}
}

@misc{newby2026,
  author = {{Supreme Court of the State of New York}},
  title  = {{Matter of Newby v. Adelphi University}},
  year   = {2026},
  month  = jan,
  day    = {28},
  note   = {2026 NY Slip Op 26021},
  url    = {https://law.justia.com/cases/new-york/other-courts/2026/2026-ny-slip-op-26021.html}
}

@techreport{jabarian2025artificial,
  author      = {Jabarian, Brian and Imas, Alex},
  title       = {Artificial Writing and Automated Detection},
  institution = {National Bureau of Economic Research},
  type        = {NBER Working Paper},
  number      = {34223},
  year        = {2025},
  month       = sep,
  doi         = {10.3386/w34223},
  url         = {https://doi.org/10.3386/w34223}
}

@misc{vanderbilt2023turnitin,
  author       = {{Vanderbilt University}},
  title        = {Guidance on {AI} Detection and Why We're Disabling
                  Turnitin's {AI} Detector},
  howpublished = {Brightspace},
  year         = {2023},
  month        = aug,
  day          = {16},
  url          = {https://www.vanderbilt.edu/brightspace/2023/08/16/guidance-on-ai-detection-and-why-were-disabling-turnitins-ai-detector/}
}

@misc{pitt2026academicintegrity,
  author       = {{University of Pittsburgh Center for Teaching and Learning}},
  title        = {Encouraging Academic Integrity},
  howpublished = {University Center for Teaching and Learning},
  year         = {2026},
  note         = {Accessed August 17, 2026},
  url          = {https://teaching.pitt.edu/resources/encouraging-academic-integrity/}
}

@misc{jhu2025detection,
  author       = {{Johns Hopkins University}},
  title        = {Detection Tools: Limitations and Alternatives},
  howpublished = {Teaching at Johns Hopkins University},
  year         = {2025},
  month        = mar,
  day          = {6},
  url          = {https://teaching.jhu.edu/university-teaching-policies/generative-ai/detection-tools/}
}

@article{silman2026publishing,
  author  = {Silman, Anna},
  title   = {AI Has Plunged the Book Publishing Industry Into Utter Chaos},
  journal = {The Wall Street Journal},
  year    = {2026},
  month   = aug,
  day     = {17},
  url     = {https://www.wsj.com/arts-culture/books/generative-ai-book-publishing-be79a287}
}

@article{romano2005optimal,
  title={Optimal testing of equivalence hypotheses},
  author={Romano, Joseph P},
  journal={Annals of statistics},
  pages={1036--1047},
  year={2005},
  publisher={JSTOR}
}

@article{nelson1995using,
  title={Using common random numbers for indifference-zone selection and multiple comparisons in simulation},
  author={Nelson, Barry L and Matejcik, Frank J},
  journal={Management Science},
  volume={41},
  number={12},
  pages={1935--1945},
  year={1995},
  publisher={INFORMS}
}

@article{frazier2014fully,
  title={A fully sequential elimination procedure for indifference-zone ranking and selection with tight bounds on probability of correct selection},
  author={Frazier, Peter I},
  journal={Operations Research},
  volume={62},
  number={4},
  pages={926--942},
  year={2014},
  publisher={INFORMS}
}

@article{balakrishnan2019hypothesis,
  title={Hypothesis testing for densities and high-dimensional multinomials},
  author={Balakrishnan, Sivaraman and Wasserman, Larry},
  journal={The Annals of Statistics},
  volume={47},
  number={4},
  pages={1893--1927},
  year={2019},
  publisher={JSTOR}
}
% For SSRN:
%\bibliography{citations_full}

\setlength{\abovedisplayskip}{4pt}
\setlength{\belowdisplayskip}{4pt}
\setlength{\abovedisplayshortskip}{2pt}
\setlength{\belowdisplayshortskip}{2pt}

%\fontsize{4}{4}\selectfont
\newpage
\setcounter{page}{1}
\fontsize{10}{12}\selectfont

\section*{\centering Online Appendices}%\vspace{0.3cm}\label{sec:Appendix}
\begin{APPENDICES}
\vspace{5mm}

\setlength{\abovedisplayskip}{4pt}
\setlength{\belowdisplayskip}{4pt}
\setlength{\abovedisplayshortskip}{2pt}
\setlength{\belowdisplayshortskip}{2pt}

%\fontsize{8}{8}\selectfont

The online appendix is organized as follows. Appendix~\ref{sec:blackbox} extends the proposed tests to the black-box setting, where the evaluator's conditional probabilities are approximated through sampling. Appendices~\ref{app:concentrationinequalroadmap}--\ref{appendixlowerbound} contain the proofs: Appendix~\ref{app:concentrationinequalroadmap} presents the proof for Theorem \ref{thm:maindiff}, Appendix~\ref{app:stattestroadmap} presents the proofs for the white-box statistical tests, Appendix~\ref{app:blackboxproofsroadmap} presents the proof for the statistical tests in the black-box with sampling setting, and Appendix~\ref{appendixlowerbound} presents the proof for the lower bound. Appendix \ref{app:clipped_log_scores} develops clipped-score analogs of our attribution and detection tests, establishing finite-sample concentration and error guarantees without requiring Assumption~\ref{assum:crossmodel}. Appendix~\ref{app:experimentdetails} provides experiment implementation details and additional experiment results, including counterparts for white-box setting experiments in a black-box with sampling setting in \ref{BBexperiments}, dataset descriptions in \ref{app:datasetdetails}, setting and details for decoding strategies and adversarial attack experiments in \ref{app:adversarial attacks}, descriptions of the benchmark detectors in \ref{app:detectordetails}, hardware descriptions in \ref{app:hardwaredetails}, and code access and run instructions in \ref{app:ourasset}. %Appendix~\ref{app:extralitrev} discusses additional related literature on asymptotic detection methods and authorship attribution.

\section{Black-box detection via sampling }\label{sec:blackbox}

A few closed-weight language models, including Google Gemini (API/Bard/Vertex AI) and Anthropic Claude (public API \& Claude.ai), do not provide log probabilities for the evaluator model $A$. We call this the black-box setting. In black-box detection, we can empirically estimate the probability distributions $p^A_n$ by sampling from $A$ given a context. In this section, we adapt our tests to empirical estimates of the sequential probabilities obtained by finite sampling from the model. Recall that $p^A_n$ denotes the next-token probability mass function (pmf) for model $A$, given context $(\textbf{X}, Y_1, Y_2, \dots, Y_{n-1})$. For a language model $A$ that enables API access, given context $(\textbf{X}, Y_1, Y_2, \dots, Y_{n-1})$, we can sample from $A$ even when the LLM provider does not allow the API command to retrieve the probability distributions from $A$. Formally, for each $n^{\text{th}}$ token, we first obtain $m$ independent and identically distributed (i.i.d.) samples $\hat{Y}^{(1)}_n,\ldots,\hat{Y}^{(m)}_n \sim p^A_n(\cdot),$ and define the empirical pmf as
\[
\hat{p}^A_n(y):=\frac{\sum_{j=1}^{m}\mathbf{1}\{\hat{Y}^{(j)}_n=y\}}{m}.
\]
Additionally, define the empirical log-perplexity 
\begin{equation}\label{eq:bbstat-nosmooth}
\hat{l}_A(\mathbf{Y}_N)
~:=~ -\frac{1}{N}\sum_{n=1}^{N}\log \hat{p}^A_n(Y_n)
\end{equation}
as an approximation of the true log-perplexity $l_A(\mathbf{Y}_N)=-\frac{1}{N}\sum_{n=1}^{N}\log p^A_n(Y_n).$
Lastly, define the sample average entropy as
\begin{eqnarray*}
    \hat{h}_N(A,A)(\textbf{Y}_N) = -\frac{1}{N} \sum_{n=1}^{N} \sum_{y_n \in \mathcal{X}} \hat{p}^{A}_n(y_n) \log(\hat{p}^{A}_n(y_n)).
\end{eqnarray*}

\subsection{Attribution among multiple LLMs with sampling}\label{sec:multibb} %\vspace{-2mm}

Given a string $\mathbf{Y}_N$, we design a statistical test to detect whether one of the models in $\mathcal{A}=\{A_1, \ldots, A_p\}$ (non-sanctioned) or one of the models in $\mathcal{B}=\{B_1, \ldots, B_q\}$ (in-house) generated the text. The null hypothesis $\mathbf{H}_0$ is that the text $\mathbf{Y}_N$ is generated by a model in $\mathcal{B}$. We first calculate the sample random variables $\hat{Z}^{M}_n=:-\log\big(\hat{p}^{M}_n(Y_n)\big)$, and sum over $\hat{Z}^{M}_n=:-\log\big(\hat{p}^{M}_n(Y_n)\big)$ for all models $M \in \mathcal{A}\cup \mathcal{B}$. Our test rejects the null $\mathbf{H}_0$ if for some $A_i \in \mathcal{A}$, we have
\begin{eqnarray*}
\hat{l}_{A_i}(\mathbf{Y}_N) < \hat{l}_{B_j}(\mathbf{Y}_N) \qquad \forall B_j \in \mathcal{B}.
\end{eqnarray*}
\begin{proposition}\label{thm:statstestmultibb}
    \small For any sample size 
$m$ such that  $m \;\ge\; \frac{1}{2\varepsilon^2}\log\!\left(\frac{2N}{\delta}\right)$, where $\varepsilon
=
\min\left(
\frac{\epsilon_1\epsilon}{16},
\frac{\epsilon_1}{16K\log K},
\frac{\epsilon_1^2}{64K},
\frac{1}{2K},
\frac{\epsilon}{4}
\right),$ 
    if Assumptions \ref{assum:crossmodel} and \ref{assum:mindiff2m} hold, the Type I error for our test is upper bounded by 
\begin{eqnarray*}
2|\mathcal{A}| \exp \left[ -\frac{N (\epsilon_1/4)}{-c_2\log(\epsilon)} 
\min \left(1, \frac{\epsilon_1/4}{-c_2\log(\epsilon)} \right) \right] 
+ 2 \exp \left[ -\frac{N (\epsilon_1/4)}{c_1\log(K)} 
\min \left(1, \frac{\epsilon_1/4}{c_1\log(K)} \right) \right]+(|\mathcal{A}|+|\mathcal{B}|)\delta
\end{eqnarray*}
and the Type II error for our test is upper bounded by
\begin{eqnarray*}
2|\mathcal{B}| \exp \left[ 
  -\frac{N (\epsilon_1/4)}{-c_2\log(\epsilon)} 
  \min \left(1, \frac{\epsilon_1/4}{-c_2\log(\epsilon)} \right) 
\right]
+ 2 \exp \left[ 
  -\frac{N \epsilon_1/4}{c_1\log(K)} 
  \min \left(1, \frac{\epsilon_1/4}{c_1\log(K)} \right) 
\right]+(|\mathcal{A}|+|\mathcal{B}|)\delta
\end{eqnarray*}
 with constants $c_1$, $c_2$, and $\epsilon$ as introduced in Theorem \ref{thm:maindiff}, and \(\delta\in(0,1)\) is a user-chosen sampling failure probability/confidence parameter that controls the event that the empirical estimates \(\hat p_n^M\) uniformly approximate the true probabilities \(p_n^M\) over tokens \(n=1,\ldots,N\)\footnote{In practice, \(\delta\) is set as a small tolerance for Monte Carlo estimation error: it upper bounds the probability that sampling noise makes at least one of the empirical next-token distributions \(\hat p_n^M\) inaccurate enough to affect the test decision, which is why it appears both in the required sampling budget through \(\log(1/\delta)\) and as the additive term \((|\mathcal{A}|+|\mathcal{B}|)\delta\) after union bounding over models. A standard calibration is to allocate a small portion \(\eta\) of the overall error budget to sampling by choosing \(\delta=\eta/(|\mathcal{A}|+|\mathcal{B}|)\), ensuring \((|\mathcal{A}|+|\mathcal{B}|)\delta\le \eta\). Since \(m\) depends only logarithmically on \(1/\delta\), tightening \(\delta\) by orders of magnitude increases the sampling cost only modestly.}.

\end{proposition} 
\proof{Proof.}
See Section \ref{app:proofofblackbox1}.   
\endproof
\hfill
\Halmos

Replacing the true next-token probabilities \(p_n^M\) by empirical estimates \(\hat p_n^M\) obtained from \(m\) samples per token introduces an additional source of uncertainty beyond the randomness in the observed text \(\mathbf{Y}_N\). Proposition \ref{thm:statstestmultibb} shows that if \(m \ge \frac{1}{2\varepsilon^2}\log\!\big(\frac{2N}{\delta}\big)\), where $\varepsilon
=
\min\left(
\frac{\epsilon_1\epsilon}{16},
\frac{\epsilon_1}{16K\log K},
\frac{\epsilon_1^2}{64K},
\frac{1}{2K},
\frac{\epsilon}{4}
\right)$, then the Type I and Type II errors still decay exponentially in \(N\). Relative to the white-box setting, two black-box effects appear: the effective margin in the exponential terms is reduced to \(\epsilon_1/4\) from the \(\epsilon_1/2\) in the white-box setting, and an additional additive term \((|\mathcal{A}|+|\mathcal{B}|)\delta\) reflects the probability that at least one sampling-based probability estimate is insufficiently accurate. Thus, the sampling precision parameter \(\delta\) contributes to both Type I and Type II error bounds through this additive term, while also determining the required sample size \(m\).

The required sample size \(m\) depends only logarithmically on the text length \(N\) and on the confidence level through \(\log(1/\delta)\), but scales as \(1/\varepsilon^2\). Consequently, tightening the sampling precision (smaller \(\delta\)) increases \(m\) slowly: for instance, when the \(\log(1/\delta)\) term is dominant, changing \(\delta\) from \(10^{-10}\) to \(10^{-20}\) increases \(\log(1/\delta)\) by a factor of \(2\), and hence approximately doubles the required \(m\). In contrast, because \(\varepsilon\) can be limited by the alphabet size \(K\) and the minimum separation margin \(\epsilon_1\) from Assumption \ref{assum:mindiff2m}, the sampling requirement can introduce polynomial dependence on \(K\) and inverse powers of \(\epsilon_1\), unlike the no-sampling setting where these quantities primarily enter through logarithmic factors in the concentration terms. Achieving a target accuracy therefore requires balancing the text length \(N\) (which governs the exponential decay terms) with the sampling effort \(m\) and the precision parameter \(\delta\). Appendix~\ref{BBexperiments} shows that, empirically, $m \leq 1000$ suffices in many settings.

\noindent\textbf{Managerial implications.} Relative to Proposition~\ref{thm:statstestmulti}, Proposition~\ref{thm:statstestmultibb} introduces a sampling-specific design decision because the manager does not observe the candidate models' next-token probabilities and must estimate them empirically. The manager therefore has three main levers: (i) the sample size \(m\) used to construct each empirical next-token distribution; (ii) the minimum text length \(N\); and (iii) the numbers of models included in the non-sanctioned set \(\mathcal{A}\) and the sanctioned set \(\mathcal{B}\). The first lever determines the reliability of the empirical distributions. The parameter \(\delta\) represents the probability of insufficiently accurate estimation for a candidate model, and the resulting increase in each error bound is \((|\mathcal{A}|+|\mathcal{B}|)\delta\). Thus, the manager can first specify how much additional error she is willing to allow because log-probabilities are unavailable and must then select \(\delta\) accordingly. Proposition~\ref{thm:statstestmultibb} provides a sufficient value of \(m\) for that choice. A smaller \(\delta\) limits the additional error more tightly but requires more samples. Holding the remaining parameters fixed, the required \(m\) increases only logarithmically with \(N\) and with \(1/\delta\). Hence, doubling the text length or imposing a substantially stricter sampling tolerance does not require a proportional increase in the number of samples used to construct each empirical distribution. The second lever is the minimum text length \(N\). Once \(m\) satisfies the sampling requirement, the non-sampling components of both error bounds decrease exponentially with \(N\), while the additional sampling term remains fixed for a given \(\delta\). The manager can therefore determine the minimum length of an employee report, student assignment, or other audited document. The third lever is the number of models included in each set. As in Proposition~\ref{thm:statstestmulti}, reducing \(|\mathcal{A}|\) lowers one component of the Type~I error bound, while reducing \(|\mathcal{B}|\) lowers one component of the Type~II error bound. In the sampling setting, reducing either set also lowers the additional sampling term and the total number of samples required. If the manager retains larger model sets while keeping the permitted sampling-related increase in the error bounds unchanged, she must select a smaller \(\delta\); the resulting sufficient \(m\) increases only logarithmically with the total number of models. Finally, \(K\) and \(\epsilon\) enter the concentration terms through \(\log K\) and \(\log(1/\epsilon)\); consequently, their effects on these terms grow only logarithmically as new models adopt larger vocabularies or finer numerical precision.

\subsection{Model $A$ or not model $A$ with sampling?}
Given a string $\mathbf{Y}_N$, for arbitrary constants $t$, we design a statistical test to detect whether evaluator model $A$ that we do not have access to its log probabilities, generated the text. The null hypothesis $\mathbf{H}_0$ is that the text $\mathbf{Y}_N$ is not generated by the evaluator model $A$; it is generated by any model $B$ that we do not have white-box access to, which could be a human or some other language generation process. The alternative $\mathbf{H}$ is that it is generated by $A$. We first calculate the random variable $Z_n^{A}=:-\log\big(p^{A}_n(Y_n)\big)$, and then we calculate $\frac{1}{N}\sum_{n=1}^{N} Z_n^{A}$. Our test rejects the null if 
\begin{eqnarray*}
   \left| \hat{l}_A(\mathbf{Y}_N) - \hat{h}_N(A,A)(\mathbf{Y}_N) \right| \le t.
\end{eqnarray*}

\begin{proposition}\label{thm:statstestsampling}

For any $t>0$, and any sample size
$m$ such that $m \ge \frac{1}{2\varepsilon^2}
\log\left(\frac{2N}{\delta}\right)$, where
$\varepsilon
=
\min\left(
\frac{t\epsilon}{16},
\frac{t}{16K\log K},
\frac{t^2}{64K},
\frac{1}{2K},
\frac{\epsilon}{4}
\right).,$ the Type II error for our test is upper bounded by
$$2 \exp \left[
-\frac{N(t/2)}{c_1\log(K)}
\min\left(1,\frac{t/2}{c_1\log(K)}\right)
\right]+\delta$$
with $c_1$ as introduced in Theorem \ref{thm:maindiff}. Also, if Assumptions \ref{assum:crossmodel} and \ref{assum:mindiff} hold, then for any positive $t < \epsilon_2/4$, and for $c_2$ as introduced in Theorem \ref{thm:maindiff}, the Type  I error of our test is upper bounded by  

$$2 \exp \left[ -\frac{N \epsilon_2/4}{-c_2\log(\epsilon)} \min \left(1, \frac{\epsilon_2/4}{-c_2\log(\epsilon)} \right) \right]
    + 2 \exp \left[ -\frac{N (\epsilon_2/4 - t)}{-c_2\log(\epsilon)} \min \left(1, \frac{\epsilon_2/4 - t}{-c_2\log(\epsilon)} \right) \right]+\delta,$$
where $\delta\in(0,1)$ is a user-chosen sampling failure probability that controls the event that the empirical estimates $\hat p_n^M$ uniformly approximate the true probabilities $p_n^M$ over tokens $n=1,\ldots,N$. The same Type~I bound holds when the sufficient condition in Equation~\ref{eq:sufficient} is imposed in place of Assumption~\ref{assum:mindiff}.
\end{proposition}
\proof{Proof.}
See Section \ref{app:blackboxproof2}. 
\endproof
\hfill
\Halmos

Proposition \ref{thm:statstestsampling} demonstrates that if \(m \ge \frac{1}{2\varepsilon^2}\log\!\big(\frac{2N}{\delta}\big)\), where $\varepsilon
=
\min\left(
\frac{t\epsilon}{16},
\frac{t}{16K\log K},
\frac{t^2}{64K},
\frac{1}{2K},
\frac{\epsilon}{4}
\right)$, then the Type I and Type II errors decay exponentially in the text length \(N\), with rates governed by the threshold \(t\) and the same logarithmic scales \(c_1\log(K)\) and \(-c_2\log(\epsilon)\). Similar to Proposition \ref{thm:statstestmultibb}, the sampling budget $m$ grows only logarithmically in \(N\) and \(1/\delta\) but scales as \(1/\varepsilon^2\), and an additive \(+\delta\) term accounts for the residual probability of sampling inaccuracy. Note that since sampling is required only for a single evaluator \(A\) rather than across \(\mathcal{A}\cup\mathcal{B}\), the sampling penalty here appears as \(\delta\) rather than \((|\mathcal{A}|+|\mathcal{B}|)\delta\). Moreover, the relevant separation is \(\epsilon_2\) from Assumption \ref{assum:mindiff}, and sampling reduces the usable gap in the Type I bound to \(\epsilon_2/4\), yielding the stricter requirement \(t<\epsilon_2/4\). This tightening reflects that a portion of the separation guaranteed by \(\epsilon_2\) must be allocated to absorb estimation error in \(\hat l_A\) and \(\hat h_N(A,A)\), leaving a smaller residual margin for controlling false positives. Appendix~\ref{BBexperiments} shows that, empirically, $m \leq 1000$ suffices in many settings.

\noindent\textbf{Managerial implications.}
Relative to Proposition~\ref{thm:statstest}, Proposition~\ref{thm:statstestsampling} introduces an additional design decision because the evaluator model's next-token probabilities are unavailable and must be estimated through sampling. The manager therefore has three main levers: (i) the sample size \(m\) used to construct each empirical next-token distribution, together with the sampling tolerance \(\delta\); (ii) the attribution tolerance \(t\); and (iii) the minimum text length \(N\). The first lever determines the reliability of the empirical distributions. The parameter \(\delta\) is the additional amount in each error bound that the manager allows for estimating the evaluator probabilities rather than observing them directly. After selecting \(\delta\), the manager can use Proposition~\ref{thm:statstestsampling} to determine a sufficient sample size \(m\). A smaller \(\delta\) limits this additional error more tightly but requires more samples. Holding the remaining parameters fixed, the sufficient \(m\) increases only logarithmically with \(N\) and \(1/\delta\). Thus, increasing the document length or tightening the permitted sampling-related increase does not require a proportional increase in the number of samples used at each token position. The second lever, \(t\), determines how strict the attribution rule is. A smaller \(t\) requires the observed statistic to be closer to the behavior expected under model \(A\) before the text is attributed to \(A\). It lowers the \(t\)-dependent component of the Type~I error bound but increases the Type~II error bound. A smaller \(t\) may therefore be appropriate when attribution can lead to disciplinary, regulatory, or other consequential action, whereas a larger \(t\) may be appropriate when the audit is used to identify cases for subsequent human review. The third lever is the minimum text length \(N\). Once \(m\) satisfies the sampling requirement, the exponential components of both error bounds decrease with \(N\), while the additional sampling term remains \(\delta\). The manager can therefore invert the bounds to determine the minimum length of an employee report, student assignment, or other audited document. If she also wants the sampling-related contribution to decrease as longer texts strengthen the other components of the guarantees, she can select a smaller \(\delta\); the resulting sufficient \(m\) increases only logarithmically with \(1/\delta\). Finally, \(K\) and \(\epsilon\) enter the concentration terms through \(\log K\) and \(\log(1/\epsilon)\); consequently, their effects on these terms grow only logarithmically as new models adopt larger vocabularies or finer numerical precision.

%In hypothesis testing, the null hypothesis reflects the default/conservative assumption. So, if we accuse someone of having used LLM $A$, they are innocent until proven guilty. Accordingly, t
%\footnote{Recall from section \ref{model} that an application of the Bayes’ rule yields that any text generation process corresponds to some sequential model $B$ regardless of whether tokens are sequentially generated. Our test does not require any information about $B$.}. 

\section{Concentration inequality proofs}\label{app:concentrationinequalroadmap}

In this section, to improve readability, we first provide the prerequisites of sub-exponential random variables that are used in our proofs in Section \ref{app:proofconsentrationprereq}, followed by a review of notations in Section \ref{app:proofnotationrev}. We provide the proof for Theorem \ref{thm:maindiff} - part (a), i.e., where the generator model $B$ and the evaluator model $A$ are the same, in Section \ref{sec:proofmainsame}. The starting paragraph in the proof provides a high-level overview, describing the proof steps, remarks, and lemmas used as part of the proof. In Section \ref{app:proofthmdiff}, we provide the proof for Theorem \ref{thm:maindiff} - part (b), i.e., where the generator model $B$ is not the same as the evaluator model $A$. Similar to the previous part, the starting paragraph of this section provides a proof overview.

%\subsection*{Same generative and evaluator models}\label{sec:samemodels}

\subsection{Prerequisites of sub-exponential random variables from the literature}\label{app:proofconsentrationprereq}

Before presenting the proofs for our main theorem, we need to clearly state the following definitions and the equivalence lemma for sub-exponential random variables from the literature.

\begin{definition} \label{def:sub-exp}
    (sub-exponential norm). The sub-exponential norm of $X \in \mathbb{R}$ is
    \begin{equation*}
        \| X\|_{\psi_1}=\inf\bigg\{t>0:\mathbb{E}[e^{\frac{|X|}{t}}]\leq 2\bigg\}.
    \end{equation*}
    If $\| X\|_{\psi_1}$ is finite, we say that $X$ is sub-exponential. 
\end{definition}

We also need the following equivalent (up to a constant factor) definition of a sub-exponential random variable from \cite{vershynin2018high}.
    \begin{definition}\label{def:serv}
    (sub-exponential random variable). Centered random variable $X\in SE(\nu^2, \alpha)$ with parameters $\nu, \alpha >0$ is sub-exponential if
    \begin{eqnarray*}
        \mathbb{E}[e^{\lambda X}]\leq e^{\frac{\lambda^2\nu^2}{2}}, \quad \forall \lambda: |\lambda|<\frac{1}{\alpha}.
    \end{eqnarray*}
\end{definition}
Next, we present a lemma from \citet{vershynin2018high} that demonstrates that the two definitions are equivalent up to a constant factor.
\begin{lemma}\label{SE properties}
    (SE properties ,\citealp{vershynin2018high}). Let $X$ be a random variable with $\mathbb{E}[X]=0$. Then, there exists a constant $c$ and constants $K_4$ and $K_5$ such that $K_4 \le c K_5$ and $K_5 \le c K_4$ and the following two properties are equivalent. 
    \begin{itemize}
        \item There exists a constant $K_4$, such that the MGF of $|X|$ is bounded, specifically
        \begin{eqnarray*}
            \mathbb{E}[e^{|X|/K_4}]\leq 2.
        \end{eqnarray*}
        \item  There exists a constant $K_5$, such that the MGF of $X$ satisfies 
    \begin{eqnarray*}
        \mathbb{E}[e^{\lambda X}]\leq e^{\frac{K_5^2\lambda^2}{2}} \quad \forall \lambda \quad s.t. \quad |\lambda| \leq \frac{1}{K_5}
    \end{eqnarray*}
    \end{itemize}
\end{lemma}

\subsection{Review of notations} \label{app:proofnotationrev}
Here, we first review the notation in Section \ref{sec:modelandbackground}: $\textbf{X}$ denotes user prompt and $\textbf{Y}$ denotes the output that consists of a string of tokens $\textbf{Y}=[Y_1, Y_2, \dots, Y_N, \dots]$. Each token is chosen from a finite vocabulary set, i.e., $Y_n \in \mathcal{X}$ with size $K:=|\mathcal{X}|$. We defined the random variable $Z_n^{A}:=-\log\big(p^{A}_n(Y_n)\big)$, and the zero-mean random variable $X_n^{A}:=Z_n^{A}-\mathbb{E}_{p^{B}_n}[Z_n^{A}]$. Additionally, we defined a random variable $S_N^{A} := \sum_{i=1}^{N} X^{A}_i$.

\subsection{Proof of Theorem \ref{thm:maindiff} - part (a)}\label{sec:proofmainsame}

We first provide a high-level structure for our proof. Recall that in part (a), we focus on the case where the generator model $B$ is the same as the evaluator model $A$.
In Remark \ref{remarkfiniteexpectation}, we show that $\mathbb{E}_{p_n^A}[Z_n^{A}]$ is upper bounded by $\log K$, and therefore it is finite. 
In Lemma \ref{lemmamartingale}, we show that the random variable $S_N^{A}=\sum_{i=1}^{N}X_i^{A}$ forms a martingale. In Lemma \ref{keylemma},  we find the sub-exponential norm for the martingale increments random variable $X_n^{A}$.
%To apply the martingale concentration bounds, we find the sub-exponential norm for the random variable $X_n$ in Lemma \ref{keylemma}. 
As the last step, we show that martingale sum $\sum_{n=1}^{N} X_n^{A}$ is sub-exponential with parameters $(\nu^2 N, \nu)$, with $\nu=c_1 \log K$.
Finally, we apply the tail bound for sub-exponential random variables to bound the probability of $|\sum_{n=1}^{N} X_n^{A}|$ being larger than a constant $t$. %Hence, we derive the concentration bounds for martingales to provide finite sample guarantee for the convergence of the random variable $S_N/N$ to zero. 
A challenge in applying the common concentration bounds for martingales is that martingale increments are not necessarily bounded. We overcome this issue by showing that the martingale differences, while not bounded, admit a light tail. In particular, we show that the martingale differences are sub-exponential.

%Suppose a string $\mathbf{Y}$ is generated by model $A$, and we evaluate the text using the same model $A$. Given a (sub-) string $\mathbf{Y}_N$ and the evaluator model $A$, we define the random variable $Z_n=-\log\big(p^{A}_n(Y_n)\big)$. Then,
%\begin{eqnarray*}
%   \mathbb{E}_{p_n^A}[Z_n] =-\sum_{y_n \in \mathcal{X}} p^{A}_n(y_n)\log p^{A}_n(y_n).
%\end{eqnarray*}
%Because we know $p^{A}_n(y_n)$, we also know the entropy $-\sum_{y_n \in \mathcal{X}} p^{A}_n(y_n)\log p^{A}_n(y_n)= \mathbb{E}_{p_n^A}[Z_n]$. 

\begin{remark}\label{remarkfiniteexpectation}
    \begin{eqnarray*}\label{eq:expectedup1} 
   \mathbb{E}_{p_n^A}[Z_n^{A}] =-\sum_{y_n \in \mathcal{X}} p^{A}_n(y_n)\log p^{A}_n(y_n) \leq \log K.
\end{eqnarray*}
\end{remark}
\proof{Proof.}
    Concavity of $-p^{A}_n(y_n)\log p^{A}_n(y_n)$ yields that its maximizer is $p^{*A}_n(y_n)=\frac{1}{K}, \forall y_n \in \mathcal{X}$. Hence, \begin{eqnarray*}
        \mathbb{E}_{p_n^A}[Z_n^{A}]=-\sum_{y_n\in \mathcal{X}}p_n^A(y_n)\log p_n^A(y_n)&\leq& -\sum_{y_n \in \mathcal{X}}p_n^{*A}(y_n)\log p_n^{*A}(y_n)\\&=& K.\frac{1}{K}.\log K=\log K.
    \end{eqnarray*}
\endproof
\hfill
\halmos

Now that we have shown $\mathbb{E}_{p_n^A}[Z_n^{A}]<\infty$, recall that we defined the zero-mean random variable $X_n^{A}=Z_n^{A}-\mathbb{E}_{p_n^A}[Z_n^{A}]$. Then, we defined $S_N^{A}=\sum_{i=1}^{N}X_i^{A}$, which as we show in Lemma \ref{lemmamartingale}, forms a martingale. 
%\subsubsection{Proof of Remark \ref{remarkfiniteexpectation}}
%\subsubsection{Proof of Lemma \ref{lemmamartingale}}

\begin{lemma} \label{lemmamartingale}
If the evaluator model $A$ is the same as the generator model $B$ that generates the string $\bf{Y}_N$, then the random variable $S_N=\sum_{i=1}^{N}X_i^{A}$ forms a martingale.
\end{lemma}
\proof{Proof.} Since $Z_n^A\geq 0$ and $\mathbb{E}_{p_n^A}[Z_n^A]\leq\log K$ by Remark~\ref{remarkfiniteexpectation}, the triangle inequality implies that $\mathbb{E}\big[|X_n^A|\big]\leq 2\log K$. To establish that the process $\{S_N^A\}_{N\geq 1}$ is a martingale, we need to verify that (i) $\mathbb{E}[|S_N^A|]<\infty$, and (ii) $\mathbb{E}[S_{N+1}^A\mid X,Y_1,\ldots,Y_N]=S_N^A$.

Property (i) is satisfied because $\mathbb{E}[|S_N^A|]\leq\sum_{n=1}^N\mathbb{E}[|X_n^A|]\leq 2N\log K<\infty$.

Property (ii) is satisfied because, by definition, $\mathbb{E}[X_{N+1}^A\mid X,Y_1,\ldots,Y_N]=0$, and therefore $\mathbb{E}[S_{N+1}^A\mid X,Y_1,\ldots,Y_N]=S_N^A$.

%$Z_n^{A}$ and as a result $\mathbb{E}_{p_n^A}(y_n)$ are positive random variables. In Remark \ref{remarkfiniteexpectation}, we showed that $\mathbb{E}_{p_n^A}[Z_n]\leq\log K$. Therefore, we have $\mathbb{E}\big[|X_n^{A}|\big] \le 2\log K$. 

%For a random variable to form a martingale, the following two properties need to be satisfied: (i) $\mathbb{E}\big[|S_{N+1}^{A}-S_N^{A}|\big]<\infty$, and (ii) $\mathbb{E}[S_{N+1}^{A}|S_N^{A}]=S_N^{A}$. 

%Property (i) is satisfied because $\mathbb{E}\big[|S_{N+1}^{A}-S_N^{A}|\big]=\mathbb{E}\big[|X_{N+1}^{A}|\big]\leq 2\log K<\infty$. 

%Property (ii) is satisfied because the martingale increments $X_n^{A}$ are, by definition, a zero-mean random variable conditioned on preceding tokens.

\endproof
\hfill
\halmos

\begin{lemma}\label{keylemma}
    The sub-exponential norm for the random variable $X_n^{A}$ is less than or equal to $\frac{2}{\log(2)}\log(K)$.
\end{lemma}

%\textbf{Proof of Lemma \ref{keylemma}.}
\proof{Proof}
    For the random variable to be sub-exponential, by Definition \ref{def:sub-exp}, we need to find $t$ such that
    \begin{eqnarray}\label{keyinequality}
        \sum_{y_n \in \mathcal{X}}p^{A}_n(y_n) e^{\frac{\big|-\log p^{A}_n(y_n)-\mathbb{E}[Z_n^{A}]\big|}{t}}\le e^{\frac{\mathbb{E}[Z_n]}{t}}\sum_{y_n \in \mathcal{X}} p^{A}_n(y_n) \frac{1}{p^{A}_n(y_n)^{1/t}}=e^{\frac{\mathbb{E}[Z_n]}{t}}\sum_{y_n \in \mathcal{X}} p^{A}_n(y_n)^{\frac{t-1}{t}} \leq 2.
    \end{eqnarray}
where the first inequality follows from the triangle inequality and that $Z_n$ and therefore $\mathbb{E}[Z_n]$ are positive. Note that $\sum_{y_n \in \mathcal{X}}p^{A}_n(y_n)^{\frac{t-1}{t}}$ is concave in $p^{A}_n(y_n)$. Hence, it attains its maximum when we have $p^{A}_n(y_i)=p^{A}_n(y_j)\quad  \forall y_i,y_j \in \mathcal{X}$, which yields 
\begin{eqnarray}\label{eq:upperbound}
    \sum_{y_n \in \mathcal{X}}p^{A}_n(y_n)^{\frac{t-1}{t}} \le K(\frac{1}{K})^{\frac{t-1}{t}}= K^{\frac{1}{t}}.
\end{eqnarray}
To analyze $e^{\frac{\mathbb{E}[Z_n^{A}]}{t}}$, recall from Remark~\ref{remarkfiniteexpectation} that
\begin{eqnarray}\label{eq:expectedup}
    \mathbb{E}_{p_n^A}[Z_n^{A}]
    =-\sum_{y_n \in \mathcal{X}} p^{A}_n(y_n)\log p^{A}_n(y_n)
    \leq \log(K).
\end{eqnarray}
Combining Equations~\eqref{keyinequality}, \eqref{eq:upperbound}, and \eqref{eq:expectedup} yields that for $X_n^A$ to be sub-exponential with norm $t$, it is sufficient that
\begin{equation*}
    K^{1/t}K^{1/t}\le 2.
\end{equation*}
Hence, $X_n^{A}$ has sub-exponential norm $\| X\|_{\psi_1}\leq\frac{2}{\log(2)}\log(K)$.
\endproof
\hfill
\halmos
%\noindent{\color{blue} Proof in \ref{app:lemma1}.}

%and there exists an absolute constant $C$ such that property $i$ implies property $j$ with parameter $K_j\leq CK_i$ for any two properties $i,j$ 

\textbf{Proof of Theorem \ref{thm:maindiff} part (a)}
\proof{Proof.}

Next, note that by Lemma \ref{keylemma} and Lemma \ref{SE properties}, there exists $c_1>0$ s.t. for $\nu=c_1 \log K$, and for any $n \in \mathbb{N}$,
\begin{eqnarray}\label{eq:singlesubexp}
    \mathbb{E}[e^{\lambda X_n^{A}} | \mathbf{Y}_{n-1}]\leq e^{\frac{\nu^2 \lambda^2}{2}} \quad \forall |\lambda| \leq \frac{1}{\nu}. 
\end{eqnarray}
By Definition \ref{def:serv}, we have $X_n \in SE(\nu^2, \nu)$. 

Our next step is to show that $\sum_{n=1}^{N} X_n$ is SE with parameters $(\nu^2 N, \nu)$. To realize that, observe that
\begin{eqnarray*}
    \mathbb{E} \left[ e^{\lambda \left( \sum_{k=1}^{n} X_k^{A} \right)} \right] = \mathbb{E} \left[ e^{\lambda \left( \sum_{k=1}^{n-1} X_k^{A} \right)} \mathbb{E} \left[ e^{\lambda X_n^{A}} \mid \mathbf{Y}_{n-1} \right] \right] 
\leq \mathbb{E} \left[ e^{\lambda \sum_{k=1}^{n-1} X_k^{A}} \right] e^{\frac{\lambda^2 \nu^2}{2}} 
\leq e^{\frac{\lambda^2  N \nu^2}{2}},
\end{eqnarray*}
where the first equation follows from the iterated law of expectation, and the first inequality follows from Equation (\ref{eq:singlesubexp}). 

Finally, if $S \in SE(\nu^2, \alpha)$ is a sub-exponential random variable, then %fromTheorem 5.2 in\citealp{arinaldo2019}
\begin{eqnarray}\label{eq:tailsub}
    \mathbb{P}(|S-\mathbb{E}[S]|\geq t_1) \leq 2\exp \bigg( -\frac{1}{2} \min\bigg(\frac{t_1^{2}}{\nu^2}, \frac{t_1}{\alpha} \bigg) \bigg).
\end{eqnarray}
Substituting for the zero-mean random variable $\sum_{n=1}^{N} X_n$ that is sub-exponential with parameters $(\nu^2 N, \nu)$, with $\nu=c_1 \log K$, we obtain 
\begin{eqnarray*}
    \mathbb{P}(|\sum_{n=1}^{N} X_n^{A}|\geq t_1) \leq 2\exp \bigg( -\frac{1}{2} \min\bigg(\frac{t_1^{2}}{N(c_1 \log K)^2}, \frac{t_1}{c_1 \log K} \bigg) \bigg).
\end{eqnarray*}
Setting $t_1=t N$ and absorbing the factor $2$ in $c_1$, we have
\begin{eqnarray*}
    \mathbb{P}\bigg(\frac{1}{N}|\sum_{i=1}^{N}X_i^{A}|\geq t \bigg) \le 2 \exp \bigg[ -\frac{N t}{2c_1\log(K)} \min \bigg(1, \frac{t}{c_1\log(K)} \bigg)  \bigg].
\end{eqnarray*}

Since Lemma~\ref{keylemma} yields $\|X_n^A\|_{\psi_1}
\leq\frac{2}{\log(2)}\log K,$ the conversion from the $\psi_1$-norm bound to Definition~\ref{def:serv} allows us to take $\nu=2\sqrt{2}\,\|X_n^A\|_{\psi_1}\leq\frac{4\sqrt{2}}{\log(2)}\log K.$ Hence, $c_1=\frac{4\sqrt{2}}{\log(2)}.$ This concludes the proof for Theorem \ref{thm:maindiff} part (a). 
\endproof
\hfill
\halmos

\subsection{Proof of Theorem \ref{thm:maindiff} - part (b)} \label{app:proofthmdiff}

Similar to the proof of part (a), here we provide a high-level structure for our proof. First, recall from the discussion under Assumption \ref{assum:crossmodel}, as proved in Equation \ref{outcomeofassumption1}, that $\mathbb{E}_{p_n^B}[Z_n^{A}]$ is upper bounded by $-\log(\epsilon)$, and therefore it is finite. 
Then, In Lemma \ref{lemmamartingale2}, we show that the random variable $S_N^{A}=\sum_{i=1}^{N}X_i^{A}$ forms a martingale. In Lemma \ref{keylemma2},  we find the sub-exponential norm for the martingale increments random variable $X_n^{A}$.
As the last step, we show that martingale sum $\sum_{n=1}^{N} X_n^{A}$ is sub exponential with parameters 
$(\nu^2 N,\nu)$, where $\nu=-c_2 \log(\epsilon)$. Finally, we apply the tail bound for sub-exponential random variables to bound the probability of $|\sum_{n=1}^{N} X_n^{A}|$ being larger than a constant $t$.

\begin{lemma} \label{lemmamartingale2}
Under assumption \ref{assum:crossmodel}, the random variable $S_N^{A}=\sum_{i=1}^{N}X_i^{A}$ forms a martingale. 
\end{lemma}
\proof{Proof.} 
$Z_n^{A}$ is a positive random variables, so  $\mathbb{E}_{p_n^B}[Z_n^{A}]$ is also positive.
In Equation (\ref{outcomeofassumption1}), we have shown that $\mathbb{E}_{p_n^B}[Z_n^{A}]$ is upper bounded by $-\log(\epsilon)$, and therefore we have $\mathbb{E}\big[|X_n^{A}|\big] \le -2\log(\epsilon)$. To establish that the process $\{S_N^A\}_{N\geq 1}$ is a martingale, we need to verify that (i) $\mathbb{E}[|S_N^A|]<\infty$, and (ii) $\mathbb{E}[S_{N+1}^A\mid X,Y_1,\ldots,Y_N]=S_N^A$.

Property (i) is satisfied because $\mathbb{E}[|S_N^A|]
\leq
\sum_{n=1}^N \mathbb{E}[|X_n^A|]
\leq
-2N\log(\epsilon)
<
\infty$.

Property (ii) is satisfied because, by definition, $\mathbb{E}[X_{N+1}^A\mid X,Y_1,\ldots,Y_N]=0,$ and therefore $\mathbb{E}[S_{N+1}^A\mid X,Y_1,\ldots,Y_N]=S_N^A$.

%For a random variable to form a martingale, the following two properties need to be satisfied: (i) $\mathbb{E}\big[|S_{N+1}^{A}-S_N^{A}|\big]<\infty$, and (ii) $\mathbb{E}[S_{N+1}^{A}|S_N^{A}]=S_N^{A}$. 

%Property (i) is satisfied because $\mathbb{E}\big[|S_{N+1}^{A}-S_N^{A}|\big]=\mathbb{E}\big[|X_{N+1}^{A}|\big]\leq -\log(\epsilon)<\infty$. 

%Property (ii) is satisfied because the martingale increments $X_n^{A}$ are, by definition, a zero-mean random variable conditioned on past tokens.
\endproof
\hfill
\halmos

%\subsubsection{Proof of Lemma \ref{keylemma2}} \label{app:proofkeylemma2}
\begin{lemma}\label{keylemma2}
Under Assumption \ref{assum:crossmodel}, the sub-exponential norm for the random variable $X_n$ is less than or equal to $-\frac{2}{\log(2)}\log(\epsilon)$. 
\end{lemma}
\proof{Proof.}
    For the random variable to be sub-exponential, by definition \ref{def:sub-exp}, we need to find $t$ such that
    \begin{eqnarray*}\label{keyinequality2}
        \sum_{y_n \in \mathcal{X}}p^{B}_n(y_n) e^{\frac{\big|-\log p^{A}_n(y_n)-\mathbb{E}[Z_n^{A}]\big|}{t}}\le e^{\frac{\mathbb{E}[Z_n^{A}]}{t}}\sum_{y_n \in \mathcal{X}}  \frac{p^{B}_n(y_n)}{p^{A}_n(y_n)^{1/t}}\le 
        e^{\frac{\mathbb{E}[Z_n^{A}]}{t}}\sum_{y_n \in \mathcal{X}}  \frac{p^{B}_n(y_n)}{\epsilon^{1/t}} \le e^{\frac{\mathbb{E}[Z_n^{A}]-\log(\epsilon)}{t}}\leq 2.
%e^{\frac{\mathbb{E}[Z_n^{A}]}{t}}\sum_{y_n \in \mathcal{X}} p_{n}(y_n)^{\frac{t-1}{t}} \leq 2.
    \end{eqnarray*}
Thus, for $X_n$ to be sub-exponential with norm $t$, it is sufficient to satisfy
\begin{equation}\label{app:reqforlemma2}
    \frac{\mathbb{E}[Z_n^{A}]-\log(\epsilon)}{t} \le \log(2).
\end{equation}
Recall equation \ref{outcomeofassumption1} that states
\begin{eqnarray*}
    \mathbb{E}_{p^{B}_n}[Z_n^{A}]= \sum_{y_n \in \mathcal{X}} -p^{B}_n(y_n) \log(p^{A}_n(y_n)) \le -\log(\epsilon),
\end{eqnarray*}
and therefore $\mathbb{E}_{p^{B}_n}[Z_n^{A}]-\log(\epsilon)\leq -2\log(\epsilon)$. Substituting in (\ref{app:reqforlemma2}) yields that $X_n^{A}$ has sub-exponential norm $\| X\|_{\psi_1}\leq -\frac{2}{\log(2)}\log(\epsilon)$.
\endproof
\hfill
\halmos

\textbf{Proof of Theorem \ref{thm:maindiff} part (b)}
\proof{Proof.}
First note that, by Lemma \ref{keylemma2} and Lemma \ref{SE properties}, there exists $c_2>0$ such that for $\nu=-c_2 \log(\epsilon)$, by Lemma \ref{keylemma2} and Lemma \ref{SE properties}, and for any $n \in \mathbb{N}$,
\begin{eqnarray}
    \mathbb{E}[e^{\lambda X_n^{A}} | \mathbf{Y}_{n-1}]\leq e^{\frac{\nu^2 \lambda^2}{2}} \quad \forall |\lambda| \leq \frac{1}{\nu}. 
\end{eqnarray}
By Definition \ref{def:serv}, we have $X_n \in SE(\nu^2, \nu)$. 
Then, following the exact same steps as in proof of Theorem \ref{thm:maindiff}-part (a),
we conclude that $\sum_{n=1}^{N}X_n^{A}$ is sub-exponential with parameter $S_N^{A} \in SE(\nu^2 N,\nu)$, where $\nu=-c_2\log(\epsilon)$. Again, following the exact same steps we described in proof of part (a)
, from (\ref{eq:tailsub}) we have
\begin{eqnarray*}
    \mathbb{P}(|\sum_{n=1}^{N} X_n^{A}|\geq t_1) \leq 2\exp \bigg( -\frac{1}{2} \min\bigg(\frac{t_1^{2}}{N({-c_2\log \epsilon)^2}}, \frac{t_1}{-c_2\log \epsilon} \bigg) \bigg).
\end{eqnarray*}
Finally, setting $t_1=t N$, we obtain
\begin{eqnarray*}
    \mathbb{P}\bigg(\frac{1}{N}|\sum_{n=1}^{N}X_n^{A}|\geq t \bigg) \le 2 \exp \bigg[ -\frac{N t}{-2c_2\log(\epsilon)} \min \bigg(1, \frac{t}{-c_2\log(\epsilon)} \bigg)  \bigg].
\end{eqnarray*}

Since Lemma~\ref{keylemma2} yields $\|X_n^A\|_{\psi_1}
\leq
-\frac{2}{\log(2)}\log(\epsilon),$ the conversion from the $\psi_1$-norm bound to Definition~\ref{def:serv} allows us to take $\nu
=
2\sqrt{2}\,\|X_n^A\|_{\psi_1}
\leq
-\frac{4\sqrt{2}}{\log(2)}\log(\epsilon).$ Thus, $c_2
=
\frac{4\sqrt{2}}{\log(2)}.$ This concludes the proof for Theorem \ref{thm:maindiff} part (b).
\endproof
\hfill
\halmos

\section{Statistical test proofs}\label{app:stattestroadmap}

In Section \ref{app:testmulti}, we present the proof for Proposition \ref{thm:statstestmulti}. In Section \ref{app:stattestthm}, we first provide a proof overview in the first paragraph, and then provide the proof for Proposition \ref{thm:statstest}.

%\subsubsection{Proof of Proposition \ref{thm:statstesttwo}} \label{app:testtwo}

\subsection{Proof of Proposition \ref{thm:statstestmulti}}\label{app:testmulti}
\proof{Proof.}
Type I error occurs if a model $B_j \in \mathcal{B}$ generates the text string $\mathbf{Y}_N$, but for one model $A_i \in \mathcal{A}$, we have 
\begin{eqnarray*}
    \frac{1}{N} \sum_{n=1}^{N} Z^{A_i}_n <  \frac{1}{N} \sum_{n=1}^{N} Z^{B_j}_n .
\end{eqnarray*}
This yields for some $A_i \in \mathcal{A}$, we have
\begin{eqnarray*}
     h_N(B_j,A_i)-h_N(B_j,A_i)+\frac{1}{N} \sum_{n=1}^{N} Z^{A_i}_n  \le  h_N(B_j,B_j)-h_N(B_j,B_j)+ \frac{1}{N}\sum_{n=1}^{N} Z^{B_j}_n .
\end{eqnarray*}
First, note that by Assumption \ref{assum:mindiff2m}, we have $h_N(B_j,A_i)-h_N(B_j,B_j) \ge \epsilon_1$. Hence, the Type I error occurs only if at least for one of the models $A_i \in \mathcal{A}$ we have
\begin{eqnarray*}
     \epsilon_1-h_N(B_j,A_i)+\frac{1}{N} \sum_{n=1}^{N} Z^{A_i}_n  \le  -h_N(B_j,B_j)+ \frac{1}{N}\sum_{n=1}^{N} Z^{B_j}_n .
\end{eqnarray*}
Equivalently, Type I error occurs only if at least for one of the models $A_i \in \mathcal{A}$ we have
\begin{eqnarray*}
     \epsilon_1 \le \bigg|-h_N(B_j,A_i)+\frac{1}{N} \sum_{n=1}^{N} Z^{A_i}_n\bigg|    + \bigg|-h_N(B_j,B_j)+ \frac{1}{N}\sum_{n=1}^{N} Z^{B_j}_n \bigg|.
\end{eqnarray*}
So, Type I error only occurs if, for at least one of the models $A_i \in \mathcal{A}$, we have
either
$|-h_N(B_j,A_i)+\frac{1}{N} \sum_{n=1}^{N} Z^{A_i}_n|>\epsilon_1/2$ or $|-h_N(B_j,B_j)+ \frac{1}{N}\sum_{n=1}^{N} Z^{B_j}_n |>\epsilon_1/2$. Hence, we upper bound the Type I error as
\begin{eqnarray*}
    \mathbb{P}\bigg( \bigg| \frac{1}{N} \sum_{n=1}^{N} Z^{B_j}_n- h_N(B_j,B_j)(\mathbf{Y}_N) \bigg| \ge \epsilon_1/2 \bigg) + |\mathcal{A}|\mathbb{P}\bigg( \bigg| \frac{1}{N} \sum_{n=1}^{N} Z^{A_i}_n- h_N(B_j,A_i)(\mathbf{Y}_N) \bigg| \ge \epsilon_1/2 \bigg).
\end{eqnarray*}
Next, from Theorem \ref{thm:maindiff}-part (a), we have
\begin{eqnarray*}
    \mathbb{P}\bigg( \bigg| \frac{1}{N} \sum_{n=1}^{N} Z^{B_j}_n- h_N(B_j,B_j)(\mathbf{Y}_N) \bigg| \ge \epsilon_1/2 \bigg) \le 2 \exp \bigg[ -\frac{N \epsilon_1/2}{c_1\log(K)} \min \bigg(1, \frac{\epsilon_1/2}{c_1\log(K)} \bigg)  \bigg].
\end{eqnarray*}
Also, from Theorem \ref{thm:maindiff}, we have
\begin{eqnarray*}
    \mathbb{P}\bigg( \bigg| \frac{1}{N} \sum_{n=1}^{N} Z^{A_i}_n- h_N(B_j,A_i)(\mathbf{Y}_N) \bigg| \ge \epsilon_1/2 \bigg) \le  2 \exp \bigg[ -\frac{N (\epsilon_1/2)}{-c_2\log(\epsilon)} \min \bigg(1, \frac{(\epsilon_1/2)}{-c_2\log(\epsilon)} \bigg)  \bigg].
\end{eqnarray*}
Therefore, the Type I error is at most 
\begin{eqnarray*}
    2 \exp \bigg[ -\frac{N \epsilon_1/2}{c_1\log(K)} \min \bigg(1, \frac{\epsilon_1/2}{c_1\log(K)} \bigg)  \bigg]+2|\mathcal{A}| \exp \bigg[ -\frac{N (\epsilon_1/2)}{-c_2\log(\epsilon)} \min \bigg(1, \frac{(\epsilon_1/2)}{-c_2\log(\epsilon)} \bigg)  \bigg].
\end{eqnarray*}
The Type II error occurs if a model $A_i \in \mathcal{A}$ generates the text $\bf{Y}_N$, but for a model $B_j \in \mathcal{B}$ we have 
\begin{eqnarray*}
    \frac{1}{N} \sum_{n=1}^{N} Z^{B_j}_n < \frac{1}{N} \sum_{n=1}^{N} Z^{A_i}_n.
\end{eqnarray*}
This yields that for some $B_j \in \mathcal{B}$, we have
\begin{eqnarray*}
     h_N(A_i,B_j)-h_N(A_i,B_j)+\frac{1}{N} \sum_{n=1}^{N} Z^{B_j}_n  \le  h_N(A_i,A_i)-h_N(A_i,A_i)+ \frac{1}{N}\sum_{n=1}^{N} Z^{A_i}_n.
\end{eqnarray*}
Note that by Assumption \ref{assum:mindiff2m}, we have $h_N(A_i,B_j)-h_N(A_i,A_i) \ge \epsilon_1$. Hence, the Type II error occurs only if at least for one of the models $B_j \in \mathcal{B}$ we have
\begin{eqnarray*}
     \epsilon_1-h_N(A_i,B_j)+\frac{1}{N} \sum_{n=1}^{N} Z^{B_j}_n  \le  -h_N(A_i,A_i)+ \frac{1}{N}\sum_{n=1}^{N} Z^{A_i}_n .
\end{eqnarray*}
Equivalently, the Type II error occurs only if at least for one of the models $B_j \in \mathcal{B}$ we have
\begin{eqnarray*}
    \epsilon_1 \le \bigg| -h_N(A_i,B_j)+\frac{1}{N} \sum_{n=1}^{N} Z^{B_j}_n \bigg|  + \bigg| -h_N(A_i,A_i)+ \frac{1}{N}\sum_{n=1}^{N} Z^{A_i}_n \bigg|.
\end{eqnarray*}
%%%%%%%%%%
So, the Type II error only occurs if at least for one of the models $B_j \in \mathcal{B}$, we have
either
$|-h_N(A_i,B_j)+\frac{1}{N} \sum_{n=1}^{N} Z^{B_j}_n|>\epsilon_1/2$ or $|-h_N(A_i,A_i)+ \frac{1}{N}\sum_{n=1}^{N} Z^{A_i}_n |>\epsilon_1/2$. Hence, we upper bound the Type II error as
\begin{eqnarray*}
    |\mathcal{B}|\mathbb{P}\bigg( \bigg| \frac{1}{N} \sum_{n=1}^{N} Z^{B_j}_n- h_N(A_i,B_j)(\mathbf{Y}_N) \bigg| \ge \epsilon_1/2 \bigg) + \mathbb{P}\bigg( \bigg| \frac{1}{N} \sum_{n=1}^{N} Z^{A_i}_n- h_N(A_i,A_i)(\mathbf{Y}_N) \bigg| \ge \epsilon_1/2 \bigg).
\end{eqnarray*}
Next, from Theorem \ref{thm:maindiff} we have
\begin{eqnarray*}
    \mathbb{P}\bigg( \bigg| \frac{1}{N} \sum_{n=1}^{N} Z^{B_j}_n- h_N(A_i,B_j)(\mathbf{Y}_N) \bigg| \ge \epsilon_1/2 \bigg) \le 2 \exp \bigg[ -\frac{N (\epsilon_1/2)}{-c_2\log(\epsilon)} \min \bigg(1, \frac{(\epsilon_1/2)}{-c_2\log(\epsilon)} \bigg)  \bigg].
\end{eqnarray*}
Also, from Theorem \ref{thm:maindiff} part (a) we have
\begin{eqnarray*}
    \mathbb{P}\bigg( \bigg| \frac{1}{N} \sum_{n=1}^{N} Z^{A_i}_n- h_N(A_i,A_i)(\mathbf{Y}_N) \bigg| \ge \epsilon_1/2 \bigg) \le 2 \exp \bigg[ -\frac{N \epsilon_1/2}{c_1\log(K)} \min \bigg(1, \frac{\epsilon_1/2}{c_1\log(K)} \bigg)  \bigg].
\end{eqnarray*}
Therefore, the Type II error is at most 
\begin{eqnarray*}
    2|\mathcal{B}| \exp \bigg[ -\frac{N (\epsilon_1/2)}{-c_2\log(\epsilon)} \min \bigg(1, \frac{(\epsilon_1/2)}{-c_2\log(\epsilon)} \bigg)  \bigg]+2 \exp \bigg[ -\frac{N \epsilon_1/2}{c_1\log(K)} \min \bigg(1, \frac{\epsilon_1/2}{c_1\log(K)} \bigg)  \bigg].
\end{eqnarray*}
\endproof
\hfill
\halmos

%\subsubsection{Proof of Lemma \ref{lemma:bounding}}\label{app:testlemma}

\subsection{Proof of Proposition \ref{thm:statstest}} \label{app:stattestthm}

In this section, we present the proof for a generalized version of Proposition \ref{thm:statstest}. In particular, we first present Lemma~\ref{lemma:bounding}, which, under the sufficient condition in Equation~\ref{eq:sufficient}, provides an upper bound on the probability that $\left|h_N(B,A)(\mathbf{Y}_N)-h_N(A,A)(\mathbf{Y}_N)\right|$ is at most a threshold $c_3$. In the proof of Proposition~\ref{thm:statstest}, this separation probability is one of the quantities used to bound the Type~I error. Note that providing an upper bound on the Type II error does not require any assumptions.

%In particular, we first present Lemma \ref{lemma:bounding} that provides an upper bound on the probability that, under the sufficient condition in Equation (\ref{eq:sufficient}) from Remark \ref{assumption3practical}, the absolute value of the difference between the cross-entropy of the (unknown) generator model $B$ and the evaluator model $A$ and the entropy of the evaluator model $A$ over string $\mathbf{Y}_N$ be less than or equal to a threshold $c_3$, i.e., Assumption  \ref{assum:mindiff} holds for a constant $c_3$ except for an exponentially small probability in $N$. 

%In the proof of Proposition \ref{thm:statstest}, we use Lemma \ref{lemma:bounding} as one of the three inequalities that yield an upper bound on Type I error.

\begin{lemma}\label{lemma:bounding}
    Under the sufficient condition in Equation~(\ref{eq:sufficient}), for any positive constant $c_3 < \epsilon_2$, we have 
    \begin{eqnarray*}
        \mathbb{P} \bigg( |h_N(B,A)(\mathbf{Y}_N)- h_N(A,A)(\mathbf{Y}_N)| \le c_3\bigg) \le 2 \exp \bigg[ -\frac{N (\epsilon_2-c_3)}{-c_2\log(\epsilon)} \min \bigg(1, \frac{(\epsilon_2-c_3)}{-c_2\log(\epsilon)} \bigg)  \bigg].
    \end{eqnarray*}
\end{lemma}

\proof{Proof.}

%{\color{blue}

%\begin{assumption}\label{assum:mindiff} (minimum tangible difference). We assume that if the generative and evaluator models are different, then 
%\begin{eqnarray*}
%    \frac1N \sum_{n=3}^{N}  \mathbb{E} \bigg[  D_{KL}\big(p_n^{B}||p_n^{A}\big) \bigg| \textbf{Y}_{n-2} \bigg] \ge  4 \log^2\big(|\chi|\big).
%\end{eqnarray*}
%\end{assumption}
%}

We prove this lemma in three steps. 
In this proof, for notation brevity we write $D_{KL}\big(p_n^{B}||p_n^{A}\big)$ instead of $[ D_{KL}\big(p_n^{B}||p_n^{A}\big) | \mathbf{Y}_{n-2}]$, and $\mathbb{E}\big[ D_{KL}\big(p_n^{B}||p_n^{A}\big) \big]$ instead of $\mathbb{E}\big[ D_{KL}\big(p_n^{B}||p_n^{A}\big) | \mathbf{Y}_{n-2}\big]$. \smallskip

\textit{Step 1.} sub-exponential norm for $[D_{KL}\big(p_n^{B}||p_n^{A}\big)| \mathbf{Y}_{n-2}]$

For the random variable $D_{KL}\big(p_n^{B}||p_n^{A}\big)$, applying (\ref{outcomeofassumption1}) we obtain
\begin{eqnarray}\label{kl-bounding}
    D_{KL}\big(p_n^{B}||p_n^{A}\big)= \sum_{y_n \in \mathcal{X}} p^{B}_n(y_n) \big( \log(p^{B}_n(y_n))-\log(p^{A}_n(y_n)) \big) \le \sum_{y_n \in \mathcal{X}} -p^{B}_n(y_n) \log(p^{A}_n(y_n)) \le -\log(\epsilon),
\end{eqnarray}
where the first inequality holds since $\sum_{y_n \in \mathcal{X}} p_n^B(y_n)\log(p_n^B(y_n))\leq 0$, and the second inequality holds by Assumption \ref{assum:crossmodel}. For the random variable to be sub-exponential, by Definition \ref{def:sub-exp}, we need to find $t$ such that
\begin{eqnarray*}
\mathbb{E}\bigg[e^{\frac{\big|D_{KL}(p_n^{B}||p_n^{A})-\mathbb{E}[D_{KL}(p_n^{B}||p_n^{A})]\big|}{t}}\bigg] \le 2.
\end{eqnarray*}
Applying (\ref{kl-bounding}), for the random variable to be sub-exponential with norm $t$, it is sufficient to satisfy
\begin{eqnarray*}
\mathbb{E}\bigg[e^{\frac{\big|D_{KL}(p_n^{B}||p_n^{A})-\mathbb{E}[D_{KL}(p_n^{B}||p_n^{A})]\big|}{t}}\bigg] \le e^{\frac{-2\log(\epsilon)}{t}}\leq 2.
\end{eqnarray*}
Hence, $[D_{KL}\big(p_n^{B}||p_n^{A}\big)| \mathbf{Y}_{n-2}]$ has sub-exponential norm at most $-4\log(\epsilon)$. \smallskip

\textit{Step 2.} Concentration bounds for $[D_{KL}\big(p_n^{B}||p_n^{A}\big)| \mathbf{Y}_{n-2}]$

Following the same steps as in the proof for Theorem \ref{thm:maindiff}- part (a), we conclude that $\sum_{n=1}^{N} [D_{KL}\big(p_n^{B}||p_n^{A}\big)| \mathbf{Y}_{n-2}]$ is sub-exponential with parameter $S_N \in SE(\nu\sqrt{N},\alpha)$, where $\alpha=\nu=-c_2 \log(\epsilon)$ for a constant $c_2 > 0$. Then, from (\ref{eq:tailsub}) we have
\begin{eqnarray*}
&&\mathbb{P}\bigg(\bigg|\sum_{n=1}^{N} D_{KL}\big(p_n^{B}||p_n^{A}\big)- \sum_{n=1}^{N}\mathbb{E}\Big[ D_{KL}\big(p_n^{B}||p_n^{A}\big)\Big]\bigg|\geq t \bigg)
    \le 2\exp \bigg( -\frac{1}{2} \min\bigg(\frac{t^{2}}{N(-c_2 \log \epsilon)^2}, \frac{t}{-c_2 \log \epsilon} \bigg) \bigg).
\end{eqnarray*}
Next, setting $t=(\epsilon_2-c_3) N$, we obtain
\begin{eqnarray}\label{eq:boundforkl}
   \mathbb{P}\bigg(\big|\sum_{n=1}^{N} D_{KL}\big(p_n^{B}||p_n^{A}\big)- \sum_{n=1}^{N}\mathbb{E}\Big[ D_{KL}\big(p_n^{B}||p_n^{A}\big)\Big]\big|\geq t\bigg) \le 2 \exp \bigg[ -\frac{N (\epsilon_2-c_3)}{-2c_2\log(\epsilon)} \min \bigg(1, \frac{(\epsilon_2-c_3)}{-c_2\log(\epsilon)} \bigg)  \bigg].
\end{eqnarray} \smallskip

\textit{Step 3.} Tail bound for cross-entropy

First, we have 
\begin{eqnarray}\label{eq:bounding}
    &&\Big|h_N(B,A)(\mathbf{Y}_N)- h_N(A,A)(\mathbf{Y}_N)\Big|
   \nonumber \\
   &\ge&h_N(B,A)(\mathbf{Y}_N)- h_N(A,A)(\mathbf{Y}_N)
   \nonumber \\
    &=&\frac1N \sum_{n=1}^{N}  D_{KL}\big(p_n^{B}||p_n^{A}\big) +h_N(B,B)(\mathbf{Y}_N)- h_N(A,A)(\mathbf{Y}_N)\nonumber\\
    &\ge& \frac1N \sum_{n=1}^{N}  D_{KL}\big(p_n^{B}||p_n^{A}\big)  - h_N(A,A)(\mathbf{Y}_N)\nonumber\\
    &\ge&\frac{1}{N} \sum_{n=1}^{N}  D_{KL}\big(p_n^{B}||p_n^{A}\big) - \frac{1}{N} \sum_{n=3}^{N} \mathbb{E} \bigg[ D_{KL}\big(p_n^{B}||p_n^{A}\big) \bigg]+\epsilon_2,
\end{eqnarray}
where the first inequality follows from $|x|\ge x$, the second follows from $h_N(B,B)\ge0$, and the last inequality follows from
Equation~\eqref{eq:sufficient}. Hence, we have
\begin{eqnarray*}
    &&\mathbb{P} \bigg( |h_N(B,A)(\mathbf{Y}_N)- h_N(A,A)(\mathbf{Y}_N)| \le c_3\bigg) \\
    &\le& \mathbb{P} \bigg( \frac{1}{N} \sum_{n=1}^{N}  D_{KL}\big(p_n^{B}||p_n^{A}\big)  - \frac{1}{N} \sum_{n=3}^{N} \mathbb{E} \bigg[ D_{KL}\big(p_n^{B}||p_n^{A}\big) \bigg]+\epsilon_2 \le c_3\bigg) \\
    &=& \mathbb{P} \bigg( -\frac{1}{N} \sum_{n=1}^{N}  D_{KL}\big(p_n^{B}||p_n^{A}\big)  + \frac{1}{N} \sum_{n=3}^{N} \mathbb{E} \bigg[ D_{KL}\big(p_n^{B}||p_n^{A}\big) \bigg] \ge -c_3+\epsilon_2 \bigg) \\
    &\le& \mathbb{P} \bigg( -\frac{1}{N} \sum_{n=1}^{N}  D_{KL}\big(p_n^{B}||p_n^{A}\big)  + \frac{1}{N} \sum_{n=1}^{N} \mathbb{E} \bigg[ D_{KL}\big(p_n^{B}||p_n^{A}\big) \bigg] \ge -c_3+\epsilon_2 \bigg) \\
    &\le&  \mathbb{P} \bigg(  \bigg|\frac{1}{N} \sum_{n=1}^{N}  D_{KL}\big(p_n^{B}||p_n^{A}\big)   - \frac{1}{N} \sum_{n=1}^{N} \mathbb{E} \bigg[ D_{KL}\big(p_n^{B}||p_n^{A}\big) \bigg]\bigg| \ge \epsilon_2-c_3 \bigg)
    \\
    &\le& 2 \exp \bigg[ -\frac{N (\epsilon_2-c_3)}{-c_2\log(\epsilon)} \min \bigg(1, \frac{\epsilon_2-c_3}{-c_2\log(\epsilon)} \bigg)  \bigg],
\end{eqnarray*}
where the first inequality follows from (\ref{eq:bounding}). The one-to-last inequality follows from the property of absolute value. The last inequality follows from (\ref{eq:boundforkl}). This concludes the proof.
\endproof
\hfill
\halmos

\textbf{Proof of Proposition \ref{thm:statstest}}

\proof{Proof.}
Type I error occurs if a model $B \neq A$ generates the text string $\mathbf{Y}_N$, but we have 
\begin{eqnarray*}
    \bigg| \frac{1}{N} \sum_{n=1}^{N} Z_n^{A}- h_N(A,A)(\mathbf{Y}_N) \bigg| \le t.
\end{eqnarray*}
Applying triangle inequality, we have 
\begin{eqnarray*}
   \bigg| h_N(B,A)(\mathbf{Y}_N)- h_N(A,A)(\mathbf{Y}_N) \bigg|- \bigg| \frac{1}{N} \sum_{n=1}^{N} Z_n^{A}- h_N(B,A)(\mathbf{Y}_N) \bigg| \le \bigg| \frac{1}{N} \sum_{n=1}^{N} Z_n^{A}- h_N(A,A)(\mathbf{Y}_N) \bigg| .
\end{eqnarray*}
Hence, the Type I error is upper bounded as
\begin{eqnarray}\label{eq:handlemain}
    &&\mathbb{P}\bigg( \bigg| \frac{1}{N} \sum_{n=1}^{N} Z_n^{A}- h_N(A,A)(\mathbf{Y}_N) \bigg| \le t \bigg) \nonumber\\
    &\le& \mathbb{P}\bigg( \bigg| h_N(B,A)(\mathbf{Y}_N)- h_N(A,A)(\mathbf{Y}_N) \bigg|- \bigg| \frac{1}{N} \sum_{n=1}^{N} Z_n^{A}- h_N(B,A)(\mathbf{Y}_N) \bigg| \le t \bigg) \nonumber\\
    &\le&
    \mathbb{P}\bigg( \bigg| \frac{1}{N} \sum_{n=1}^{N} Z_n^{A}- h_N(B,A)(\mathbf{Y}_N) \bigg| \ge c_3-t \bigg) + \mathbb{P} \bigg( \bigg|h_N(B,A)(\mathbf{Y}_N)- h_N(A,A)(\mathbf{Y}_N)\bigg| \le c_3\bigg).
\end{eqnarray}
For any positive $c_3<\epsilon_2$, the following bound holds in each case stated in Proposition~\ref{thm:statstest}: under Assumption~\ref{assum:mindiff}, its left-hand side is zero; under the sufficient condition in Equation~\ref{eq:sufficient}, it follows from Lemma~\ref{lemma:bounding}.\begin{eqnarray}\label{eq:handle1}
    \mathbb{P} \bigg( \bigg|h_N(B,A)(\mathbf{Y}_N)- h_N(A,A)(\mathbf{Y}_N)\bigg| \le c_3\bigg) \le 2 \exp \bigg[ -\frac{N (\epsilon_2-c_3)}{-c_2\log(\epsilon)} \min \bigg(1, \frac{(\epsilon_2-c_3)}{-c_2\log(\epsilon)} \bigg)  \bigg].
\end{eqnarray}
From Theorem \ref{thm:maindiff} we know that
\begin{eqnarray}\label{eq:handle2}
    \mathbb{P}\bigg( \bigg| \frac{1}{N} \sum_{n=1}^{N} Z_n^{A}- h_N(B,A)(\mathbf{Y}_N) \bigg| \ge c_3-t \bigg) \le  2 \exp \bigg[ -\frac{N (c_3-t)}{-c_2\log(\epsilon)} \min \bigg(1, \frac{(c_3-t)}{-c_2\log(\epsilon)} \bigg)  \bigg].
\end{eqnarray}
Combining Equations (\ref{eq:handlemain}-\ref{eq:handle2}), we conclude the Type I error is upper bounded as
\begin{eqnarray*}
    2 \exp \bigg[ -\frac{N (\epsilon_2-c_3)}{-c_2\log(\epsilon)} \min \bigg(1, \frac{(\epsilon_2-c_3)}{-c_2\log(\epsilon)} \bigg)  \bigg]+2 \exp \bigg[ -\frac{N (c_3-t)}{-c_2\log(\epsilon)} \min \bigg(1, \frac{(c_3-t)}{-c_2\log(\epsilon)} \bigg)  \bigg].
\end{eqnarray*}
Setting $c_3=\epsilon_2/2$, yields that the Type I error is upper bounded as
\begin{eqnarray*}
    2 \exp \bigg[ -\frac{N (\epsilon_2/2)}{-c_2\log(\epsilon)} \min \bigg(1, \frac{(\epsilon_2/2)}{-c_2\log(\epsilon)} \bigg)  \bigg]+2 \exp \bigg[ -\frac{N (\epsilon_2/2-t)}{-c_2\log(\epsilon)} \min \bigg(1, \frac{(\epsilon_2/2-t)}{-c_2\log(\epsilon)} \bigg)  \bigg].
\end{eqnarray*}

The Type II error occurs if the model $A$ generates the text $\bf{Y}_N$, but we have 
\begin{eqnarray*}
    \bigg| \frac{1}{N} \sum_{n=1}^{N} Z_n^{A}- h_N(A,A)(\mathbf{Y}_N) \bigg| \ge t.
\end{eqnarray*}
We upper bound the probability of this event as
\begin{eqnarray*}
    \mathbb{P}\bigg( \bigg| \frac{1}{N} \sum_{n=1}^{N} Z_n^{A}- h_N(A,A)(\mathbf{Y}_N) \bigg| \ge t \bigg) \le 2 \exp \bigg[ -\frac{N t}{c_1\log(K)} \min \bigg(1, \frac{t}{c_1\log(K)} \bigg)  \bigg],
\end{eqnarray*}
where the inequality follows from Theorem \ref{thm:maindiff}-part (a). This concludes the proof.
\endproof
\hfill
\halmos

%From Assumption \ref{assum:mindiff}, we know that for any positive $c_3<\epsilon_2$, we have

%\section{Lower bound proofs}\label{app:lowerboundproof}
%\proof{Proof.}
%For the statistic $T$, consider event $A$ as the event that $T$ rejects the null hypothesis. Then, applying Theorem 14.2 from \citet{lattimore2020bandit}, we obtain
%\begin{equation}\label{eq:lecam-iid}
% \alpha_n(T) + \beta_n(T) \ge \frac{1}{2} e^{-D_{KL}(P_B^{\otimes N} || P_A^{\otimes N})}= \frac{1}{2} e^{-N \cdot D_{KL}(P_B || P_A)},
%\end{equation}
%which concludes the proof.
%\endproof
%\hfill
%\Halmos

\section{Black-box detection with sampling proofs}\label{app:blackboxproofsroadmap}

In this section, we determine how large the sample size $m$ must be in order to construct empirical log-perplexity and empirical average cross-entropy statistics that yield the same testing performance as in the white-box setting, even though, unlike in the white-box setting, we do not have direct access to the conditional probability distributions of the models. We first prove Lemma~\ref{lemma:blackboxbound}, which provides a lower bound on the number of samples $m$ needed so that the empirical quantities $\hat{l}_A(\mathbf{Y}_N)$ and $\hat{h}_N(A,A)(\mathbf{Y}_N)$ uniformly approximate their white-box counterparts $l_A(\mathbf{Y}_N)$ and $h_N(A,A)(\mathbf{Y}_N)$ with high probability. Building on this lemma, we then establish Propositions~\ref{thm:statstestmultibb} and~\ref{thm:statstestsampling}, showing that, under the corresponding sample size requirements, the tests based on these empirical black-box statistics achieve Type~I and Type~II error guarantees that match those of the white-box tests in Propositions~\ref{thm:statstestmulti} and~\ref{thm:statstest}, up to constant slack in the thresholds.

\subsection{Lemma \ref{lemma:blackboxbound} and its proof}

\begin{lemma}\label{lemma:blackboxbound}

Fix $\delta\in(0,1)$. For any $0<\varepsilon\le
\min\left\{\frac{1}{2K},\frac{\epsilon}{4}\right\}$,
if we choose the sampling size according to
\begin{equation*}\label{eq:mchoice}
m \;\ge\; \frac{1}{2\varepsilon^2 }\log\!\left(\frac{2N}{\delta}\right).
\end{equation*}
Then, with probability of at least $(1-\delta)$ over the sampling used to build $\hat p^A_n$, we have
\begin{equation*}
\big|\widehat{l}_A(\mathbf{Y}_N)-l_A(\mathbf{Y}_N)\big|
~\le~ \frac{4\varepsilon}{\epsilon}.
\end{equation*}
Additionally, 
\begin{eqnarray*}
   |\hat{h}_N(A,A)(\textbf{Y}_N)- h_N(A,A)(\textbf{Y}_N)|
\;\le\;
 2 K \varepsilon \log K + (K \varepsilon)^{1/2}.
\end{eqnarray*}
\end{lemma}

%Fix $\delta\in(0,1)$, then for any $\varepsilon>0,\,$ if we choose the sampling size according to

\proof{Proof.}

Fixing a vocabulary order on the language model $\mathcal{X}=\{1,2,3,\ldots,K\}$,
we form the corresponding right-continuous  empirical cdf
$\hat{F}_n(k):=\sum_{i\le k}\hat{p}^A_n(i)$. Let $F_n(k):=\sum_{i\le k}p^A_n(i)$ be the true cdf. The Dvoretzky–Kiefer–Wolfowitz (DKW) inequality implies that, for each $n$, for all $\varepsilon>0$,
\begin{equation}\label{eq:dkw}
\mathbb{P}\!\left(\sup_{k\in[K]}\big|\hat{F}_n(k)-F_n(k)\big|>\varepsilon\right)\le 2e^{-2m\varepsilon^2}.
\end{equation}
The bound in (\ref{eq:dkw}) holds for any fixed $n$. We next aim to find a guarantee that the bound holds simultaneously for all $n=1,\ldots,N$. Let
\[
A_n := \Big\{\sup_{k\in[K]} \big|\hat{F}_n(k)-F_n(k)\big|>\varepsilon \Big\}.
\]
Denote the undesirable event that the $n^{\text{th}}$ empirical cdf deviates from its true counterpart by more than $\varepsilon$. Applying the union bound on (\ref{eq:dkw}) yields
\[
\mathbb{P}\!\left(\bigcup_{n=1}^{N} A_n\right)
\le \sum_{n=1}^{N} \mathbb{P}(A_n)
\le 2N e^{-2m\varepsilon^2}.
\]
To ensure that this joint failure probability does not exceed a prescribed level $\delta \in (0,1)$,
we require
\[
2N e^{-2m\varepsilon^2}\le \delta.
\]
Simplifying the expression yields
\begin{equation}
m \;\ge\; \frac{1}{2\varepsilon^2}\log\!\left(\frac{2N}{\delta}\right).
\end{equation}
Hence, if $m$ satisfies~\eqref{eq:mchoice}, then with a probability of at least $(1-\delta)$, for all $n=1,\ldots,N$,
\[
\sup_{k\in[K]}\big|\hat{F}_n(k)-F_n(k)\big|\le \varepsilon.
\]

%\noindentIntuitively, the DKW inequality controls the uniform deviation between empirical and true CDFs for a single distribution, while the union bound extends this guarantee to hold across all $N$ empirical distributions simultaneously.
%Applying Boole's Inequality/ union bound, for 
%\begin{eqnarray}\label{eq:mchoice}
 %   m \;\ge\; \frac{1}{2\varepsilon^2}\log\!\left(\frac{2N}{\delta}\right),
%\end{eqnarray}
%we have with probability $(1-\delta)$ for all $n=1,\ldots,N$, 
%\begin{equation}
%\sup_{k\in[K]}\big|\hat{F}_n(k)-F_n(k)\big|\le \varepsilon.
%\end{equation}

We denote this event by $\mathcal{A}:=\Big\{\sup_{k\in[K]}\big|\hat{F}_n(k)-F_n(k)\big|\le \varepsilon\Big\}$. Under $\mathcal{A}$, the empirical and the true cdf satisfy 
$\sup_{k\in[K]}|\hat F_n(k)-F_n(k)|\le\varepsilon$. For discrete ordered alphabets, each probability mass entry can be expressed as 
$\hat p^A_n(i)=\hat F_n(i)-\hat F_n(i-1)$ and $p^A_n(i)=F_n(i)-F_n(i-1)$, with $F_n(0)=0$. 
Hence,
\[
\big|\hat p^A_n(i)-p^A_n(i)\big|
  = \Big|\big(\hat F_n(i)-F_n(i)\big)-\big(\hat F_n(i-1)-F_n(i-1)\big)\Big|
  \le \big|\hat F_n(i)-F_n(i)\big| + \big|\hat F_n(i-1)-F_n(i-1)\big|
  \le 2\varepsilon,
\]
which implies
\begin{equation}\label{distanceforp}
    \max_{Y\in\mathcal X}\big|\hat p^A_n(Y)-p^A_n(Y)\big|\le 2\varepsilon.
\end{equation}
Denote by $\mathrm{TV}(\hat p^A_n,p^A_n)$, the total variation distance between $\hat p^A_n$ and $p^A_n$. Summing over all $i\in[K]$ yields
\begin{equation}\label{deltap}
    \|\hat p^A_n-p^A_n\|_1
   = \sum_{i=1}^K |\hat p^A_n(i)-p^A_n(i)|
   \le 2K\varepsilon,
\end{equation}
and
\begin{equation}\label{deltaTV}
    \mathrm{TV}(\hat p^A_n,p^A_n)
   = \tfrac12\|\hat p^A_n-p^A_n\|_1
   \le K\varepsilon.
\end{equation}
Fix $\varepsilon \le \frac{1}{2K}$, so that, under the event $\mathcal{A}$, equation (\ref{deltaTV}) simplify to
\begin{equation}\label{TVbound}
    \operatorname{TV}(\hat p^A_n,p^A_n)\le K\varepsilon\le\tfrac12.
\end{equation}

%Thus, on the event $\mathcal{A}$, the empirical and true distributions are uniformly close in both $\ell_1$ and total–variation distance, with each probability mass differing by at most $2\varepsilon$.
%We denote this event by $\mathcal{A}$. Hence, for discrete ordered alphabets, under event $\mathcal{A}$ we have,
%\[
%\|\hat{p}^A_n-p^A_n\|_1 \;\le\; 2K\varepsilon
%;\qquad
%\mathrm{TV}(\hat{p}^A_n,p^A_n)\;=\;\tfrac12\|\hat{p}^A_n-p^A_n\|_1 \;\le\; K\varepsilon; \qquad \max_{Y \in \mathcal{X}}|\hat{p}^A_n(Y)-p^A_n(Y)| \;\le\; 2\,\varepsilon
%\]
%Fix $\varepsilon \;<\; \frac{1}{2K},$
%so that on the event $\mathcal{A}$ we have
%\[
%\operatorname{TV}(\hat p^A_n,p^A_n)\;=\;\tfrac12\|\hat p^A_n-p^A_n\|_1
%\;\le\; K\varepsilon \;\le\; \tfrac12.
%\]

Recall that $H(p^{A}_n,p^{A}_n)=-\sum_{y\in \mathcal{X}}^{K}p^{A}_n(y)\log p^{A}_n(y)$, and define $h_2(x):=-x\log x-(1-x)\log(1-x)$. For the pmfs $\hat p^A_n, p^A_n$ on the $K$-point alphabet, the Fannes–Audenaert continuity bound for Shannon entropy yields
\begin{equation}\label{eq:FA}
\big|H(\hat{p}^A_n, \hat{p}^A_n)-H(p^A_n, p^A_n)\big|
\;\le\;
\operatorname{TV}(\hat{p}^A_n, p^A_n)\,\log(K-1)+h_2\big(\operatorname{TV}(\hat{p}^A_n, p^A_n)\big),
\end{equation}
where $\operatorname{TV}(\hat{p}^A_n, p^A_n)\in[0,1]$. Simplifying (\ref{eq:FA}) using the upper-bound in (\ref{TVbound}) yields
\begin{equation}\label{eq:FA-direct}
\big|H(\hat p^A_n, \hat p^A_n)-H(p^A_n, p^A_n)\big|
\;\le\;
(K\varepsilon)\,\log(K-1) \;+\; h_2(K\varepsilon).
\end{equation}

Upper-bounding (\ref{eq:FA-direct}) using $h_2(x)\le x\log\!\frac{e}{x}$ for $x\in(0,\tfrac12]$ yields
\begin{equation}\label{eq:FA-expanded}
\big|H(\hat p^A_n, \hat p^A_n)-H(p^A_n, p^A_n)\big|
\;\le\;
K\varepsilon\,\log\!\frac{e(K-1)}{K\varepsilon} \le K \varepsilon+ K \varepsilon \log K + (K \varepsilon)^{1/2}.
\end{equation}
This yields
\begin{eqnarray*}
    \frac1N\sum_{n=1}^{N}\big|H(\hat p^A_n, \hat p^A_n)-H(p^A_n, p^A_n)\big|
\;\le\;
K \varepsilon+ K \varepsilon \log K + (K \varepsilon)^{1/2}.
\end{eqnarray*}
Hence, by triangle inequality
\begin{eqnarray*}
    |\hat{h}_N(A,A)(\textbf{Y}_N)- h_N(A,A)(\textbf{Y}_N)| \le K \varepsilon+ K \varepsilon \log K + (K \varepsilon)^{1/2}
\end{eqnarray*}

Since $K$ is such that $\log K\ge 1$, we have
$K\varepsilon+K\varepsilon\log K
\le 2K\varepsilon\log K$,
which gives the stated entropy bound.

%In our setting,
%\[
%\operatorname{TV}(\hat p^A_n,p^A_n)\;\le\;K\varepsilon \;\le\; \tfrac12,
%\]
%hence applying \eqref{eq:FA} with $p=\hat p^A_n$, $q=p^A_n$ gives

Next, to calculate the error in estimating the log-perplexity using sampling, we have
for $\varepsilon \le \epsilon/4$, whenever $p^A_n(Y_n)>0$, on the event $\mathcal{A}$, applying the Mean Value Theorem, we have
\begin{equation}\label{eq:logdev-nosmooth}
\big|\log \hat{p}^A_n(Y_n)-\log p^A_n(Y_n)\big|
~\le~
\frac{|\hat{p}^A_n(Y_n)-p^A_n(Y_n)|}{\min\{p^A_n(Y_n),\,\hat p^A_n(Y_n)\}}
~\le~
\frac{2\varepsilon}{\,\min\{p^A_n(Y_n),\,\hat p^A_n(Y_n)\}\,} \le \frac{4 \varepsilon}{\epsilon},
\end{equation}
where the second inequality follows from (\ref{distanceforp}), and last inequality follows from the positivity condition that yields $\min\{\,p^A_n(Y_n),\,\hat{p}^A_n(Y_n)\,\} \;\ge\; \frac{\epsilon}{2}$.
This yields 
\begin{equation*}\label{eq:bb-main-bound-nosmooth}
\big|\widehat{l}_A(\mathbf{Y}_N)-l_A(\mathbf{Y}_N)\big|
~\le~ \frac{4\varepsilon}{\epsilon}.
\end{equation*}

\endproof
\hfill
\Halmos

\subsection{Proof of Proposition \ref{thm:statstestmultibb}} \label{app:proofofblackbox1}
\proof{Proof.}
As we show in Lemma \ref{lemma:blackboxbound}, by setting 
\begin{eqnarray*}
   \varepsilon
=
\min\left(
\frac{\epsilon_1\epsilon}{16},
\frac{\epsilon_1}{16K\log K},
\frac{\epsilon_1^2}{64K},
\frac{1}{2K},
\frac{\epsilon}{4}
\right),
\end{eqnarray*}
and
\begin{eqnarray*}
    m \ge \frac{1}{2\varepsilon^2 }\log\!\left(\frac{2N}{\delta}\right)
\end{eqnarray*}
with probability of at least $1-\delta$, for each model $M \in \mathcal{A} \cup \mathcal{B}$, we have
\begin{eqnarray*}
    \big|\hat{l}_M(\mathbf{Y}_N)-l_M(\mathbf{Y}_N)\big|
~\le~ \frac{4\varepsilon}{\epsilon}\le \epsilon_1/4.
\end{eqnarray*}
Hence, a union bound yields that with probability of at least $1-\big(|\mathcal{A}|+|\mathcal{B}|\big)\delta$, for all models $M \in \mathcal{A} \cup \mathcal{B}$, we have 
\begin{eqnarray*}
    \Bigg| \frac{1}{N} \sum_{n=1}^{N} Z_n^{M}  -\frac{1}{N} \sum_{n=1}^{N} \hat{Z}_n^{M}  \Bigg| \le \frac{\epsilon_1}{4}.
\end{eqnarray*}
This yields that with probability of at least $1-\big(|\mathcal{A}|+|\mathcal{B}|\big)\delta$ for any model $A_i \in \mathcal{A}$ and any model $B_j \in \mathcal{B}$, we have 
\begin{eqnarray*}
  \Bigg| \frac{1}{N} \sum_{n=1}^{N} Z_n^{A_i} - \frac{1}{N} \sum_{n=1}^{N} Z_n^{B_j}  \Big| - \Big| \frac{1}{N} \sum_{n=1}^{N} \hat{Z}_n^{A_i} - \frac{1}{N} \sum_{n=1}^{N} \hat{Z}_n^{B_j}  \Bigg| \le \frac{\epsilon_1}{2}.
\end{eqnarray*}
Therefore, with a probability of at least $1-\big(|\mathcal{A}|+|\mathcal{B}|\big)\delta$, Type I and Type II errors of the test reduce to that in Proposition \ref{thm:statstestmulti}, with the mere difference that we replace $\epsilon_1$ with $\frac{\epsilon_1}{2}$. 
\endproof
\hfill
\Halmos

\subsection{Proof of Proposition \ref{thm:statstestsampling}}\label{app:blackboxproof2}

\proof{Proof.}

As we show in Lemma \ref{lemma:blackboxbound}, by setting 
$$\varepsilon
=
\min\left(
\frac{t\epsilon}{16},
\frac{t}{16K\log K},
\frac{t^2}{64K},
\frac{1}{2K},
\frac{\epsilon}{4}
\right),$$
and
\begin{eqnarray*}
    m \ge \frac{1}{2\varepsilon^2 }\log\!\left(\frac{2N}{\delta}\right)
\end{eqnarray*}
with a probability of at least $1-\delta$, we have
\begin{eqnarray*}
    \big|\hat{l}_A(\mathbf{Y}_N)-l_A(\mathbf{Y}_N)\big|
~\le~ \frac{4\varepsilon}{\epsilon}\le t/4.
\end{eqnarray*}
and
\begin{eqnarray*}
     \frac1N\sum_{n=1}^{N}\big|H(\hat p^A_n)-H(p^A_n)\big|
\;\le\;
 2 K \varepsilon \log K + (K \varepsilon)^{1/2} \le t/4.
\end{eqnarray*}
Hence, with probability of at least $1-\delta$, we have
\begin{eqnarray*}
  \Bigg| \Big| \frac{1}{N} \sum_{n=1}^{N} Z_n^{A} - h_N(A,A)(\mathbf{Y}_N)  \Big| - \Big|  \hat{l}_A(\mathbf{Y}_N) - \hat{h}_N(A,A)(\mathbf{Y}_N)  \Big|  \Bigg| \le \frac{t}{2}.
\end{eqnarray*}
Therefore, on this event, a Type~II error implies
\[
\left|
l_A(\mathbf{Y}_N)-h_N(A,A)(\mathbf{Y}_N)
\right|>\frac{t}{2},
\]
which gives the stated Type~II bound. A Type~I error implies
\[
\left|
l_A(\mathbf{Y}_N)-h_N(A,A)(\mathbf{Y}_N)
\right|\le \frac{3t}{2}.
\]
Applying the Type~I argument of Proposition
\ref{thm:statstest} with threshold $3t/2$ gives a bound no
larger than the displayed Type~I bound, since
$t<\epsilon_2/4$ implies
\[
\frac{\epsilon_2}{2}\ge\frac{\epsilon_2}{4},
\qquad
\frac{\epsilon_2}{2}-\frac{3t}{2}
\ge
\frac{\epsilon_2}{4}-t.
\]
%Therefore, with probability of at least $1-\delta$ the problem reduces to the problem in Proposition \ref{thm:statstest} with the mere difference that we replace $\epsilon_2$ with $\frac{\epsilon_2}{2}$. 
\endproof
\hfill
\Halmos

\section{Proof of Corollary \ref{cor:lecam-iid}} \label{appendixlowerbound}

\proof{Proof.}\label{app:proofofcorr}
For the statistic $T$, consider event $A$ as the event that $T$ rejects the null hypothesis. Then, applying Theorem 14.2 from \citet{lattimore2020bandit}, we obtain
\begin{equation}\label{eq:lecam-iid1}
 \alpha_N(T) + \beta_N(T) \ge \frac{1}{2} e^{-D_{KL}(P_B^{\otimes N} || P_A^{\otimes N})}= \frac{1}{2} e^{-N \cdot D_{KL}(P_B || P_A)},
\end{equation}
which concludes the proof.
\endproof
\hfill
\Halmos

\section{Clipped Log Scores}
\label{app:clipped_log_scores}
%%%%%%%%%%%%%%%%%%%%%%%%%%%%%%%%%%%%%%%%%%%%%%%%%%%%%%%%%%%%%%%%%%%%%%%%%%%%

This appendix studies versions of the statistical tests in
Propositions~\ref{thm:statstestmulti} and~\ref{thm:statstest}, where the logarithmic score used by evaluator model \(A\) is clipped. The generation mechanism of every model remains unchanged. That is, if model \(B\) generates the text, then
\[
Y_n\mid \mathcal F_{n-1}\sim p_n^B(\cdot),
\qquad
\mathcal F_n:=\sigma(\mathbf X,Y_1,\ldots,Y_n).
\]
Throughout this appendix, fix a chosen clipping level $\tau\in(0,1),$ and define $L_\tau:=\log(1/\tau).$

\noindent\textbf{Bias-corrected clipped scores.} For evaluator model \(A\), define the clipped counterpart of the log score by $\widetilde z_n^A(y):=-\log\!\left(\max\{p_n^A(y),\tau\}\right).$ Note that for every \(y\in\mathcal X\), we have $0\leq \widetilde z_n^A(y)\leq L_\tau.$ Define the clipping correction term as
\begin{equation}
r_{\tau,n}^{A}
:=
\sum_{\substack{y\in\mathcal X:\\0<p_n^A(y)<\tau}}
p_n^A(y)
\log\!\left(\frac{\tau}{p_n^A(y)}\right)
\geq 0.
\label{eq:clipping_correction}
\end{equation}
Because \(p_n^A(\cdot)\) is determined by the the preceding tokens, \(r_{\tau,n}^{A}\) is \(\mathcal F_{n-1}\)-measurable. We define the bias-corrected clipped counterpart of \(Z_n^A\) by
\begin{equation}
\widetilde Z_n^{A}
:=
\widetilde z_n^{A}(Y_n)+r_{\tau,n}^{A}.
\label{eq:corrected_clipped_score}
\end{equation}
We then define the clipped counterpart of \(l_A\) by $\widetilde l_A(\mathbf Y_N):=\frac1N\sum_{n=1}^N\widetilde Z_n^{A}$ and the clipped counterpart
of \(h_N(B,A)\) by $\widetilde h_N(B,A)(\mathbf Y_N)
:=
\frac1N\sum_{n=1}^N
\mathbb E_B\!\left[\widetilde Z_n^A\mid\mathcal F_{n-1}\right].$ The conditional mean for \(\widetilde Z_n^A\) under
generator \(B\) is
\begin{equation}
\mathbb E_B\!\left[\widetilde Z_n^A\mid\mathcal F_{n-1}\right]
=
\sum_{y\in\mathcal X}p_n^B(y)\widetilde z_n^A(y)+r_{\tau,n}^{A}.
\label{eq:corrected_clipped_conditional_mean}
\end{equation}
Note that the weights in the conditional mean are the unmodified generator probabilities \(p_n^B\). 
When the generator and evaluator are the same, the correction in \eqref{eq:clipping_correction} yields
\begin{equation}
\widetilde h_N(A,A)(\mathbf Y_N)
=
h_N(A,A)(\mathbf Y_N).
\label{eq:corrected_self_center_average}
\end{equation}
That is because $\mathbb E_A\!\left[\widetilde Z_n^A\mid\mathcal F_{n-1}\right]
=
\sum_{y\in\mathcal X}p_n^A(y)\widetilde z_n^A(y)+r_{\tau,n}^{A}
=
-\sum_{y\in\mathcal X}p_n^A(y)\log p_n^A(y).$

%Thus, when \(A\) generates and evaluates the text, the bias-corrected clipped statistic remains centered at the same ordinary entropy target used in the main paper.

%%%%%%%%%%%%%%%%%%%%%%%%%%%%%%%%%%%%%%%%%%%%%%%%%%%%%%%%%%%%%%%%%%%%%%%%%%%%
\subsection{Concentration of the bias-corrected clipped score}
\label{app:clipped_concentration}
%%%%%%%%%%%%%%%%%%%%%%%%%%%%%%%%%%%%%%%%%%%%%%%%%%%%%%%%%%%%%%%%%%%%%%%%%%%%

\begin{lemma}
\label{lemma:clipped_concentration}
Let \(B\) be any text-generation process and let \(A\) be any evaluator
model. No lower bound on the positive values of \(p_n^B\) or \(p_n^A\) is
required. Then, for every \(u>0\),
\begin{equation*}
\mathbb P_B\left\{
\left|
\widetilde l_A(\mathbf Y_N)
-
\widetilde h_N(B,A)(\mathbf Y_N)
\right|
\geq u
\right\}
\leq
2\exp\left(
-\frac{2Nu^2}{L_\tau^2}
\right).
\end{equation*}
\end{lemma}
\proof{Proof.}
Define $\widetilde X_n^A
:=
\widetilde Z_n^{A}
-
\mathbb E_B\!\left[\widetilde Z_n^A\mid\mathcal F_{n-1}\right].$
Using \eqref{eq:corrected_clipped_score} and
\eqref{eq:corrected_clipped_conditional_mean}, we have
\begin{align*}
\widetilde X_n^A
&=
\widetilde z_n^{A}(Y_n)+r_{\tau,n}^{A}
-
\left\{
\sum_{y\in\mathcal X}p_n^B(y)\widetilde z_n^A(y)+r_{\tau,n}^{A}
\right\}
=
\widetilde z_n^{A}(Y_n)-\sum_{y\in\mathcal X}p_n^B(y)\widetilde z_n^A(y).
\end{align*}
Because \(p_n^B\) is the conditional law of \(Y_n\) given
\(\mathcal F_{n-1}\), we have $\mathbb E_B\left[
\widetilde X_n^A\mid\mathcal F_{n-1}
\right]
=0.$ Therefore,
\(\{\widetilde X_n^A\}_{n=1}^N\) is a martingale-difference sequence. Conditional on \(\mathcal F_{n-1}\), the random variable
\(\widetilde z_n^{A}(Y_n)\) belongs to the interval \([0,L_\tau]\).
Therefore, Azuma–Hoeffding's lemma (note that clipped random variables make the martingale increments conditionally sub-Gaussian) yields
\begin{equation*}
\mathbb E_B\left[
\exp\left\{
\lambda \widetilde X_n^A
\right\}
\middle|\mathcal F_{n-1}
\right]
\leq
\exp\left(
\frac{\lambda^2L_\tau^2}{8}
\right)
\qquad
\forall\lambda\in\mathbb R.
\end{equation*}
Iterating conditional expectations yields
\begin{equation*}
\mathbb E_B\left[
\exp\left\{
\lambda\sum_{n=1}^N
\widetilde X_n^A
\right\}
\right]
\leq
\exp\left(
\frac{N\lambda^2L_\tau^2}{8}
\right).
\end{equation*}
Hence, for every \(\lambda>0\), we have $\mathbb P_B\left\{
\sum_{n=1}^N \widetilde X_n^A\geq Nu
\right\}
\leq
\exp\left(
-\lambda Nu+\frac{N\lambda^2L_\tau^2}{8}
\right).$ The right-hand side is minimized at $\lambda=\frac{4u}{L_\tau^2},$
which yields $\mathbb P_B\left\{
\frac1N\sum_{n=1}^N \widetilde X_n^A\geq u
\right\}
\leq
\exp\left(
-\frac{2Nu^2}{L_\tau^2}
\right).$ Applying the same argument to
\(-\widetilde X_n^A\) and using a union bound concludes the proof.
\endproof
\hfill
\Halmos

The clipping correction does not enlarge the stochastic fluctuation scale:
it is \(\mathcal F_{n-1}\)-measurable and cancels after conditional centering. The concentration scale is therefore determined by \(L_\tau=\log(1/\tau)\). We next present a one-sided comparison result that we will then use for the multi-model attribution test.

\begin{lemma}
\label{lemma:clipped_pairwise_comparison}
Let \(B\) be the true generator and let \(A\) be another evaluator model.
Suppose that, for some deterministic \(d>0\),
\begin{equation}
\widetilde h_N(B,A)(\mathbf Y_N)
-
\widetilde h_N(B,B)(\mathbf Y_N)
\geq d
\qquad
\mathbb P_B\text{-almost surely}.
\label{eq:pairwise_corrected_gap}
\end{equation}
Then,
\begin{equation}
\mathbb P_B\left\{
\widetilde l_A(\mathbf Y_N)
\leq
\widetilde l_B(\mathbf Y_N)
\right\}
\leq
\exp\left(
-\frac{Nd^2}{2L_\tau^2}
\right).
\label{eq:pairwise_clipped_comparison_bound}
\end{equation}
\end{lemma}
\proof{Proof.}
Define the pairwise score difference $D_n
:=
\widetilde Z_n^{A}
-
\widetilde Z_n^{B}$
and its conditional mean $\mu_n
:=
\mathbb E_B\left[
D_n\mid\mathcal F_{n-1}
\right].$
By definition, $\frac1N\sum_{n=1}^N\mu_n
=
\widetilde h_N(B,A)
-
\widetilde h_N(B,B).$ Conditional on \(\mathcal F_{n-1}\), the correction
\(r_{\tau,n}^{A}-r_{\tau,n}^{B}\) is fixed, while $\widetilde z_n^{A}(Y_n)-\widetilde z_n^{B}(Y_n)
\in[-L_\tau,L_\tau].$ Therefore, conditional on \(\mathcal F_{n-1}\),
\(D_n\) belongs to an interval of length \(2L_\tau\). Azuma–Hoeffding inequality therefore yields
\begin{equation*}
\mathbb E_B\left[
\exp\left\{
-\lambda
\left(
D_n-\mu_n
\right)
\right\}
\middle|\mathcal F_{n-1}
\right]
\leq
\exp\left(
\frac{\lambda^2L_\tau^2}{2}
\right).
\end{equation*}
Iterating conditional expectations and applying Chernoff's inequality yields
\begin{equation*}
\mathbb P_B\left\{
\sum_{n=1}^N
\left(
D_n-\mu_n
\right)
\leq -Nd
\right\}
\leq
\exp\left(
-\lambda Nd+\frac{N\lambda^2L_\tau^2}{2}
\right).
\end{equation*}
Optimizing at $\lambda=\frac{d}{L_\tau^2}$
yields $\mathbb P_B\left\{
\sum_{n=1}^N
\left(
D_n-\mu_n
\right)
\leq -Nd
\right\}
\leq
\exp\left(
-\frac{Nd^2}{2L_\tau^2}
\right).$ Under \eqref{eq:pairwise_corrected_gap}, the event $\widetilde l_{A}\leq\widetilde l_{B}$
implies $\sum_{n=1}^N
\left(
D_n-\mu_n
\right)
\leq -Nd$ and \eqref{eq:pairwise_clipped_comparison_bound} follows.
\endproof
\hfill
\Halmos

\subsection{Clipped analogue of Proposition~\ref{thm:statstestmulti}}
\label{app:clipped_attribution}
%%%%%%%%%%%%%%%%%%%%%%%%%%%%%%%%%%%%%%%%%%%%%%%%%%%%%%%%%%%%%%%%%%%%%%%%%%%%

Consider the attribution hypotheses as in Section~\ref{sec:multi}: $\mathbf H_0:B\in\mathcal B,
\quad
\mathbf H_1:B\in\mathcal A,$
where $\mathcal A=\{A_1,\ldots,A_p\},
\quad
\mathcal B=\{B_1,\ldots,B_q\}.$ The bias-corrected clipped attribution test rejects \(\mathbf H_0\) if
there exists an \(A_i\in\mathcal A\) such that
\begin{equation}
\widetilde l_{A_i}(\mathbf Y_N)
<
\widetilde l_{B_j}(\mathbf Y_N)
\qquad
\forall B_j\in\mathcal B.
\label{eq:clipped_attribution_test}
\end{equation}

The following assumption is the clipped analogue of
Assumption~\ref{assum:mindiff2m}. It requires the true model to have a
uniformly smaller conditional expected bias-corrected clipped score than
every model in the competing set.

\begin{assumption}
\label{assum:clipped_attribution_separation}
(minimum clipped-score separation).
There exists a constant \(\widetilde\epsilon_1>0\) such that, for every
\(A_i\in\mathcal A\) and \(B_j\in\mathcal B\),
\begin{equation}
\widetilde h_N(B_j,A_i)(\mathbf Y_N)
-
\widetilde h_N(B_j,B_j)(\mathbf Y_N)
\geq
\widetilde\epsilon_1
\qquad
\mathbb P_{B_j}\text{-almost surely},
\label{eq:clipped_gap_BA}
\end{equation}
and
\begin{equation}
\widetilde h_N(A_i,B_j)(\mathbf Y_N)
-
\widetilde h_N(A_i,A_i)(\mathbf Y_N)
\geq
\widetilde\epsilon_1
\qquad
\mathbb P_{A_i}\text{-almost surely}.
\label{eq:clipped_gap_AB}
\end{equation}
\end{assumption}

\begin{proposition}
\label{prop:clipped_statstestmulti}
Suppose Assumption~\ref{assum:clipped_attribution_separation} holds.
Then the Type~I error of the test in
Equation~\eqref{eq:clipped_attribution_test} satisfies
\begin{equation*}
\sup_{B_j\in\mathcal B}
\mathbb P_{B_j}
\left\{
\text{reject }\mathbf H_0
\right\}
\leq
|\mathcal A|
\exp\left(
-\frac{N\widetilde\epsilon_1^2}{2L_\tau^2}
\right).
\end{equation*}
The Type~II error satisfies
\begin{equation*}
\sup_{A_i\in\mathcal A}
\mathbb P_{A_i}
\left\{
\text{fail to reject }\mathbf H_0
\right\}
\leq
|\mathcal B|
\exp\left(
-\frac{N\widetilde\epsilon_1^2}{2L_\tau^2}
\right).
\end{equation*}
\end{proposition}

\proof{Proof.}
Fix \(B_j\in\mathcal B\) as the true generator. If a Type~I error occurs,
then there exists \(A_i\in\mathcal A\) such that $\widetilde l_{A_i}
<
\widetilde l_{B_j}.$ Therefore, by a union bound, $\mathbb P_{B_j}
\left\{
\text{Type~I error}
\right\}
\leq
\sum_{A_i\in\mathcal A}
\mathbb P_{B_j}
\left\{
\widetilde l_{A_i}
<
\widetilde l_{B_j}
\right\}.$ By \eqref{eq:clipped_gap_BA} and
Lemma~\ref{lemma:clipped_pairwise_comparison}, each probability on the
right-hand side is at most $\exp\left(
-\frac{N\widetilde\epsilon_1^2}{2L_\tau^2}
\right).$ This proves the first part of the proposition. Next, fix \(A_i\in\mathcal A\) as the true generator. If the test fails to
reject \(\mathbf H_0\), then the true model \(A_i\) does not have a score
below every model in \(\mathcal B\). Hence, there exists
\(B_j\in\mathcal B\) such that $\widetilde l_{B_j}
\leq
\widetilde l_{A_i}.$ Thus, by a union bound, $\mathbb P_{A_i}
\left\{
\text{Type~II error}
\right\}
\leq
\sum_{B_j\in\mathcal B}
\mathbb P_{A_i}
\left\{
\widetilde l_{B_j}
\leq
\widetilde l_{A_i}
\right\}.$ Then, \eqref{eq:clipped_gap_AB} and
Lemma~\ref{lemma:clipped_pairwise_comparison} prove the second part of the proposition.
\endproof
\hfill
\Halmos

\subsection{A worst case KL-based sufficient condition for clipped-score separation}
\label{app:clipped_KL_separation}
%%%%%%%%%%%%%%%%%%%%%%%%%%%%%%%%%%%%%%%%%%%%%%%%%%%%%%%%%%%%%%%%%%%%%%%%%%%%

For this subsection, define the normalized clipped conditional distribution
of model \(M\) by 
\begin{equation*}
\overline p_{n,\tau}^{M}(y)
:=
\frac{\max\{p_n^M(y),\tau\}}
{\displaystyle\sum_{z\in\mathcal X}\max\{p_n^M(z),\tau\}}.
\end{equation*}
The clipping correction satisfies $r_{\tau,n}^{M}
=
\log\left(
\sum_{z\in\mathcal X}\max\{p_n^M(z),\tau\}
\right)
-
D_{\mathrm{KL}}
\left(
p_n^M\|
\overline p_{n,\tau}^{M}
\right).$ Since KL divergence is non-negative and $\sum_{z\in\mathcal X}\max\{p_n^M(z),\tau\}
\leq 1+K\tau,$ combined with the definition of bias-corrected clipped-score gap, we have
\begin{align}
\widetilde h_N(B,A)-\widetilde h_N(B,B)
\geq\;&
\frac1N\sum_{n=1}^N
\sum_{y\in\mathcal X}
p_n^B(y)
\log
\frac{\overline p_{n,\tau}^{B}(y)}
     {\overline p_{n,\tau}^{A}(y)}
-
\log(1+K\tau).
\label{eq:clipped_gap_normalized_log_ratio}
\end{align}

For each \(n\), we have $$\sum_{y\in\mathcal X}
p_n^B(y)
\log
\frac{\overline p_{n,\tau}^{B}(y)}
     {\overline p_{n,\tau}^{A}(y)}
\geq
D_{\mathrm{KL}}
\left(
\overline p_{n,\tau}^{B}
\|
\overline p_{n,\tau}^{A}
\right)
-
\left\|
p_n^B-\overline p_{n,\tau}^{B}
\right\|_1
\max_{y\in\mathcal X}
\left|
\log
\frac{\overline p_{n,\tau}^{B}(y)}
     {\overline p_{n,\tau}^{A}(y)}
\right|.$$
Using the definition of \(\overline p_{n,\tau}^{B}\) and the triangle inequality, we obtain
\begin{equation}
\left\|
p_n^B-\overline p_{n,\tau}^{B}
\right\|_1
\leq
\frac{
2\displaystyle\sum_{y\in\mathcal X}
\max\{\tau-p_n^B(y),0\}
}{
1+\displaystyle\sum_{y\in\mathcal X}
\max\{\tau-p_n^B(y),0\}
}.
\label{eq:normalized_clipped_L1_bound}
\end{equation}
Since $\frac{\tau}{1+K\tau}
\leq
\overline p_{n,\tau}^{M}(y)
\leq 1
\quad
\text{for every model }M\text{ and }y\in\mathcal X,$ we also have
\begin{equation}
\max_{y\in\mathcal X}
\left|
\log
\frac{\overline p_{n,\tau}^{B}(y)}
     {\overline p_{n,\tau}^{A}(y)}
\right|
\leq
\log\left(\frac{1+K\tau}{\tau}\right).
\label{eq:normalized_clipped_log_ratio_bound}
\end{equation}
Combining \eqref{eq:clipped_gap_normalized_log_ratio}--%
\eqref{eq:normalized_clipped_log_ratio_bound} yield
\begin{align*}
\widetilde h_N(B,A)-\widetilde h_N(B,B)
\geq\;
\frac1N\sum_{n=1}^N
D_{\mathrm{KL}}
\left(
\overline p_{n,\tau}^{B}
\|
\overline p_{n,\tau}^{A}
\right)
\nonumber-
\frac1N\sum_{n=1}^N
\frac{
2\displaystyle\sum_{y\in\mathcal X}
\max\{\tau-p_n^B(y),0\}
}{
1+\displaystyle\sum_{y\in\mathcal X}
\max\{\tau-p_n^B(y),0\}
}
\log\left(\frac{1+K\tau}{\tau}\right)
-
\log(1+K\tau).
\end{align*}

\begin{remark}
\label{remark:KL_sufficient_for_clipped_separation}
Suppose that, for every \(A_i\in\mathcal A\) and \(B_j\in\mathcal B\),
\begin{equation}
\frac1N\sum_{n=1}^N
D_{\mathrm{KL}}
\left(
\overline p_{n,\tau}^{B_j}
\|
\overline p_{n,\tau}^{A_i}
\right)
\geq
\epsilon_1
\qquad
\mathbb P_{B_j}\text{-almost surely},
\label{eq:normalized_clipped_KL_BA}
\end{equation}
and
\begin{align}
&
\frac1N\sum_{n=1}^N
\frac{
2\displaystyle\sum_{y\in\mathcal X}
\max\{\tau-p_n^{B_j}(y),0\}
}{
1+\displaystyle\sum_{y\in\mathcal X}
\max\{\tau-p_n^{B_j}(y),0\}
}
\log\left(\frac{1+K\tau}{\tau}\right)+
\log(1+K\tau)
\leq
\frac{\epsilon_1}{2}
\qquad
\mathbb P_{B_j}\text{-almost surely}.
\label{eq:normalized_clipped_error_BA}
\end{align}
Suppose also that \eqref{eq:normalized_clipped_KL_BA} and
\eqref{eq:normalized_clipped_error_BA} hold with \(A_i\) and \(B_j\)
interchanged, almost surely under \(A_i\). Then,
Assumption~\ref{assum:clipped_attribution_separation} holds with $\widetilde\epsilon_1=\frac{\epsilon_1}{2}.$
\end{remark}
%%%%%%%%%%%%%%%%%%%%%%%%%%%%%%%%%%%%%%%%%%%%%%%%%%%%%%%%%%%%%%%%%%%%%%%%%%%%

%%%%%%%%
\subsection{Clipped analogue of Proposition~\ref{thm:statstest}}
\label{app:clipped_A_not_A}
%%%%%%%%%%%%%%%%%%%%%%%%%%%%%%%%%%%%%%%%%%%%%%%%%%%%%%%%%%%%%%%%%%%%%%%%%%%%

We next consider the hypotheses as in Section~\ref{sec:human}: $\mathbf H_0:
B\neq A,
\quad
\mathbf H_1:
B=A.$ Using \eqref{eq:corrected_self_center_average}, for a threshold \(t>0\), reject \(\mathbf H_0\) if $\left|
\widetilde l_A(\mathbf Y_N)
-
h_N(A,A)(\mathbf Y_N)
\right|
\leq t.$ The separation condition is as follows.
\begin{assumption}
\label{assum:clipped_A_not_A_separation}
There exists \(\widetilde\epsilon_2>0\) such that every non-\(A\) source \(B\)
under consideration satisfies
\begin{equation}
\left|
\widetilde h_N(B,A)(\mathbf Y_N)
-
h_N(A,A)(\mathbf Y_N)
\right|
\geq
\widetilde\epsilon_2
\qquad
\mathbb P_B\text{-almost surely}.
\label{eq:clipped_A_separation}
\end{equation}
\end{assumption}

The correction \(r_{\tau,n}^{A}\) appears in both terms in \eqref{eq:clipped_A_separation}. Hence, \eqref{eq:clipped_A_separation} is equivalently
\begin{equation*}
\left|
\frac1N\sum_{n=1}^N
\left\{
\sum_{y\in\mathcal X}p_n^B(y)\widetilde z_n^A(y)
-
\sum_{y\in\mathcal X}p_n^A(y)\widetilde z_n^A(y)
\right\}
\right|
\geq
\widetilde\epsilon_2.
\end{equation*}
Thus, the bias correction preserves the ordinary entropy center when \(A\)
generates, while the null-versus-alternative separation is determined by
the difference between the two expected clipped evaluator scores.

\begin{proposition}
\label{prop:clipped_statstest}
Suppose Assumption~\ref{assum:clipped_A_not_A_separation} holds. For every
threshold $0<t<\widetilde\epsilon_2,$
the Type~II error satisfies
\begin{equation*}
\mathbb P_A
\left\{
\text{fail to reject }\mathbf H_0
\right\}
\leq
2\exp\left(
-\frac{2Nt^2}{L_\tau^2}
\right).
\end{equation*}
Uniformly over all non-\(A\) sources \(B\) satisfying Assumption~\ref{assum:clipped_A_not_A_separation}, the Type~I error
satisfies
\begin{equation*}
\mathbb P_B
\left\{
\text{reject }\mathbf H_0
\right\}
\leq
2\exp\left(
-\frac{2N(\widetilde\epsilon_2-t)^2}{L_\tau^2}
\right).
\end{equation*}
\end{proposition}
\proof{Proof.}
Suppose first that \(A\) generated the text. By \eqref{eq:corrected_self_center_average},
$\widetilde h_N(A,A)=h_N(A,A).$
A Type~II error occurs only if $\left|
\widetilde l_{A}
-
\widetilde h_N(A,A)
\right|
>t.$
Applying Lemma~\ref{lemma:clipped_concentration} with \(B=A\) proves the first part of the proposition. Next, suppose a non-\(A\) source \(B\) generated the text. On the Type~I
error event, $\left|
\widetilde l_{A}
-
h_N(A,A)
\right|
\leq t.$
By the reverse triangle inequality, 
\begin{align*}
\left|
\widetilde l_{A}
-
\widetilde h_N(B,A)
\right|
\geq
\left|
\widetilde h_N(B,A)
-
h_N(A,A)
\right|
-
\left|
\widetilde l_{A}
-
h_N(A,A)
\right|
\geq
\widetilde\epsilon_2-t.
\end{align*}
Therefore, $\mathbb P_B
\left\{
\text{Type~I error}
\right\}
\leq
\mathbb P_B\left\{
\left|
\widetilde l_{A}
-
\widetilde h_N(B,A)
\right|
\geq
\widetilde\epsilon_2-t
\right\}.$ Applying Lemma~\ref{lemma:clipped_concentration} with generator \(B\) and
evaluator \(A\) proves the second part of the proposition.
\endproof
\hfill
\Halmos

A balanced threshold is $t=\frac{\widetilde\epsilon_2}{2}.$
For this choice, both error bounds in
Proposition~\ref{prop:clipped_statstest} are at most
$2\exp\left(
-\frac{N\widetilde\epsilon_2^2}{2L_\tau^2}
\right).$
The following more general bound may be useful if the clipped separation
holds with high probability rather than almost surely.

\begin{remark}
\label{remark:high_probability_clipped_separation}
For any \(c>t\), the same triangle-inequality argument gives
\begin{align*}
\mathbb P_B
\left\{
\text{Type~I error}
\right\}
\leq\;
2\exp\left(
-\frac{2N(c-t)^2}{L_\tau^2}
\right)
+
\mathbb P_B\left\{
\left|
\widetilde h_N(B,A)
-
h_N(A,A)
\right|
<c
\right\}.
\end{align*}
Thus, any separate high-probability lower bound on the clipped separation
can be combined with
Lemma~\ref{lemma:clipped_concentration}.
\end{remark}

%%%%%%%%%%%%%%%%%%%%%%%%%%%%%%%%%%%%%%%%%%%%%%%%%%%%%%%%%%%%%%%%%%%%%%%%%%%%
%%%%%%%%%%%%%%%%%%%%%%%%%%%%%%%%%%%%%%%%%%%%%%%%%%%%%%%%%%%%%%%%%%%%%%%%%%%%
\subsection{Relationship with the original \(A\)-versus-not-\(A\) separation}
\label{app:clipped_relationship_original}
%%%%%%%%%%%%%%%%%%%%%%%%%%%%%%%%%%%%%%%%%%%%%%%%%%%%%%%%%%%%%%%%%%%%%%%%%%%%

Suppose that the cross-entropy \(h_N(B,A)\) is finite. Definitions yield
\begin{align}
\widetilde h_N(B,A)-h_N(A,A)
=
h_N(B,A)-h_N(A,A)
-
\frac1N\sum_{n=1}^N
\sum_{\substack{y\in\mathcal X:\\0<p_n^A(y)<\tau}}
\left\{
p_n^B(y)-p_n^A(y)
\right\}
\log\left(\frac{\tau}{p_n^A(y)}\right).
\label{eq:detection_clipping_distortion_identity}
\end{align}

\begin{corollary}
\label{cor:original_detection_separation_implies_clipped}
Suppose Assumption~\ref{assum:mindiff} holds with parameter
\(\epsilon_2>0\). If there exists
\(\widetilde\epsilon_2\in(0,\epsilon_2]\) such that
\begin{align*}
&
\left|
\frac1N\sum_{n=1}^N
\sum_{\substack{y\in\mathcal X:\\0<p_n^A(y)<\tau}}
\left\{
p_n^B(y)-p_n^A(y)
\right\}
\log\left(\frac{\tau}{p_n^A(y)}\right)
\right| \leq
\epsilon_2-\widetilde\epsilon_2
\qquad
\mathbb P_B\text{-almost surely},
\end{align*}
then, Assumption~\ref{assum:clipped_A_not_A_separation} holds.
\end{corollary}

\proof{Proof.}
Equation~\eqref{eq:detection_clipping_distortion_identity} and the reverse
triangle inequality yield $\left|
\widetilde h_N(B,A)-h_N(A,A)
\right|
\geq
\epsilon_2-(\epsilon_2-\widetilde\epsilon_2)
=
\widetilde\epsilon_2.$
\endproof
\hfill
\Halmos

\begin{remark}
\label{remark:clipping_below_original_floor}
Suppose Assumption~\ref{assum:crossmodel} holds with parameter
\(\epsilon\), there is no support mismatch between the relevant generator
and evaluator, and \(\tau\leq\epsilon\). Then, no positive probability is
affected by clipping and \(r_{\tau,n}^M=0\) for every relevant model \(M\).
Consequently, under each generator \(B\),
\[
\widetilde l_M(\mathbf Y_N)=l_M(\mathbf Y_N),
\qquad
\widetilde h_N(B,M)(\mathbf Y_N)=h_N(B,M)(\mathbf Y_N)
\qquad
\mathbb P_B\text{-almost surely}.
\]
So, the clipped tests coincide almost surely with the tests in the main body.
\end{remark}

%%%%%%%%%%%%%%%%%%%%%%%%%%%%%%%%%%%%%%%%%%%%%%%%%%%%%%%%%%%%%%%%%%%%%%%%%%%%
%%%%%%%%%%%%%%%%%%%%%%%%%%%%%%%%%%%%%%%%%%%%%%%%%%%%%%%%%%%%%%%%%%%%%%%%%%%%
\subsection{Numerical example for upper-bounds with \(\tau=10^{-3}\)}
\label{app:clipped_tau_numeric}
%%%%%%%%%%%%%%%%%%%%%%%%%%%%%%%%%%%%%%%%%%%%%%%%%%%%%%%%%%%%%%%%%%%%%%%%%%%%

For the illustrative clipping level $\tau=10^{-3},
$ we have
\[
L_\tau=3\log(10)\approx 6.9078,
\qquad
L_\tau^2\approx47.7171,
\qquad
2L_\tau^2\approx95.434.
\]
Thus, Proposition~\ref{prop:clipped_statstestmulti} yields $\operatorname{Type~I}
\leq
|\mathcal A|
\exp\left(
-\frac{N\widetilde\epsilon_1^2}{95.434}
\right)$ and $\operatorname{Type~II}
\leq
|\mathcal B|
\exp\left(
-\frac{N\widetilde\epsilon_1^2}{95.434}
\right).$ For the threshold
\(t=\widetilde\epsilon_2/2\),
Proposition~\ref{prop:clipped_statstest} yields
\begin{equation}
\max\{\operatorname{Type~I},\operatorname{Type~II}\}
\leq
2\exp\left(
-\frac{N\widetilde\epsilon_2^2}{95.434}
\right).
\label{eq:numeric_clipped_detection}
\end{equation}

To illustrate the magnitude of these guarantees, suppose that the effective
separation after clipping is $\widetilde\epsilon_1=\widetilde\epsilon_2=1.$ A separation of one nat per token corresponds to the evaluator assigning the generator's tokens, on average, $e$ times lower probability than its own, a gap of the order commonly observed between human-written and LLM-generated text under a fixed evaluator; readers can rescale $N$ by $1/\widetilde\epsilon^{2}$ for other values.

For attribution between one model in \(\mathcal A\) and one model in
\(\mathcal B\), requiring either error probability to be at most
\(\alpha\) requires $N
\geq
95.434\log(1/\alpha).$ Thus, an error guarantee of at most \(0.01\) requires $N\geq
95.434\log(100)
\approx439.49,$ or at least \(440\) tokens, while an error guarantee of at most \(0.001\) requires $N\geq
95.434\log(1000)
\approx659.24,$ or at least \(660\) tokens. More generally, when multiple candidate models are included, guaranteeing that both Type~I and Type~II errors are at most \(\alpha\) is ensured by $N
\geq
95.434
\max\left\{
\log\left(\frac{|\mathcal A|}{\alpha}\right),
\log\left(\frac{|\mathcal B|}{\alpha}\right)
\right\}.$

For the \(A\)-versus-not-\(A\) detection test, the leading factor \(2\)
in \eqref{eq:numeric_clipped_detection} requires $N
\geq
95.434\log(2/\alpha).$ Hence, when \(\widetilde\epsilon_2=1\), a Type~I and Type~II error guarantee
of at most \(0.01\) requires $N\geq
95.434\log(200)
\approx505.64,$ or at least \(506\) tokens. Tightening the desired guarantee to
\(0.001\) requires $N\geq
95.434\log(2000)
\approx725.39,$ or at least \(726\) tokens. Thus, for example, a teacher or employer
seeking these guarantees can translate the statistical requirement into
a minimum amount of auditable text: under an effective separation of one,
roughly \(500\)--\(700\) tokens are sufficient for the
\(A\)-versus-not-\(A\) test at the \(1\%\)--\(0.1\%\) error levels,
respectively.

\section{Additional Experiments \& Experiment Details}\label{app:experimentdetails}

\subsection{Black-Box Detection with Sampling Experiments}\label{BBexperiments}

Human text is the null, and text generated by the focal evaluator model is the alternative. We use XSum, SQuAD, and Reddit WritingPrompts with Llama3 (8B), GPT-NeoX Erebus (20B), and Qwen (32B). The first 30 human-written tokens are used as the shared prompt, using the document field for XSum, the context field for SQuAD, and the story field for WritingPrompts; See Appendix~\ref{app:datasetdetails} for dataset details. For computational purposes, we restrict the input texts to a 512-token pre-processing context window, generations are produced from the first 30 prompt tokens with generation capped at 200 model tokens, and each human/LLM pair is trimmed to the shorter word length before evaluation. In this experiment, no additional maximum-token cap is imposed after this trimming, so the statistic is computed over the full resulting text. For each token position, we estimate the evaluator's next-token distribution using \(m \in \{10,100,1000\}\) samples and compare these finite-sampling results with an exact-probability benchmark computed using the evaluator's exact next-token probabilities under the same context window. We define this benchmark as ``Exact probabilities (same context window),'' which uses the evaluator's exact next-token probabilities under the same context window used in this black-box experiment. We report AUROC and TPR at FPR \(\leq 1\%\).

For AUROC, increasing (m) from 10 to 100 and then to 1000 likewise generally moves finite-sampling performance toward the exact-probability benchmark in Table~\ref{tab:blackbox_sampling_auroc_adjusted}. By \(m=1000\), Writing converges fully to the benchmark for all three models, with AUROC values of 1.000, 0.999, and 1.000 for Llama, Neox, and Qwen, respectively. XSum also approaches the benchmark, reaching 0.97, 0.96, and 0.99 for the three models. SQuAD improves with increasing \(m\), but leaves a larger gap from the exact-probability benchmark than in Writing and XSum: at \(m=1000\), AUROC for SQuAD reaches 0.93, 0.92, and 0.81. This pattern is consistent with the short, factual, Wikipedia-derived nature of responses in SQuAD, where human and LLM continuations can be closer under the evaluator than in creative writing or news-style text. 

\begin{table}[H]
\centering
\caption{AUROC - Black-box detection with sampling.}
\label{tab:blackbox_sampling_auroc_adjusted}
\resizebox{\textwidth}{!}{
\begin{tabular}{lccccccccc}
\toprule
 & \multicolumn{3}{c}{XSum} & \multicolumn{3}{c}{SQuAD} & \multicolumn{3}{c}{WritingPrompts} \\
\cmidrule(lr){2-4} \cmidrule(lr){5-7} \cmidrule(lr){8-10}
Method 
& Llama3 (8B) & GPT-NeoX Erebus (20B) & Qwen (32B) 
& Llama3 (8B) & GPT-NeoX Erebus (20B) & Qwen (32B) 
& Llama3 (8B) & GPT-NeoX Erebus (20B) & Qwen (32B) \\
\midrule
Exact probabilities (same context window)             & 0.99 & 0.98 & 1.00 & 0.98 & 0.95 & 0.93 & 0.99 & 1.00 & 1.00 \\
Black-box, $m=10$     & 0.97 & 0.60 & 0.99 & 0.84 & 0.57 & 0.72 & 1.00 & 0.89 & 1.00 \\
Black-box, $m=100$    & 0.97 & 0.90 & 0.99 & 0.92 & 0.82 & 0.79 & 1.00 & 0.99 & 1.00 \\
Black-box, $m=1000$   & 0.98 & 0.96 & 0.99 & 0.93 & 0.92 & 0.81 & 1.00 & 1.00 & 1.00 \\
\bottomrule
\end{tabular}
}
\end{table}

For TPR at FPR \(\leq 1\%\), increasing \(m\) from 10 to 100 and then to 1000 generally moves the black-box statistic toward the exact-probability benchmark computed under the same context window, i.e., \textit{Exact probabilities (same context window)} in Table~\ref{tab:TPRblackboxsampling}. Writing essentially converges for all three models by $m=1000$. In XSum, the remaining gap for Neox/XSum can be interpreted as a finite-sampling gap relative to the exact-probability benchmark: in the black-box construction, \(p_n^A\) is replaced by the empirical distribution \(\hat p_n^A\), and for Neox/XSum, \(m=1000\) is not sufficient to fully preserve the low-FPR separation obtained with exact probabilities. This is plausible for news text because named entities, factual details, and topic-specific continuations can distribute probability mass across several admissible next tokens, making the empirical approximation require a larger \(m\).

SQuAD is the main exception: at (m=1000), several sampled results remain below the exact-probability benchmark. SQuAD consists of Wikipedia-derived factual contexts, where human and LLM continuations can be very similar under the evaluator, especially when the answer context is short. When this underlying difference between human and LLM continuations is small, estimation error in the sampled next-token probabilities has a larger impact on the test statistic, so convergence of the black-box detector to the exact-probability benchmark requires a larger sampling budget (m). Consequently, while (m=1000) is sufficient to nearly recover the exact-probability benchmark in WritingPrompts and most XSum settings, it remains insufficient for some SQuAD configurations. The managerial takeaway is that for creative writing and for most news-style contexts, \(m \leq 1000\) can yield reliable TPR at low FPR, but for factual Wikipedia-like assignments with short factual answers, the framework must be used cautiously. In such settings, users should increase \(m\) when computation allows and design assignments to require explanatory writing around the answer, e.g., reasoning, interpretation, or comparison, rather than only a short factual response.

%For AUROC, increasing \(m\) produces a more consistent movement toward the exact-probability benchmark because AUROC aggregates performance over all thresholds rather than focusing only on the low-FPR region. By \(m=1000\), Writing is essentially at the exact-probability benchmark, XSum is close for Llama and Qwen and substantially improved for Neox, while SQuAD remains the weakest dataset and does not fully converge to the benchmark. This again reflects the factual nature of SQuAD: the limitation is not only finite sampling, but also the closeness of human and LLM continuations in factual contexts. The practical implication is to interpret AUROC and TPR at FPR \(\leq 1\%\) jointly in such domains, rather than relying only on the low-FPR operating point.

\begin{table}[H]
\centering
\caption{TPR at FPR $\leq 1\%$ - Black-box detection with sampling.}
\label{tab:TPRblackboxsampling}
\resizebox{\textwidth}{!}{
\begin{tabular}{lccccccccc}
\toprule
 & \multicolumn{3}{c}{XSum} & \multicolumn{3}{c}{SQuAD} & \multicolumn{3}{c}{WritingPrompts} \\
\cmidrule(lr){2-4} \cmidrule(lr){5-7} \cmidrule(lr){8-10}
Method 
& Llama3 (8B) & GPT-NeoX Erebus (20B) & Qwen (32B) 
& Llama3 (8B) & GPT-NeoX Erebus (20B) & Qwen (32B) 
& Llama3 (8B) & GPT-NeoX Erebus (20B) & Qwen (32B) \\
\midrule
Exact probabilities (same context window)             & 0.93 & 0.83 & 0.98 & 0.87 & 0.48 & 0.52 & 1.00 & 1.00 & 1.00 \\
Black-box, $m=10$     & 0.78 & 0.03 & 0.95 & 0.02 & 0.00 & 0.05 & 1.00 & 0.25 & 1.00 \\
Black-box, $m=100$    & 0.86 & 0.37 & 0.95 & 0.22 & 0.09 & 0.12 & 1.00 & 0.80 & 1.00 \\
Black-box, $m=1000$   & 0.89 & 0.71 & 0.96 & 0.50 & 0.47 & 0.24 & 0.99 & 0.92 & 1.00 \\
\bottomrule
\end{tabular}
}
\end{table}

\subsection{Dataset details} \label{app:datasetdetails}
\begin{figure}[H]
    \centering    \includegraphics[width=0.35\linewidth]{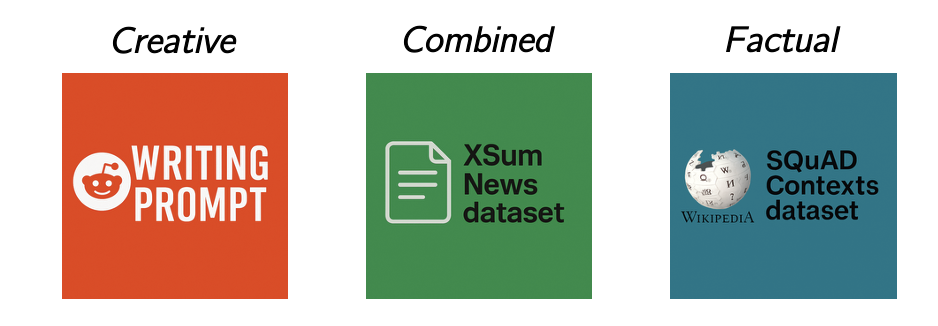}
    \caption{Overview of the three datasets used in our detection experiments.}
    \label{fig:datasets-overview}
\end{figure}
We use the codes in the official GitHub repository for DetectGPT \citep{mitchell2023detectgpt}, which is publicly available under the MIT license in \url{https://github.com/eric-mitchell/detect-gpt} for our white-box experiments. 
The datasets for this set of experiments are XSum, SQuAD, and 
Reddit writingPrompts that we briefly describe below. We use the codes in the official GitHub repository for the RAID dataset \cite{dugan2024raid} for our black-box experiments. The RAID dataset uses a variety of datasets from 8 domains, briefly described below. For details about the RAID dataset, please refer to Appendix E.1 in \cite{dugan2024raid}.

\textbf{XSum.} 
The Extreme Summarization (XSum) dataset comprises 226{,}711 BBC news articles paired with single-sentence summaries for each article. Each entry includes a \texttt{document} (the full news article), a \texttt{summary} (a concise, one-sentence abstraction), and an \texttt{id} (the BBC article identifier). The dataset is designed to support the task of extreme abstractive summarization, where models must generate highly condensed summaries that capture the essence of the source text. The XSum dataset is publicly available under the Creative Commons Attribution-ShareAlike 4.0 International (CC BY-SA 4.0) license and can be accessed at \url{https://huggingface.co/datasets/EdinburghNLP/xsum}. In our experiments, we utilize the first 30 tokens of the \texttt{document} column as input for creating LLM-generated text.

% The dataset is partitioned into training (204{,}045 examples), validation (11{,}332 examples), and test (11{,}334 examples) splits. 

\textbf{SQuAD.}
The Stanford Question Answering Dataset (SQuAD) is a benchmark for machine reading comprehension, consisting of over 100,000 question-answer pairs derived from 536 Wikipedia articles. Each example includes a \texttt{context} paragraph, a \texttt{question}, and one or more \texttt{answers}, where each answer is a span of text from the corresponding context. SQuAD is publicly available under the Creative Commons Attribution-ShareAlike 4.0 International (CC BY-SA 4.0) and can be accessed at \url{https://huggingface.co/datasets/rajpurkar/squad}. In our experiments, we utilize the first 30 tokens of the \texttt{context} column as input for creating LLM-generated text.
%The dataset is partitioned into training (87,599 examples) and validation (10,570 examples) splits. 
%under the Creative Commons Attribution ShareAlike 4.0 International license

\textbf{Reddit WritingPrompts.}
The Reddit WritingPrompts dataset comprises over 300,000 prompt-story pairs collected from the r/WritingPrompts subreddit. Each entry includes a \texttt{prompt} (a creative writing prompt) and a corresponding \texttt{story} (a user-generated narrative response). The dataset is designed to support research in open-ended story generation and narrative modeling. The dataset is publicly available under the MIT License and can be accessed at \url{https://huggingface.co/datasets/euclaise/writingprompts}. In our experiments, we utilize the first 30 tokens of the \texttt{story} field as input for creating LLM-generated text.
%The dataset is partitioned into training (272,600 examples), validation (15,620 examples), and test (15,138 examples) splits. 
%under the MIT license

\textbf{RAID.}
The Robust AI Detection (RAID) dataset is a large-scale benchmark constructed to evaluate the reliability of LLM-generated text detectors across a wide range of textual domains. It comprises over 6 million generations created by 11 language models under 4 decoding strategies and 11 adversarial attacks. RAID includes 8 carefully selected text domains to capture a broad spectrum of natural language styles, structures, and levels of creativity. The domains are

\textbf{Abstracts.} Sourced from arXiv.org, this domain consists of paper abstracts across multiple disciplines. Texts are technical, information-dense, and syntactically formal, reflecting real-world scientific writing in academic publishing.

\textbf{Books.} Drawn from the BookCorpus dataset, this domain features long-form fictional passages including dialogue and narration. It captures informal, imaginative writing with strong character and plot structure diversity.

\textbf{News.} Extracted from the CNN/DailyMail summarization corpus, the news domain includes professionally written news stories. These texts are structured with leads and summaries, and often exhibit a neutral tone, making them ideal for assessing generation consistency in journalistic formats.

\textbf{Poetry.} Taken from Project Gutenberg, the poetry domain includes English poems from public domain authors. The texts exhibit high stylistic variability, creative syntax, line breaks, and metaphorical expressions, which are characteristics that typically challenge detection.

\textbf{Recipes.} Based on the RecipeNLG dataset, this domain contains step-by-step cooking instructions and ingredient lists. It features a highly structured and instructional tone with imperative phrasing and domain-specific vocabulary.

\textbf{Reddit.} Sourced from the WritingPrompts subreddit, these texts are user-generated responses to open-ended prompts. The domain features creative fiction in a casual tone, reflecting informal narrative style and internet native conventions.

\textbf{Reviews.} Pulled from Amazon product reviews, this domain contains subjective, opinionated writing with informal grammar and varying tone. It reflects real-world user-generated content that combines evaluation, anecdote, and justification.

\textbf{Wikipedia.} This domain includes encyclopedic entries from Wikipedia, selected for their balanced, objective tone, structured format, and high factual density. Texts are varied in topic and are typically well-edited and linguistically standardized.

To avoid data contamination, all human-written documents were sourced from publicly available corpora with timestamps preceding 2022. For each document, the dataset provides one generation per model, decoding strategy, and adversarial attack. The dataset is publicly available at \url{https://huggingface.co/datasets/liamdugan/raid} under the MIT License.

%\subsection{LLM details}

\subsection{Decoding strategies \& Adversarial attacks} \label{app:adversarial attacks}

\subsubsection{Decoding strategies}
A decoding strategy determines how tokens are selected from the LLM's (conditional) probability distribution in the text generation process. Greedy decoding selects the most likely (top-ranked) token from the (conditional) probability distribution. Previous research has shown that sampling from the probability distribution rather than selecting the top-ranked token (as done in greedy decoding) can make the LLM-generated text less detectable \cite{ippolito2019automatic}. Based on these findings, \cite{dugan2024raid} generates two outputs per prompt: one with greedy decoding and the other with random sampling. 

The repetition penalty, as defined in \cite{keskar2019ctrl}, reduces the likelihood of generating previously used tokens through down-weighting the probability of choosing the same word in the following tokens using a parameter $\theta$. This penalty examines whether the less-repetitive output (that has undergone repetition penalty) can reduce the detectability of LLM-generated text. Formally, Given a temperature \( T > 0 \) and a set of scores \( s_i \in \mathbb{R}^d \) for each token \( i \) in a vocabulary, the probability \( p_i \) of predicting the $i^{\text{th}}$ token is 
\[
p_i = \frac{\exp(s_i / T)}{\sum_j \exp(s_j / T)}.
\]
The repetition penalty defined by \cite{keskar2019ctrl} modifies the distribution as 
\[
\tilde{p}_i = \frac{\exp\left(\frac{s_i}{T \cdot \mathbb{I}(i \in g)}\right)}{\sum_j \exp\left(\frac{s_j}{T \cdot \mathbb{I}(j \in g)}\right)},
\]
where \( g \) is the list of previously generated tokens, and the indicator function \( \mathbb{I}(c) \) is
\[
\mathbb{I}(c) =
\begin{cases}
\theta & \text{if } c \text{ is true}, \\
1 & \text{otherwise}.
\end{cases}
\]

\cite{dugan2024raid} uses \cite{keskar2019ctrl}'s method and generates two additional outputs: with and without repetition penalty. Following \cite{keskar2019ctrl}, we use $\theta$ = 1.2 for the experiments.

\subsubsection{Adversarial attacks}

Adversarial attacks in LLM detection are strategically crafted paraphrases or perturbations applied to LLM-generated text that aim to bypass detection by reducing detectable differences between LLM-generated and human-written text. \cite{dugan2024raid} produces different adversarially attacked versions of texts for each LLM-generated text. Below are brief descriptions of the attacks and sources from which \cite{dugan2024raid} acquired them.

\textbf{Insert Paragraphs.} \cite{bhat2020effectively} injects extra line breaks by putting \texttt{\textbackslash n\textbackslash n} between sentences. \cite{dugan2024raid} does the following process: First, they use \texttt{Punkt} to break the LLM-generated text input into individual sentences. Then, they sample $\theta$ percent of the spans generated by \texttt{Punkt}, and add \texttt{\textbackslash n\textbackslash n} between sentences to cause a paragraph break.

\textbf{Article Deletion.} \cite{liang2023mutation,guerrero2022mutation} first search through the text and capture all articles ``the," ``a," and ``an" using POS tagging. Then, using random sampling, select a fixed $\theta$ percent of the articles and apply the article deletion to them.

\textbf{Alternative Spelling.} \cite{liang2023gpt} constructs a mapping between British and American spelling of words using an American to
British English dictionary (\url{https://github.com/hyperreality/
American-British-English-Translator}). Then, searches in the LLM-generated text to identify all the words (that also exist) in the dictionary, randomly samples a fixed $\theta$ percent of the words (with two spellings), and replaces them with the alternative spelling. 

\textbf{Misspelling.} \cite{liang2023mutation,gagiano2021robustness} use Wikipedia's commonly misspelled English words dictionary (\url{https://en.wikipedia.org/wiki/Commonly_ misspelled_English_words}), and find the LLM-generated words that also exist in the misspelling dictionary. 

\textbf{Upper Lower Swap.} \cite{gagiano2021robustness} randomly selects a fixed $\theta$ percent of the words in the LLM-generated text, and changes the first letter to uppercase if originally lowercase, and vice versa. 

\textbf{Number Swap.} \cite{bhat2020effectively} first searches through the text, and captures all numerical digits in the LLM-generated text. Then, randomly selects a fixed $\theta$ percent of the digits and replaces them with a randomly selected digit from 0 to 9. 

\textbf{Whitespace Addition.} \cite{cai2023evade,gagiano2021robustness} capture all the spaces between tokens, randomly select a fixed $\theta$ percent of the spaces, and insert one additional space to each selected space. Note that since the sampling is performed with replacement, the same inter-token space can be chosen multiple times, resulting in a possible insertion of more than one extra space between tokens. 

\textbf{Paraphrasing.} \cite{krishna2023paraphrasing,sadasivan2023can} transform the source LLM-generated text into an alternative text with the same semantic content (i.e., the conveyed meaning is equivalent, but the content is statistically and stylistically different). \cite{dugan2024raid} uses the paraphrasing method in \cite{krishna2023paraphrasing}'s DIPPER-11B using their HuggingFace repository. DIPPER-11B is a paraphrasing model built on the T5-XXL architecture. The process includes two control parameters: one for lexical diversity (a measure of the variety of the words used in the text) and another for reordering, which adjusts syntactic transformation. DIPPER-11B encodes the LLM-generated text as input and then decodes a paraphrase as output. \cite{dugan2024raid} uses the default setting from \cite{krishna2023paraphrasing} to apply the paraphrasing attack on all texts in the dataset.

\textbf{Synonym Swap.} \cite{pu2023deepfake} follows the process described in \cite{dugan2024raid}: first, they mask all tokens, replacing each with a mask-fill from the top-20 candidates in BERT \cite{devlin2019bert}. Next, they assign the Part-of-Speech (POS) tag, which is a label assigned to each word in a sentence to indicate its grammatical category (such as noun, verb, adjective, or adverb), to each of the mask-fill candidates and remove all those candidates whose POS tag does not match the POS tag for the token in the input text. After creating the final list of valid candidate swaps, they select a fixed $\theta$ percent of the texts for performing the Synonym Swap adversarial attack.

%Synonym Swap (SS): swaps tokens with highly similar BERT (Devlin et al. [2019]) candidate tokens.

%Synonym swap replaces surface-level tokens with semantically equivalent but lexically distinct items, shifting the token distribution away from the detector models' reference distribution. Whitespace manipulations alter standard sub-word segmentation by inserting or removing token boundaries, creating a discrepancy between the text's actual token frequency and what the detector expects.

\textbf{Homoglyph.} \cite{wolff2020attacking,gagiano2021robustness}  swap some English characters with visually similar yet technically different characters from the Cyrillic scripts in \cite{wolff2020attacking}. Specifically, \cite{dugan2024raid} uses homoglyphs for the following ASCII characters: a, A, B, e, E, c, p, K, O, P, M, H, T, X, C,
y, o, x, I, i, N, and Z. They identify all these characters in the LLM-generated text, and then replace all of them (i.e., $\theta$=100\%) with the homoglyph counterparts. If a character has more than one homoglyph, they randomly select between them. 

\textbf{Zero-width Space.} \cite{guerrero2022mutation} inserts 
the Unicode zero-width space character \texttt{U+200B}, which is an invisible character that occupies space in the underlying text encoding without being visible to human readers. \cite{dugan2024raid} inserts a zero-width space before and after every visible character in the perturbed text. Note that \cite{dugan2024raid}'s dataset for the Homoglyph and Zero-width space attacks is incomplete, not allowing us to provide results for them. Nevertheless, these two attacks are easy to remove with a simple preprocessing step we implement in our code (details in Appendix \ref{app:ourasset}): First, remove or normalize invisible characters, such as zero-width spaces. Then map suspicious non-ASCII characters to standard ASCII or a canonical form. To find the $\theta$ that \cite{dugan2024raid} uses for each adversarial attack, please refer to Table 11 in their paper.

\subsection{Detector details}  \label{app:detectordetails}
RAID \citep{dugan2024raid} evaluates 12 detection methods from neural classifiers, metric-based (zero-shot) detectors, and commercial detectors. All detectors were evaluated using their official implementations or web interfaces without fine-tuning on RAID data. Below is a brief description of each detector and how it was run in \cite{dugan2024raid}. To check how \cite{dugan2024raid} finds the thresholds to achieve FPR of 5\%, please refer to Appendix F-2 in \cite{dugan2024raid}.

\noindent\textbf{RoBERTa (GPT2):} fine-tuned to distinguish between human and GPT-2-generated text. It was evaluated using the released Hugging Face checkpoint without modification. RoBERTa-B (GPT2) is a smaller RoBERTa-base version trained on the same dataset. It is computationally lighter, but typically less accurate than the large model.

\noindent\textbf{RoBERTa-B (ChatGPT)} (\citealp{guo2023close}) fine-tuned to detect generations from ChatGPT. \cite{dugan2024raid} downloads and queries the detector via HuggingFace datasets with the unique identifier \texttt{Hello-SimpleAI/chatgpt-detector-roberta}.

\noindent\textbf{RADAR} (\citealp{hu2023radar}) a robust RoBERTa-based detector adversarially trained to resist paraphrasing and rewriting attacks. Used as released without retraining. \cite{dugan2024raid} downloads and queries the detector via HuggingFace datasets with the unique identifier \texttt{TrustSafeAI/RADAR-Vicuna-7B}.

\noindent\textbf{GLTR} (\citealp{gehrmann2019gltr}) evaluates the likelihood of text according to an (evaluator) LLM, bins tokens based on their likelihoods, and uses the bins as signals for detection. \cite{dugan2024raid} uses the default settings from the GLTR repository (\url{https://github.com/HendrikStrobelt/detecting-fake-text}), which is setting the (evaluator) LLM to GPT2 small and the cutoff threshold at rank=10.

%Uses GPT-2 to compute token-level log-probabilities and highlights sequences with unusually high predictability, indicating machine origin. Run with the default GPT-2 model.

\noindent\textbf{Fast-DetectGPT} (\citealp{bao2024fast}) DetectGPT evaluates log-probability curvature under paraphrased perturbations. Low curvature regions suggest LLM generations. It was implemented using DetectGPT's official repository with standard settings. A more efficient variant of DetectGPT that avoids expensive perturbation by approximating curvature metrics is Fast-DetectGPT \cite{bao2024fast}. It was implemented using the default repository configuration, GPTNeo-2.7B as the scoring model, and GPT-J-7 B as the reference model.

\textbf{Binoculars} (\citealp{hans2024spotting}) uses
perplexity divided by cross-perplexity for detection. RAID repository uses the code from the Binocular GitHub repository and calculates perplexity using the default models from Binocular's repository, which are Falcon 7B and Falcon 7B Instruct. 
%compares token-level log-probabilities from a suspected generator model and a reference model to identify generation discrepancies. Used as released in Hans et al. (2024).

\textbf{LLMDet} (\citealp{wu2023llmdet}) uses proxy-perplexity, an approximation of perplexity calculated by repeatedly sampling n-grams from models rather than running the models, of the input text from 10 different small language models. None of the evaluator models used in the detector was used for text generation.

\textbf{GPTZero} (\cite{tiangptzero}) relies on two metrics: \textit{perplexity}, which measures how predictable the text is under an evaluator LLM, and \textit{burstiness}, which captures the variation in perplexity across sentences. Human-written text typically exhibits higher burstiness, while LLM-generated text tends to be more predictable \cite{gptzero_perplexity_burstiness}. \cite{dugan2024raid} queried the detector using the v2 API (\url{https://api.gptzero.me/v2/predict/text}) and threshold on the \texttt{completely\_generated\_prob} field.

\textbf{Originality} \cite{dugan2024raid} queried this detector through the v1 API (\url{https://api.originality.ai/api/v1/scan/ai}) and threshold on the \texttt{score} field in the output JSON. They queried the multilingual "version 3'' of the detector for the Czech and German news domains.

%A commercial platform targeting plagiarism and AI detection in web publishing. Accessed via API with default thresholds.

\textbf{Winston} \cite{dugan2024raid} queried this model through the v1 API (\url{https://api.gowinston.ai/functions/v1/predict}), set the input language to English, unless the text is in German, where they set the language to German. They threshold on the \texttt{score} field in the output JSON.
%A commercial tool for educational and publishing sectors. Texts submitted via web interface.
  
\textbf{ZeroGPT} \cite{dugan2024raid} queried the API (\url{https://api.zerogpt.com/api/detect/detectText}), and used the \texttt{isHuman} field as the classifier output.

%A widely used web-based detector that returns binary labels and probability scores. Evaluated using its public interface.

\noindent\textbf{Comparison with Binoculars (white-box access detection).} \cite{hans2024spotting} attempts to normalize away domain difficulty using a two-model construction: It pairs an observer model with a closely related performer model, computes the observer's usual perplexity on the candidate text, and then computes a cross-perplexity term that measures how well the observer agrees with the performer's next-token distribution; taking a (log-)ratio of these quantities reduces sensitivity to unfamiliar topics that can inflate likelihood-based scores. The key point is that the performer is not assumed to be the true generator of the text; it is an internal proxy meant to represent typical machine continuation in that context. This method works particularly well for longer, open-ended generations where model-generated text tends to follow trajectories that remain consistently machine-typical across LLMs, while human writing introduces more idiosyncratic choices that reduce cross-model agreement. However, this method degrades when outputs are short or highly constrained, e.g., in fact-heavy datasets, because humans and LLMs often select a similar, narrow set of high-probability tokens. When generators become very large/strong, the method can also degrade because the performer may fail to be a representative proxy for the target generator class, and the separation intuition in \cite{hans2024spotting} that the performer is closer to the observer than humans are can weaken as generators approach human-like compatibility with the observer’s predictions; in both cases, the cross-perplexity baseline no longer plays its intended role and the ratio becomes less useful.

\noindent\textbf{Comparison with Fast-DetectGPT and Binoculars (no access to evaluator).} Fast-DetectGPT works well in adversarial attack settings when small random perturbations produce a clear, consistent scoring shift, typically on longer, regularly tokenized passages where LLM outputs follow high-probability continuations (e.g., greedy decoding), but it weakens on short or lexically variable text where the perturbation signal is noisy. Binoculars helps when perplexity in isolation is entangled with topic or style by contrasting two models’ likelihoods to cancel domain and prompt effects, though it can fail if that normalization also removes the key gap (e.g., both models are similarly mismatched to the domain, or the generator is close to a reference model), and it is also less reliable on short inputs. By contrast, our method is more robust when decoding heuristics introduce occasional low-probability word choices that can look human to perplexity alone, because it can still detect the resulting mismatch between surprisal and contextual uncertainty.

\subsection{Hardware} \label{app:hardwaredetails}
All experiments were conducted using a RunPod cloud instance configured with a B200 GPU, 28 vCPUs, and 283 GB of RAM. The setup provided 150 GB of disk and pod volume and achieved bandwidth speeds of up to 20,801 Mbps in the EU-RO-1 region. Experiments were executed in an on-demand secure cloud environment available at \url{https://www.runpod.io}. Our complete set of experiments required a total of approximately 12 hours of runtime. Because our method is nonparametric and does not require training, the runtime scales linearly with the number of input texts in the dataset. Figure \ref{fig:hardware} demonstrates the RunPod cloud configuration.
\begin{figure}[H]
    \centering
    \includegraphics[width=1.2\textwidth]{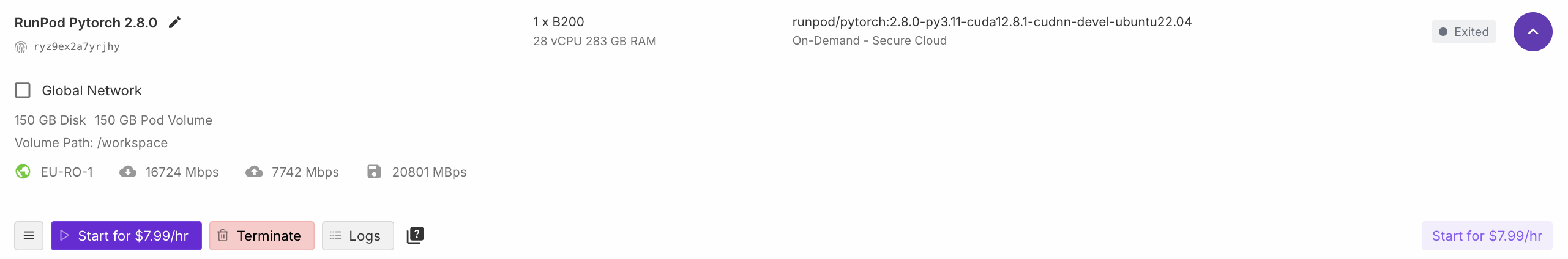}
    \caption{RunPod cloud configuration used for all experiments.}
    \label{fig:hardware}
\end{figure}

\subsection{Code instructions}\label{app:ourasset}
\subsubsection*{Contents.}
We include our self-contained codes as part of our supplementary materials to enable replication of all experiments presented in this work. The repository comprises three Jupyter notebooks: 
\texttt{Whiteboxattribution.ipynb} for identifying the generating model among two candidate sets (codes for the experiments in Section \ref{exp:attribution}), \texttt{Whiteboxdetection.ipynb} for statistical detection tests in a white-box setting (codes for the experiments in Section \ref{whiteboxnumerical}), and \texttt{Blackbox\_adversarial.ipynb} for evaluating black-box detection and robustness against adversarial attacks (codes for the experiments in Section \ref{blackboxnumerical}). All experiments were run on RunPod (B200) and used Hugging Face models and datasets where applicable. 

\subsubsection*{How to run.}
Each notebook is fully self-contained and does not require external data downloads or model training, facilitating straightforward execution on personal devices. Each notebook includes i) Installation and environment setup (in the first cells), ii) Data loading and preparation, iii) Full experiment code with results.
No setup beyond running the notebook cells is required since all dependencies are installed inline.

\subsubsection*{Datasets.}
Most datasets used in the experiments are downloaded programmatically from Hugging Face. 

However, if you want to run the WritingPrompts experiments, you will need to download the WritingPrompts dataset manually from \url{https://github.com/facebookresearch/WritingPrompts}. Then, you need to place the downloaded data in the following path: \texttt{data/writingPrompts/}.

Portions of this code are adapted from DetectGPT, available at \url{https://github.com/eric-mitchell/detect-gpt/tree/main}. We thank the authors of DetectGPT for open-sourcing their codebase.

\end{APPENDICES}

%%%%%%%%%%%%%%%%%
\end{document}